\documentclass[twoside,letterpaper]{article}

\usepackage[preprint]{aistats2027}
\usepackage[round]{natbib}
\setcitestyle{authoryear,round}

\usepackage[utf8]{inputenc}
\usepackage[T1]{fontenc}
\usepackage[hidelinks]{hyperref}
\usepackage{url}
\usepackage{booktabs}
\usepackage{array}
\usepackage{tabularx}
\usepackage{amsfonts}
\usepackage{amsmath}
\usepackage{amssymb}
\usepackage{amsthm}
\usepackage{microtype}
\usepackage{xcolor}
\usepackage{pifont}
\usepackage{graphicx}
\usepackage{subcaption}
\usepackage{algorithm}
\usepackage{algorithmic}
\usepackage{multirow}
\usepackage{placeins}
\usepackage{needspace}
\usepackage{bm}
\usepackage{mathtools}
\usepackage{titletoc}
\usepackage{listings}

\definecolor{lstkeyword}{RGB}{0,0,190}
\definecolor{lstconst}{RGB}{125,0,150}
\definecolor{lststring}{RGB}{163,21,21}
\definecolor{lstcomment}{RGB}{80,110,80}
\definecolor{lstfunction}{RGB}{0,95,135}
\definecolor{lstbackground}{RGB}{247,247,250}
\definecolor{lstrule}{RGB}{160,160,170}
\definecolor{lstnumber}{RGB}{140,140,150}
\AtBeginDocument{\renewcommand{\eqref}[1]{\textup{Eq.~\ref{#1}}}}


\theoremstyle{plain}
\newtheorem{theorem}{Theorem}
\newtheorem{proposition}[theorem]{Proposition}
\newtheorem{lemma}[theorem]{Lemma}
\newtheorem{corollary}[theorem]{Corollary}
\theoremstyle{definition}
\newtheorem{example}[theorem]{Example}

\newcommand{\reals}{\mathbb{R}}

\newcommand{\loss}{\mathcal{L}}

\newcommand{\param}{\bm{\theta}}
\newcommand{\grad}{\nabla}
\newcommand{\mat}[1]{\bm{#1}}
\newcommand{\transpose}{\top}
\newcommand{\norm}[1]{\left\lVert #1 \right\rVert}
\newcommand{\frobnorm}[1]{\left\lVert #1 \right\rVert_{\mathrm{F}}}
\newcommand{\abs}[1]{\left\lvert #1 \right\rvert}
\newcommand{\ie}{\textit{i.e.},}
\newcommand{\eg}{\textit{e.g.},}

\definecolor{cmarkgreen}{RGB}{0,145,60}
\definecolor{xmarkred}{RGB}{205,35,40}
\newlength{\cmarkwidth}
\DeclareRobustCommand{\cmark}{\textcolor{cmarkgreen}{\ding{51}}}
\DeclareRobustCommand{\xmark}{\settowidth{\cmarkwidth}{\ding{51}}%
  \textcolor{xmarkred}{\makebox[\cmarkwidth]{\ding{55}}}}

\begin{document}

\runningtitle{Same Function, Different Trajectory}
\runningauthor{Bum Jun Kim}

\begingroup
\renewcommand{\thefootnote}{\fnsymbol{footnote}}
\twocolumn[
\aistatstitle{Same Function, Different Trajectory: No Standard Optimizer\\ Is Equivariant to Transformer Reparameterization}
\aistatsauthor{Bum Jun Kim\rlap{\footnotemark[1]}}
\aistatsaddress{The University of Tokyo\\ \texttt{bumjun.kim@weblab.t.u-tokyo.ac.jp}}
]
\footnotetext[1]{Corresponding author}
\endgroup

\begin{abstract}
We show that two Transformers representing the same function can follow different training trajectories. For example, doubling a query matrix and halving its key matrix preserves the model's function but changes the gradient scales seen by the optimizer. Our theoretical analysis establishes that standard block-local first-order optimizers fail to scale updates with parameters under a shared learning rate. For AdamW, transforming the first- and second-moment estimates alone does not correct the mismatch between parameter and update scales. To make runs restarted from equivalent checkpoints follow the same trajectory, we balance matrix pairs and transform their moment estimates together, bringing distinct but equivalent training states to the same starting point. To correct scale imbalance between paired Transformer matrices without changing predictions, we propose Symmetry-Aware Gradient Geometry Stabilization (SAGGS), which periodically equalizes the Frobenius norms of exact symmetry pairs, and its damped variant, Soft SAGGS. SAGGS reduces perplexity (PPL) by $0.374\pm0.026$ at 101M parameters on WikiText-103, and Soft SAGGS reduces PPL by $0.780\pm0.040$ in a 31M modern-architecture replication, both under scale imbalance. In a 123.8M checkpoint-resume experiment, reciprocally rescaling matrix pairs barely changes validation loss at restart but increases final PPL by $3.78\pm0.42$; hard and soft rebalancing recover to within $0.02$ PPL of the control.
\end{abstract}

\begin{figure*}[t!]
\centering
\includegraphics[width=\linewidth]{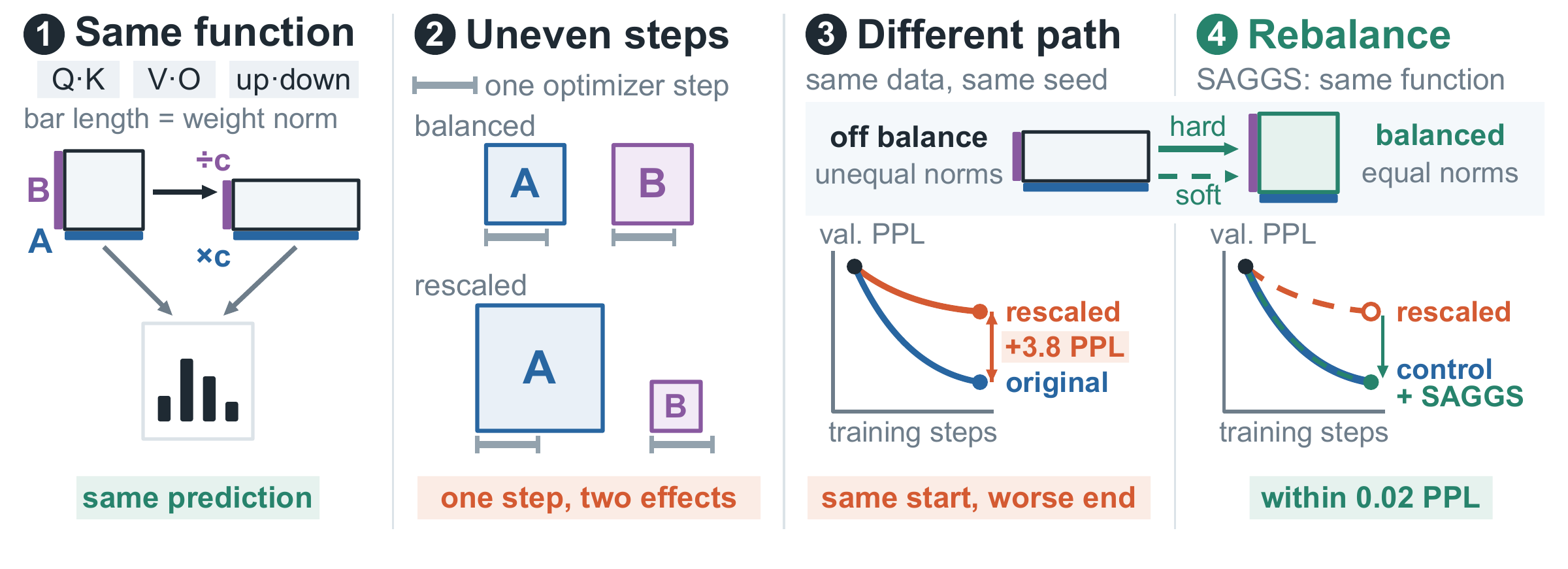}
\caption{Same function, different trajectory. (1) Multiplying $\mat{A}$ by $c$ and dividing its partner $\mat{B}$ by $c$ leaves every prediction unchanged. The rectangle's sides are the Frobenius norms of $\mat{A}$ and $\mat{B}$, so the rectangle changes shape but keeps its area. (2) The same numerical update causes unequal relative changes in the two blocks. (3) Resuming training from the rescaled checkpoint, whose pair norms are imbalanced but whose function is unchanged, ends with much worse perplexity (PPL) than resuming from the original. (4) Hard and Soft SAGGS rebalance the pairs without changing the current function and recover from this degradation.}
\label{fig:overview}
\end{figure*}

\section{Introduction}
\label{sec:intro}

Transformers~\citep{vaswani2017attention} serve as the standard architecture for large language models~\citep{radford2019language,touvron2023llama}. Each Transformer layer consists of a multi-head self-attention block and a multi-layer perceptron (MLP). The attention block is built from query, key, value, and output projections, and the MLP is based on rectified linear units (ReLU) or Swish-gated linear units (SwiGLU)~\citep{shazeer2020glu}.

Several of these weight matrices form pairs whose scales can be traded against each other without changing the output. For example, attention compares queries and keys through the product $(\mat{W}^Q)^\transpose\mat{W}^K$ of the query and key matrices, so replacing $(\mat{W}^Q,\mat{W}^K)$ with $(c\mat{W}^Q,c^{-1}\mat{W}^K)$ for any $c>0$ leaves every attention score, and hence every prediction, unchanged. We call a pair of matrices $(\mat{A},\mat{B})$ an exact symmetry pair when replacing the pair with $(c\mat{A},c^{-1}\mat{B})$ preserves the model's function for every $c>0$. Rescaling the exact symmetry pairs of a model in this way yields a whole set of parameterizations that all compute the same function, which we call a symmetry orbit.

These parameterizations represent the same function but lead to different training trajectories. The reason is that the gradients scale inversely to the weights: Multiplying $\mat{A}$ by $c$ multiplies its gradient by $c^{-1}$, and multiplying $\mat{B}$ by $c^{-1}$ multiplies its gradient by $c$. A gradient-descent step moves the enlarged matrix $c^2$ times less, relative to its size, than in the original model and its shrunken partner $c^2$ times more, so the two models no longer compute the same function after the step. Figure~\ref{fig:overview} illustrates this effect.

Concretely, symmetry orbits in a Transformer arise from attention query--key (Q--K) and value--output (V--O) pairs, ReLU up--down projections, and the linear value--output path of SwiGLU (Sec.~\ref{sec:symmetries}). This inverse gradient-scaling law holds for all of these pairs and follows from the chain rule~\citep[Theorem~3]{dinh2017sharp}. We use this law to determine how a rescaling distorts the optimizer's updates and what is needed for rescaled models to train identically.

Standard optimizers already include mechanisms such as adaptive gradient normalization, but these mechanisms still do not remove this dependence on the parameterization. Adam normalizes gradients using moving averages of gradients and squared gradients~\citep{kingma2015adam}, which cancels constant-factor scaling of the gradients: Both models receive the same numerical update (Figure~\ref{fig:overview}(2)). To keep the two models equivalent, however, the update of the enlarged matrix would have to be $c$ times larger and the update of its shrunken partner $c$ times smaller. The same update is too small for the enlarged matrix and too large for its shrunken partner. Transforming Adam's moment estimates along with the weights does not fix this mismatch, because both models still use the same learning rate (Sec.~\ref{sec:theory}). Appendix~\ref{app:related} discusses in detail how this work relates to and differs from prior studies of rescaling-aware optimization, weight balancing, and symmetry-induced learning dynamics~\citep{neyshabur2015path,stock2019equinormalization,kunin2021neural}.

To remove the scale imbalance between paired matrices without changing the function, we introduce Symmetry-Aware Gradient Geometry Stabilization (SAGGS), which periodically rescales each exact symmetry pair so that its two matrices have equal Frobenius norms (Sec.~\ref{sec:method}). To avoid disrupting optimizer state that has adapted to imbalanced scales, Soft SAGGS applies only a fraction of this correction at each rebalancing step. Both operations leave the current function unchanged and only rebalance the weights from which training continues. SAGGS also gives an exact restart procedure: Applied together with a matching transformation of the moment estimates, SAGGS moves any two checkpoints on the same symmetry orbit to the same balanced state, from which training follows the same trajectory (Corollary~\ref{cor:canonical_restart}). We evaluate both variants in deliberate rescaling tests, which rescale the pairs of layer $\ell$ by $c_\ell=e^{\lambda\ell}$ to impose pair-scale imbalance without changing the function, as well as in checkpoint-resume workflows and in training without such rescaling (Sec.~\ref{sec:experiments}). In our seven-seed comparisons, rebalancing restores near-baseline PPL after an imposed rescaling, including a checkpoint edit, and leaves PPL nearly unchanged when no imbalance is imposed.

Our contributions are fourfold.

\begin{itemize}
\item We identify exact symmetry orbits in Transformer architectures (Sec.~\ref{sec:symmetries}) and show their effect on training.

\item Our theoretical analysis proves that, under a shared learning rate, no standard optimizer, including gradient descent, AdamW, and Lion, keeps two equivalent models equivalent during training (Theorem~\ref{thm:equivariant_rules}). Even with correctly transformed moment estimates, AdamW does so only if its learning rate, $\epsilon$, and weight decay are also rescaled coordinate by coordinate (Theorem~\ref{thm:adam_equivariance} and Example~\ref{rem:adam_moment_covariance}). We further prove that canonicalizing both weights and moments at restart makes the training resumed from any two checkpoints on the same symmetry orbit identical (Corollary~\ref{cor:canonical_restart}).

\item We propose SAGGS, which periodically moves each exact symmetry pair to the unique point on its orbit where the two matrices have equal Frobenius norms, regardless of where on the orbit the pair starts (Proposition~\ref{cor:saggs_orbit}). Its damped variant, Soft SAGGS, moves each pair only part of the way there, shrinking the log ratio of the two norms by a constant factor at each rebalancing step (Sec.~\ref{sec:soft_rebalancing}).

\item Under deliberate rescaling, seven-seed paired comparisons show consistent PPL reductions from rebalancing, of $0.374\pm0.026$ for a 101M WikiText-103 model and of $0.780\pm0.040$ for a 31M modern-architecture replication, and also for training with Lion (Sec.~\ref{sec:lm_comparisons}). In a 123.8M checkpoint-resume workflow, an edit that barely changes validation loss at restart increases final PPL by $3.78\pm0.42$, and both rebalancing variants recover to within $0.02$ PPL of the control (Sec.~\ref{sec:checkpoint_recovery}).
\end{itemize}

\section{Symmetries and Gradients}
\label{sec:symmetries}

This section identifies which pairs of weight matrices in a Transformer can trade scale without changing the model's function, and how such a trade changes their gradients. SAGGS rebalances exactly these pairs (Sec.~\ref{sec:method}), and the optimizer analysis of Sec.~\ref{sec:theory} starts from this change in the gradients.

\subsection{Exact Reparameterization Symmetries}
\label{sec:symmetry_catalogue}

Consider an $L$-layer Transformer~\citep{vaswani2017attention}. At layer $\ell \in \{1, \ldots, L\}$, the self-attention uses projections $\mat{W}^Q_\ell$, $\mat{W}^K_\ell$, $\mat{W}^V_\ell$, and $\mat{W}^O_\ell$, and the MLP uses $\mat{W}^1_\ell$ and $\mat{W}^2_\ell$. Let $\loss(\param)$ denote the data loss evaluated at parameters $\param$. The equations omit biases.

Table~\ref{tab:symmetry_catalogue} records the scalar pair symmetries used in the optimizer analysis. The Q--K and V--O rows are scalar instances of the multi-head attention symmetries characterized by \citet{tran2025equivariant}; the ReLU up--down row uses standard positive homogeneity~\citep{dinh2017sharp,neyshabur2015path}. The Gaussian error linear unit (GELU)~\citep{hendrycks2016gaussian} is not positively homogeneous, so rescaling its up--down pair changes the function, and SAGGS excludes this pair. The SwiGLU MLP~\citep{shazeer2020glu} is defined as
\begin{equation}
  \mathrm{SwiGLU}_\ell(\mat{x}) \coloneqq \mat{W}^2_\ell [ \mathrm{Swish}(\mat{W}^{\mathrm{gate}}_\ell \mat{x}) \odot (\mat{W}^{\mathrm{up}}_\ell \mat{x}) ],
  \label{eq:swiglu}
\end{equation}
where $\odot$ denotes element-wise multiplication. For a fixed gate, the output is linear in the value path $\mat{W}^{\mathrm{up}}_\ell\mat{x}$, so rescaling $\mat{W}^{\mathrm{up}}_\ell$ by $c$ and $\mat{W}^2_\ell$ by $c^{-1}$ leaves the output unchanged; the gate $\mat{W}^{\mathrm{gate}}_\ell$ is not rescaled.

\begin{table}[t!]
\centering
\caption{Scalar pair rescalings $(\mat{A},\mat{B})\mapsto(c\mat{A},c^{-1}\mat{B})$, $c>0$. Rows marked \cmark~Yes preserve the forward pass exactly and are used by SAGGS.}
\label{tab:symmetry_catalogue}
\vspace{4pt}
\resizebox{\columnwidth}{!}{%
\begin{tabular}{@{}lccll@{}}
\toprule
\textbf{Sub-layer} & $\mat{A}$ & $\mat{B}$ & \textbf{Mechanism} & \textbf{Exact} \\
\midrule
\multirow[t]{2}{*}{Self-attention} & $\mat{W}^Q_\ell$ & $\mat{W}^K_\ell$ & Bilinear in logits & \cmark~Yes \\
 & $\mat{W}^V_\ell$ & $\mat{W}^O_\ell$ & Linear composition & \cmark~Yes \\
\midrule
MLP (ReLU) & $\mat{W}^1_\ell$ & $\mat{W}^2_\ell$ & Positive homogeneity & \cmark~Yes \\
MLP (GELU) & $\mat{W}^1_\ell$ & $\mat{W}^2_\ell$ & Non-homogeneous & \xmark~No \\
\midrule
SwiGLU & $\mat{W}^{\mathrm{up}}_\ell$ & $\mat{W}^2_\ell$ & Linear in value path & \cmark~Yes \\
\bottomrule
\end{tabular}%
}
\end{table}

The same pairs remain exact in the modern architecture that we use for the 31M replication and the checkpoint-resume experiments (Sec.~\ref{sec:experiments}). This architecture combines SwiGLU with grouped-query attention (GQA), root mean square layer normalization (RMSNorm), and rotary position embeddings (RoPE); under GQA, the Q--K and V--O rescalings use one factor per group (Appendix~\ref{app:gqa_detail}).

\subsection{Gradient Covariance Identity}
\label{sec:grad_law}

For an exact scalar symmetry pair, the standard chain-rule identity for reciprocal rescaling~\citep[Theorem~3]{dinh2017sharp} gives, at differentiable points,
\begin{equation}
  \grad_{\mat{A}'}\loss=c^{-1}\grad_{\mat{A}}\loss,\qquad \grad_{\mat{B}'}\loss=c\grad_{\mat{B}}\loss,
  \label{eq:grad_transform}
\end{equation}
where $(\mat{A}',\mat{B}')\coloneqq(c\mat{A},c^{-1}\mat{B})$ and each gradient of the same data loss is evaluated at its own parameterization. The two parameterizations compute the same function yet give the optimizer different gradients. In Sec.~\ref{sec:theory}, we show that, under a shared learning rate, no standard optimizer scales its updates with the weights, so two equivalent models stop computing the same function after an update. We also show that canonicalizing both weights and moments at restart makes two such models train identically.

Appendix~\ref{app:arch_variants} gives the matrix shapes, the single-head attention and MLP equations, the bias co-transforms, and the argument that layer normalization leaves the pairs exact; the same appendix also discusses cross-attention, mixture-of-experts, and state-space blocks, which our experiments do not use. Lemma~\ref{lem:invariance_grad_law} in Appendix~\ref{app:grad_law_proof} states the general diagonal-map form of \eqref{eq:grad_transform} used in the proofs.

\section{Equivariance and Restarts}
\label{sec:theory}

Two models on the same orbit compute the same function but give the optimizer different gradients (\eqref{eq:grad_transform}). This section asks whether an optimizer can compensate for this difference and keep the two models equivalent after every update. If the optimizer cannot, the position of a model on its orbit affects training, which motivates rebalancing (Sec.~\ref{sec:method}). We fix one balanced representative of each orbit (Sec.~\ref{sec:orbit_canonicalization}), derive what AdamW needs to keep two runs equivalent (Sec.~\ref{sec:optimizer_equivariance}), show that restarting both runs from the same balanced state makes the resumed runs identical (Sec.~\ref{sec:canonical_restart}), and prove that no standard optimizer keeps two runs equivalent under a shared learning rate (Sec.~\ref{sec:equivariant_rules}).

\subsection{Scalar Orbit Canonicalization}
\label{sec:orbit_canonicalization}

All points on an orbit compute the same function, so any one of these points can represent the orbit. We use the point at which the two matrices of each exact pair have equal Frobenius norms. The exact restart (Sec.~\ref{sec:canonical_restart}) and SAGGS (Sec.~\ref{sec:rebalancing}) both move a model to this point.

\begin{proposition}[Unique balanced representative of a scalar orbit]
\label{cor:saggs_orbit}
For an exact symmetry pair $(\mat{A}_p,\mat{B}_p)$ with positive Frobenius norms, let $\mat{A}_p^\star \coloneqq \mat{A}_p/c_p^\star$ and $\mat{B}_p^\star \coloneqq c_p^\star\mat{B}_p$ with $c_p^\star \coloneqq (\frobnorm{\mat{A}_p}/\frobnorm{\mat{B}_p})^{1/2}$. These matrices form the unique Frobenius-balanced point on the orbit, with $\frobnorm{\mat{A}_p^\star} = \frobnorm{\mat{B}_p^\star}$. The balancing map $(\mat{A}_p,\mat{B}_p)\mapsto(\mat{A}_p^\star,\mat{B}_p^\star)$ is independent of the starting orbit representative, idempotent, and forward-preserving.
\end{proposition}

Balancing sends every point of an orbit to the same point, so two checkpoints that differ only by an orbit move end up with identical weights after balancing.

\subsection{AdamW Trajectory Equivariance}
\label{sec:optimizer_equivariance}

The gradient normalization built into Adam cannot make parameter updates equivariant. With covariantly transformed moments, Adam's normalized direction cancels the gradient scale when $\epsilon_i=0$; a common positive $\epsilon_i$ generally prevents exact cancellation. Even when this cancellation is exact, two runs that start on the same orbit do not stay equivalent. Adam then applies the same update to both runs, whereas staying on the same orbit requires each parameter to move in proportion to its own scale, that is, $c$ times more for the enlarged block and $c$ times less for the shrunken block.

\paragraph{AdamW setup.}
Let $S\coloneqq\mathrm{diag}(s_i)$, $s_i>0$, act on disjoint exact symmetry pairs as $(\mat{A},\mat{B})\mapsto(c\mat{A},c^{-1}\mat{B})$ and as the identity elsewhere.  Every sampled data loss $\loss_t$ is assumed differentiable at the iterates and satisfies $\loss_t(S\param)=\loss_t(\param)$.  Consider AdamW, with common $(\beta_1,\beta_2)$ and bias-correction counters. Let $m_t$ and $v_t$ denote its moving averages of the data gradients and their elementwise squares, respectively. These averages estimate the first moment and second raw moment of the gradient; hats denote bias-corrected estimates. We refer to these estimates simply as moments. After updating the moments, the parameter update is
\begin{equation}
 \theta_{t+1,i}\coloneqq(1-\eta_{t,i}\omega_{t,i})\theta_{t,i} -\eta_{t,i}\frac{\hat m_{t+1,i}} {\sqrt{\hat v_{t+1,i}}+\epsilon_{t,i}}.
 \label{eq:adam_coordinate_update}
\end{equation}
Here $\eta_{t,i}>0$, $\omega_{t,i},\epsilon_{t,i}\geq0$, all denominators are positive, and the gradients entering the moments are the data gradients, without clipping~\citep{pascanu2013difficulty} or an additional parameter penalty. The following theorem gives the exact conditions under which AdamW, including its moments, keeps two runs on the same orbit.

\begin{theorem}[AdamW trajectory equivariance]
\label{thm:adam_equivariance}
Under the AdamW setup above and the state transformation $(\param',m',v')\coloneqq(S\param,S^{-1}m,S^{-2}v)$, the parameter-update formula commutes with $S$ for every state with positive denominators if and only if, at every step and coordinate,
\begin{equation}
 \eta'_{t,i}=s_i\eta_{t,i},\qquad \epsilon'_{t,i}=s_i^{-1}\epsilon_{t,i},\qquad \omega'_{t,i}=s_i^{-1}\omega_{t,i}.
 \label{eq:adam_equivariance_rule}
\end{equation}
Consequently, with these rules and a shared sequence of sampled losses, orbit-related initial states remain orbit-related at every step.
\end{theorem}

In words, when a coordinate is multiplied by $s_i$, its learning rate must be multiplied by $s_i$ as well, while its $\epsilon$ and weight decay must be divided by $s_i$. A learning rate shared by the two runs violates this rule for every nontrivial move $S\neq I$. The following example shows how large the resulting error can be.

\begin{example}[Moment covariance alone is insufficient]
\label{rem:adam_moment_covariance}
If the two runs keep the same hyperparameters, the trajectory equivariance of Theorem~\ref{thm:adam_equivariance} fails in general even when $\epsilon=0$ and $\omega=0$.  In particular, let $f(a,b)\coloneqq ab$ and $\loss(a,b)\coloneqq(ab)^2/2$.  One Adam step from zero moments, with common learning rate $\eta>0$, starting at $(1,1)$ and $(c,c^{-1})$, respectively, gives
\begin{equation}
 \abs{f(a'_1,b'_1)-f(a_1,b_1)} =\eta(c+c^{-1}-2).
 \label{eq:adam_functional_counterexample}
\end{equation}
The primed loss after that step is unbounded as $c\to\infty$, although both initial losses equal $1/2$ and their moments transform correctly.
\end{example}

Both runs move each weight by $\eta$, whereas equivalence would require the primed run to move $a'$ by $c\eta$ and $b'$ by $\eta/c$: The step is too small for the enlarged weight and too large for the shrunken one. The failure arises even without moment lag, clipping, weight decay, or an $\epsilon$ floor. With $c$ set by the depth schedule $c_\ell=e^{\lambda\ell}$ of our deliberate rescaling tests (Sec.~\ref{sec:experiments}), the discrepancy is $2\eta[\cosh(\lambda\ell)-1]$ and grows exponentially with the layer index~$\ell$.

\subsection{Exact Canonical Restart}
\label{sec:canonical_restart}

With shared hyperparameters, AdamW cannot keep two equivalent runs equivalent (Theorem~\ref{thm:adam_equivariance}). A restart avoids this problem: If two runs resume from the same numerical state, both runs follow the same trajectory. For the weights, balancing already provides this common state, because the balancing map sends every point of an orbit to the same point (Proposition~\ref{cor:saggs_orbit}); the map below extends this canonicalization to the moments.

\paragraph{Canonical state map.}
Let $S$ and the disjoint exact pairs be as in Theorem~\ref{thm:adam_equivariance}.  Assume each paired matrix has positive Frobenius norm.  Let $R_{\param}$ multiply each $\mat{A}$ block by $\rho^{-1/2}$ and each $\mat{B}$ block by $\rho^{1/2}$, where $\rho\coloneqq\frobnorm{\mat{A}}/\frobnorm{\mat{B}}$, and leave other coordinates unchanged.  Define the state canonicalizer
\begin{equation}
 \mathcal C(\param,m,v) \coloneqq(R_{\param}\param,R_{\param}^{-1}m,R_{\param}^{-2}v).
 \label{eq:state_canonicalizer}
\end{equation}

\begin{corollary}[Exact canonical restart]
\label{cor:canonical_restart}
For the state canonicalizer $\mathcal C$ and exact-pair map $S$ defined above, with positive norms on every pair,
\begin{equation}
 \mathcal C(S\param,S^{-1}m,S^{-2}v)=\mathcal C(\param,m,v).
 \label{eq:canonical_restart_identity}
\end{equation}
Thus canonicalizing both orbit-related checkpoints and their moments before the first resumed update, and then using common hyperparameters, counters, and randomness, gives identical subsequent training trajectories.  The same conclusion holds when both sets of moments are reset to zero before canonicalization.
\end{corollary}

In other words, if the weights and moments of both checkpoints are canonicalized before training resumes, the two runs are identical from the first resumed update onward. The restart needs no equivariant optimizer, because canonicalization removes the difference between the two runs before the optimizer sees either run. The checkpoint-resume experiments of Sec.~\ref{sec:checkpoint_recovery} test a practical version of this idea: These experiments rebalance an edited branch periodically and compare its final PPL with the final PPL of a control.

\subsection{Limits of Block-Local Updates}
\label{sec:equivariant_rules}

AdamW becomes trajectory-equivariant only after its hyperparameters are transformed coordinate by coordinate (Theorem~\ref{thm:adam_equivariance}). We now show that no standard first-order optimizer keeps two equivalent runs equivalent under a shared learning rate. The obstruction comes from block locality: Standard optimizers update each block using only the gradient history of that block. Write $\param\coloneqq(\{(\mat{A}_p,\mat{B}_p)\}_{p\in\mathcal P},\bm\xi)$ for the disjoint exact pairs together with the remaining parameters, and let $\mathcal G\coloneqq\{S_{\bm c}:\bm c\in(0,\infty)^{\mathcal P}\}$ be the group of orbit moves $S_{\bm c}\param\coloneqq(\{(c_p\mat{A}_p,c_p^{-1}\mat{B}_p)\}_p,\bm\xi)$. For the exact pairs in Table~\ref{tab:symmetry_catalogue}, every sampled loss satisfies $\loss_t\circ S=\loss_t$ for all $S\in\mathcal G$.

A first-order method is a sequence of maps $\mathcal U_t$ with $\param_{t+1}\coloneqq\param_t+\mathcal U_t(\param_{1:t},g_{1:t})$ and $g_\tau\coloneqq\grad\loss_\tau(\param_\tau)$; hyperparameters, step counters, and optimizer state with a fixed initialization are absorbed into $\mathcal U_t$.  The method is orbit-equivariant if
\begin{equation}
 \mathcal U_t(S\param_{1:t}, S^{-1}g_{1:t}) =S\mathcal U_t(\param_{1:t},g_{1:t}),
 \label{eq:orbit_equivariance}
\end{equation}
for all $S\in\mathcal G$ and all $t$, wherever both sides are defined. Here $S^{-1}g_{1:t}$ is exactly the gradient history produced at orbit-related iterates (Lemma~\ref{lem:invariance_grad_law}).  For a pair $p$, we write $x_A\coloneqq g^{A_p}_{1:t}$ and $x_B\coloneqq g^{B_p}_{1:t}$ for the gradient histories of its two blocks.

\begin{theorem}[Block-local obstruction to orbit equivariance]
\label{thm:equivariant_rules}
Fix a pair $p$ and suppose that the increment of $\mat{A}_p$ has the form
\begin{equation}
 \Delta\mat{A}_{p,t} \coloneqq-\eta_t\Psi_t(x_A)-\eta_t\omega_t\mat{A}_{p,t},
 \label{eq:block_local_rule}
\end{equation}
with $\eta_t>0$, $\omega_t\ge0$, and $\Psi_t$ a function of the gradient history of the block $\mat{A}_p$ alone.  If $\mathcal U$ is orbit-equivariant, then
\begin{equation}
 \Psi_t(c^{-1}x)=c\Psi_t(x),
 \label{eq:degree_minus_one}
\end{equation}
for all $c>0$ and all histories $x$. Consequently $\Psi_t\equiv0$ whenever $\Psi_t$ is positively homogeneous of some degree $k\neq-1$, or is bounded on a neighborhood of the zero history.  The same holds for $\mat{B}_p$.
\end{theorem}

By \eqref{eq:degree_minus_one}, an equivariant block-local rule must turn gradients that are $c$ times smaller into an update that is $c$ times larger, since the block itself is $c$ times larger. Standard optimizers never meet this requirement: When the gradients get smaller, their updates get smaller or stay the same. Gradient descent and heavy-ball momentum have degree~$1$ and scale their updates with the gradients. Degree-zero rules ignore the gradient scale: AdaGrad, RMSProp, Adam, and AdamW with a zero stabilizer $\epsilon=0$; the sign-based rules signSGD and Lion; and the zero-stabilizer limit of Shampoo. Adam and AdamW with $\epsilon>0$ keep their updates bounded as the gradients vanish. None of these optimizers is orbit-equivariant under a shared learning rate. Block locality and the homogeneity or boundedness assumption are both needed: A block-local rule of degree~$-1$, or a rule that couples the two blocks of a pair, can be equivariant; Appendix~\ref{app:theory_details_moved} gives these examples and the precise conditions for each optimizer above.

\paragraph{Relative-update scaling.}
Theorem~\ref{thm:equivariant_rules} shows that equivalence breaks; the next result measures by how much, in terms of how far each block moves relative to its own size. The gradient covariance in~\eqref{eq:grad_transform}, combined with homogeneity of the block-local direction, gives the following scaling law.

\begin{corollary}[Relative-update scaling for homogeneous rules]
\label{cor:half_compensation}
Let both blocks of pair $p$ follow~\eqref{eq:block_local_rule} with $\omega_t=0$ and with $\Psi_t$ positively homogeneous of degree $k$. Assume both current block norms are positive. For the orbit-related histories $(\param_{1:t},g_{1:t})$ and $(S_c\param_{1:t},S_c^{-1}g_{1:t})$, where $S_c$ acts on pair $p$ with factor $c$, the increments satisfy $\Delta\mat{A}'=c^{-k}\Delta\mat{A}$ and $\Delta\mat{B}'=c^{k}\Delta\mat{B}$, so that
\begin{equation}
 \begin{aligned}
 \frac{\frobnorm{\Delta\mat{A}'}}{\frobnorm{\mat{A}'}} &=c^{-(1+k)}\frac{\frobnorm{\Delta\mat{A}}}{\frobnorm{\mat{A}}},\\
 \frac{\frobnorm{\Delta\mat{B}'}}{\frobnorm{\mat{B}'}} &=c^{1+k}\frac{\frobnorm{\Delta\mat{B}}}{\frobnorm{\mat{B}}}.
 \end{aligned}
 \label{eq:relative_update_tilt}
\end{equation}
\end{corollary}

For gradient descent with $k=1$, the relative update of the enlarged block shrinks by $c^2$ and the relative update of the shrunken block grows by $c^2$; degree-zero rules such as Lion or Adam with $\epsilon=0$ reduce both factors to $c$. Under the depth schedule $c_\ell=e^{\lambda\ell}$, the B-to-A log-ratio of relative updates shifts by $2(1+k)\lambda\ell$: $4\lambda\ell$ for gradient descent and $2\lambda\ell$ for degree-zero rules. Gradient normalization at best halves this shift, from $4\lambda\ell$ to $2\lambda\ell$, and does not remove the shift. SAGGS instead removes this shift at each rebalance: Measured from the balanced point of Proposition~\ref{cor:saggs_orbit}, each rebalance resets $c$ to $1$ for every pair (Sec.~\ref{sec:method}).

Proofs for this section are in Appendix~\ref{app:proofs}.

\section{SAGGS Rebalancing}
\label{sec:method}

No standard optimizer keeps two equivalent models equivalent under a shared learning rate (Theorem~\ref{thm:equivariant_rules}); thus, the position of a model on its orbit affects training. SAGGS leaves the optimizer unchanged and instead periodically moves the model along its orbit to the balanced point, where the two matrices of each exact pair have equal Frobenius norms (Proposition~\ref{cor:saggs_orbit}). The current function stays the same, and training continues from balanced pairs. Hard SAGGS makes this move at once (Sec.~\ref{sec:rebalancing}); its damped variant, Soft SAGGS, makes only part of the move at each rebalance so as not to disrupt optimizer state that has adapted to the imbalance (Sec.~\ref{sec:soft_rebalancing}).

\subsection{Hard SAGGS}
\label{sec:rebalancing}

Hard SAGGS applies the balancing map of Proposition~\ref{cor:saggs_orbit} to every exact pair once every $T_{\mathrm{bal}}$ optimization steps (Algorithm~\ref{alg:rebalancing}). Each application shrinks the larger matrix of a pair and enlarges the smaller one by the same factor, so that both norms become the geometric mean of the two; in Figure~\ref{fig:overview}(4), the rectangle becomes a square of the same area. Because each application is an exact orbit move, the forward map and the current data loss are unchanged at each rebalance.

\begin{algorithm}[t!]
\caption{Symmetry-Aware Rebalancing}
\label{alg:rebalancing}
\begin{algorithmic}[1]
\REQUIRE Parameters $\param$, interval $T_{\mathrm{bal}}$, disjoint exact symmetry pairs $(\mat{A}_p,\mat{B}_p)$, $p\in\mathcal{P}$
\FOR{each optimization step $t$}
  \STATE Perform standard optimizer update on $\param$
  \IF{$t \bmod T_{\mathrm{bal}} = 0$}
    \FOR{each pair $p \in \mathcal{P}$}
      \IF{$\frobnorm{\mat{A}_p}=0$ or $\frobnorm{\mat{B}_p}=0$}
        \STATE Skip this pair, since no positive norm ratio is defined
      \ELSE
      \STATE $\rho_p \leftarrow \frobnorm{\mat{A}_p} / \frobnorm{\mat{B}_p}$
      \STATE $\mat{A}_p \leftarrow \rho_p^{-1/2}\mat{A}_p$
      \STATE $\mat{B}_p \leftarrow \rho_p^{1/2}\mat{B}_p$
      \ENDIF
    \ENDFOR
  \ENDIF
\ENDFOR
\end{algorithmic}
\end{algorithm}

\subsection{Soft SAGGS}
\label{sec:soft_rebalancing}
Hard SAGGS moves each pair to its balanced point in a single step but leaves the optimizer state, which was built from gradients at the old scales, unchanged. Lion, for example, takes the sign of a combination of the current gradient and a momentum buffer, an exponential moving average (EMA) of past gradients~\citep{chen2023symbolic}. After a hard rebalance, the gradient of the shrunken block grows and the gradient of the enlarged block shrinks (\eqref{eq:grad_transform}), while the buffer keeps the old scales, so the sign of the combination, and hence the update, can change abruptly. In a Lion ablation without imposed rescaling, hard SAGGS was less consistent across seeds, whereas state-covariant SAGGS, which also rescales the buffer by the same factors as the gradients, removed most of this variation (Appendix~\ref{app:exp64_moved}).

To avoid abrupt changes in the update without modifying the optimizer state, Soft SAGGS makes each correction smaller: At each rebalance, Soft SAGGS moves each pair only part of the way to its balanced point, so the gradient scales change in several smaller steps rather than in one jump. Concretely, Soft SAGGS keeps the schedule and the norm ratio $\rho_p$ of Algorithm~\ref{alg:rebalancing} but replaces its full correction with the damped rebalancing map
\begin{equation}
  \begin{gathered}
  \mat{A}_p \leftarrow c^{\mathrm{eff}}_p(\alpha)^{-1}\mat{A}_p,\\
  \mat{B}_p \leftarrow c^{\mathrm{eff}}_p(\alpha)\mat{B}_p,
  \end{gathered}
  \label{eq:soft_saggs}
\end{equation}
where $c^{\mathrm{eff}}_p(\alpha)\coloneqq\rho_p^{\alpha/2}$ and $\alpha \in (0, 1]$ controls how much of the full correction is applied. Forward equivalence is preserved exactly for any $\alpha$, because this map is an exact orbit move with factor $1/c^{\mathrm{eff}}_p$ on a pair from Table~\ref{tab:symmetry_catalogue}. Setting $\alpha = 1$ recovers hard SAGGS, and as $\alpha \to 0^+$ the map approaches the identity. We refer to \eqref{eq:soft_saggs} with $\alpha < 1$ as Soft SAGGS, and we adopt $\alpha = 0.5$ as the default whenever a single value is required.

\paragraph{Soft and hard SAGGS.}
Each Soft SAGGS rebalance multiplies the log-norm ratio $\log\rho_p$ of a pair by $1-\alpha$, so repeated rebalances shrink the imbalance geometrically instead of removing the imbalance at once. For $\alpha<1$, however, the rebalanced weights still depend on where on the orbit the pair starts, and applying the map again moves the weights further; only hard SAGGS sends every point of an orbit to the same point, as the exact restart of Corollary~\ref{cor:canonical_restart} requires.

Appendix~\ref{app:soft_contraction_moved} derives these properties, Appendix~\ref{app:statecov_detail} discusses optimizer-state handling and defines state-covariant SAGGS, Appendix~\ref{app:notation_setup_moved} gives the bias co-transforms for affine projections, Appendix~\ref{app:implementation} provides a reference implementation of hard and Soft SAGGS, and Appendix~\ref{app:practical_full} covers computational cost, the interaction with weight decay, and numerical guards.


\section{Experiments}
\label{sec:experiments}

All points on an orbit compute the same function, yet the analysis of Sec.~\ref{sec:theory} predicts that the position of a model on its orbit affects training. The experiments below measure how much an imbalanced position degrades language-model training and how much of this degradation rebalancing recovers. We create such a position with an orbit edit, a deliberate rescaling of exact pairs, at initialization (Sec.~\ref{sec:lm_comparisons}) and in a checkpoint before training resumes (Sec.~\ref{sec:checkpoint_recovery}), and we also train without an edit to check whether rebalancing costs anything when no imbalance is imposed. Case studies on released checkpoints show imbalance arising in ordinary training and test exact canonical restart (Appendix~\ref{sec:qualitative_cases}).

\subsection{Experimental Setup and Statistics}
\label{sec:stat_methodology}
\label{sec:methodology}

Unless otherwise stated, we train a GPT-2-style decoder-only Transformer with pre-layer normalization (Pre-LN) on WikiText-103 using AdamW~\citep{loshchilov2019decoupled} with weight decay and gradient clipping. The primary 101M model uses ReLU. The 31M replication and the checkpoint experiments of Sec.~\ref{sec:checkpoint_recovery} use a modern architecture with RMSNorm, RoPE, SwiGLU, and GQA, together with T5 SentencePiece tokenization.

The orbit edit rescales exact pairs of layer $\ell$ by $c_\ell=e^{\lambda\ell}$, which leaves the function unchanged but imbalances deeper pairs more strongly. We compare training without an edit, at $\lambda=0$, with training after a deliberate edit at $\lambda=0.2$. The experiments index layers as $\ell=0,\ldots,L-1$, so the largest scale is $e^{\lambda(L-1)}$, about $100$ for a 24-layer model at $\lambda=0.2$. Vanilla uses the same optimizer and hyperparameters without rebalancing.

We report the mean paired PPL difference $\bar\Delta$ over seven paired seeds. Unless stated otherwise, a difference is Vanilla minus the method, so a positive value favors the method, and $\pm$ denotes a sample standard deviation (SD). The one-sided sign-flip $p$-value $p_{\mathrm{sf}}^{+}$ takes its smallest possible value, $0.0078$, exactly when all seven differences are positive.

\subsection{Paired Language-Model Comparisons}
\label{sec:lm_comparisons}
\label{sec:exp45_multiseed}
\label{sec:exp65_core_n7}
\label{sec:exp_lion}
\label{sec:exp66_lion_soft_n7}

\subsubsection{101M WikiText-103 Comparison}

In our main comparison, we train the 101M ReLU Transformer with $L=24$ for 30K AdamW steps, starting either without an edit or after a Q--K edit at $\lambda=0.2$. Hard SAGGS rebalances all exact pairs every 100 steps.

\begin{table}[t!]
\centering
\caption{101M WikiText-103 comparison with $n=7$. Method columns report mean PPL; $\bar\Delta$ reports the mean $\pm$ sample SD of paired Vanilla-minus-SAGGS differences. $d$ is paired Cohen's $d$ and $p_{\mathrm{sf}}^{+}$ is the one-sided sign-flip $p$-value.}
\label{tab:exp65_core_n7}
\vspace{4pt}
\resizebox{\columnwidth}{!}{%
\begin{tabular}{@{}lccccc@{}}
\toprule
$\lambda$ & Vanilla & SAGGS & $\bar\Delta$ & $d$ & $p_{\mathrm{sf}}^{+}$ \\
\midrule
$0.0$ & $22.881$ & $22.906$ & $-0.025\pm0.034$ & $-0.75$ & $0.9453$ \\
$0.2$ & $23.277$ & $22.904$ & $+0.374\pm0.026$ & $14.21$ & $0.0078$ \\
\bottomrule
\end{tabular}%
}
\end{table}

At $\lambda=0.2$, SAGGS reduces PPL by $0.374\pm0.026$ across seven paired seeds, with $p_{\mathrm{sf}}^{+}=0.0078$ (Table~\ref{tab:exp65_core_n7}). At $\lambda=0$, the difference is $-0.025\pm0.034$ PPL. With SAGGS, the edit no longer matters: PPL is $22.904$ with the edit and $22.906$ without the edit. Without an edit, Vanilla ends with a Q--K log-norm ratio of only about $0.15$, compared with about $3$ after the edit, and removing this small imbalance barely changes PPL.

\subsubsection{Architecture and Tokenizer Replication}
To test whether these results depend on the GPT-2 architecture or the GPT-2 tokenizer, we repeat the comparison on a 31M model with the modern architecture and T5 SentencePiece tokenization. Over seven paired seeds, Soft SAGGS reduces PPL by $0.780\pm0.040$ at $\lambda=0.2$, with $p_{\mathrm{sf}}^{+}=0.0078$, while the difference without an edit is $+0.019\pm0.024$ PPL (Table~\ref{tab:exp63_modern_n7_cells}). As in the 101M comparison, rebalancing helps after the edit and barely changes PPL without one, so its benefit is not specific to the GPT-2 architecture or the GPT-2 tokenizer.

\subsubsection{Lion Comparisons}
Lion~\citep{chen2023symbolic} uses only the sign of each update coordinate and therefore ignores the gradient scale, but, like Adam, Lion is still not orbit-equivariant (Theorem~\ref{thm:equivariant_rules}). We train the 101M ReLU model for 15K Lion steps with an all-pairs edit and seven paired seeds. We rebalance with Soft SAGGS, using $\alpha=0.5$ every 100 steps, because a hard rebalance can abruptly change Lion's updates (Sec.~\ref{sec:soft_rebalancing}). At $\lambda=0.2$, all seven seeds favor Soft SAGGS: $\bar\Delta=+34.465\pm25.170$ PPL and $p_{\mathrm{sf}}^{+}=0.0078$ (Table~\ref{tab:exp66_lion_soft_n7_cells}). At $\lambda=0$, the difference is $-0.055\pm0.232$ PPL. Without rebalancing, the edit more than doubles the mean PPL of Lion, from $24.15$ to $58.32$, whereas the edited runs with Soft SAGGS reach $23.86$.

\subsection{Checkpoint Recovery and Diagnostics}
\label{sec:checkpoint_recovery}
\label{sec:exp68_checkpoint_surgery}
\label{sec:exp70_state_preserved}

Training often resumes from a checkpoint with a fresh optimizer, for example when the checkpoint is imported without its optimizer state. To measure how much an imbalanced checkpoint degrades training in this setting, we train the 123.8M model for 10K steps with each of seven seeds, reset its optimizer, and clone the checkpoint into four 5K-step continuations: control, edited, edited with SAGGS, and edited with Soft SAGGS. The three edited branches receive the same exact all-pairs edit at $\lambda=0.2$, and SAGGS and Soft SAGGS rebalance at the restart and every 50 steps thereafter.

\begin{table}[t!]
\centering
\caption{123.8M checkpoint-surgery resume with $n=7$. Values are final PPL, reported as mean $\pm$ sample SD across seeds. Paired contrasts appear in Table~\ref{tab:exp68b_checkpoint_surgery_paired_moved}.}
\label{tab:exp68b_checkpoint_surgery_cells}
\small
\begin{tabular}{@{}lc@{}}
\toprule
\textbf{Resume condition} & \textbf{Final PPL} \\
\midrule
Control resume & $20.84\pm0.25$ \\
Surgery resume, no SAGGS & $24.62\pm0.58$ \\
Surgery + SAGGS & $20.85\pm0.25$ \\
Surgery + Soft SAGGS & $20.85\pm0.25$ \\
\bottomrule
\end{tabular}
\end{table}

At restart, the edited checkpoints are indistinguishable from the original by their loss: The edit and any first rebalance change the validation loss by less than $7.4\times10^{-5}$. Nevertheless, the edited run without rebalancing finishes $3.775\pm0.416$ PPL worse than the control, with $p_{\mathrm{sf}}^{+}=0.0078$ (Table~\ref{tab:exp68b_checkpoint_surgery_paired_moved}). Hard and Soft SAGGS finish within $0.02$ PPL of the control (Table~\ref{tab:exp68b_checkpoint_surgery_cells}), so rebalancing recovers almost all of the degradation caused by the edit.

Instead of being reset, the AdamW moments can be kept and transformed along with the weights. On the 31M model, keeping and transforming the moments leaves a nonzero gap of $0.058\pm0.017$ PPL relative to a control that also keeps its optimizer state (Table~\ref{tab:exp70_state_preserved_cells}), consistent with Theorem~\ref{thm:adam_equivariance}: Under a shared learning rate, transformed moments alone do not keep the two runs equivalent.

Case studies on public checkpoints (Appendix~\ref{sec:qualitative_cases}) show that pair imbalance also arises in ordinary training: In Pythia 410M, the query and key norms drift apart during pre-training by a factor of up to about $6$ in deep layers. The case studies also test canonical restart on GPT-2 774M, fine-tuned from the released checkpoint and from an exact Q--K edit of this checkpoint: Canonicalizing both checkpoints at restart makes the two runs identical, whereas without canonicalization the edited run predicts a different top-1 token than the canonicalized runs at $7.3\%$ of held-out positions after 200 updates.

Appendix~\ref{app:exp_details} gives the training configurations, hardware, and seed protocol, and Appendix~\ref{app:moved_details} defines the paired statistics, states their test assumptions, and separates primary from supporting comparisons. Appendix~\ref{app:moved_round2} gives the full protocols, tables, and paired contrasts for Secs.~\ref{sec:lm_comparisons} and~\ref{sec:checkpoint_recovery}, together with optimizer-state diagnostics for reset and stale moments. Appendix~\ref{app:additional_experiments} verifies the predicted gradient-norm scaling, measures natural drift, and reports stress tests and replications across corpora, architectures, and model sizes; some of these experiments use SAGGS+P, which adds to SAGGS an auxiliary penalty on pair imbalance (Appendix~\ref{par:saggs_hyperparameters}).

\section{Conclusion}
\label{sec:conclusion}

Even two Transformers that represent the same function can train differently. Our analysis characterizes exact AdamW equivariance, identifies the obstruction for standard block-local updates, and establishes exact restart equivalence through canonicalization of weights and moments. SAGGS and Soft SAGGS provide simple function-preserving corrections for pair-scale imbalance. The normalization layers of a Transformer, weight decay, and gradient clipping do not restore pair balance (Appendix~\ref{app:discussion_moved}). A rebalance rescales each pair only as much as the pair is imbalanced, so rebalancing can be kept on throughout training. Across paired language-model experiments and checkpoint-resume workflows, these corrections recover the performance lost to this imbalance, in particular after an exact orbit edit. These findings show that the pair-scale imbalance introduced by a parameterization strongly affects training, whether training starts from initialization or resumes from a checkpoint.

\clearpage
\section*{AI use statement}
Generative AI tools were used as assistive tools for writing and editing, mathematical proofreading and consistency checking, and code development and debugging. Any AI-assisted outputs incorporated into this work were reviewed and validated by the authors as appropriate. The authors take full responsibility for the final content of this work, including any text, claims, or artifacts produced with the assistance of generative AI.

\bibliographystyle{apalike}
\bibliography{lrs}

@inproceedings{chen2023symbolic,
  author       = {Xiangning Chen and
                  Chen Liang and
                  Da Huang and
                  Esteban Real and
                  Kaiyuan Wang and
                  Hieu Pham and
                  Xuanyi Dong and
                  Thang Luong and
                  Cho{-}Jui Hsieh and
                  Yifeng Lu and
                  Quoc V. Le},
  title        = {{Symbolic Discovery of Optimization Algorithms}},
  booktitle    = {NeurIPS},
  year         = {2023}
}

@inproceedings{dinh2017sharp,
  author       = {Laurent Dinh and
                  Razvan Pascanu and
                  Samy Bengio and
                  Yoshua Bengio},
  title        = {{Sharp Minima Can Generalize For Deep Nets}},
  booktitle    = {{ICML}},
  volume       = {70},
  pages        = {1019--1028},
  year         = {2017}
}

@inproceedings{du2018algorithmic,
  author       = {Simon S. Du and
                  Wei Hu and
                  Jason D. Lee},
  title        = {{Algorithmic Regularization in Learning Deep Homogeneous Models: Layers are Automatically Balanced}},
  booktitle    = {NeurIPS},
  pages        = {382--393},
  year         = {2018}
}

@inproceedings{glorot2010understanding,
  author       = {Xavier Glorot and
                  Yoshua Bengio},
  title        = {{Understanding the difficulty of training deep feedforward neural networks}},
  booktitle    = {{AISTATS}},
  volume       = {9},
  pages        = {249--256},
  year         = {2010}
}

@inproceedings{gupta2018shampoo,
  author       = {Vineet Gupta and
                  Tomer Koren and
                  Yoram Singer},
  title        = {{Shampoo: Preconditioned Stochastic Tensor Optimization}},
  booktitle    = {{ICML}},
  volume       = {80},
  pages        = {1837--1845},
  year         = {2018}
}

@inproceedings{he2015delving,
  author       = {Kaiming He and
                  Xiangyu Zhang and
                  Shaoqing Ren and
                  Jian Sun},
  title        = {{Delving Deep into Rectifiers: Surpassing Human-Level Performance on ImageNet Classification}},
  booktitle    = {{ICCV}},
  pages        = {1026--1034},
  year         = {2015}
}

@article{hendrycks2016gaussian,
  author       = {Dan Hendrycks and
                  Kevin Gimpel},
  title        = {{Bridging Nonlinearities and Stochastic Regularizers with Gaussian Error Linear Units}},
  journal      = {CoRR},
  volume       = {abs/1606.08415},
  year         = {2016}
}

@inproceedings{huang2020improving,
  author       = {Xiao Shi Huang and
                  Felipe P{\'{e}}rez and
                  Jimmy Ba and
                  Maksims Volkovs},
  title        = {{Improving Transformer Optimization Through Better Initialization}},
  booktitle    = {{ICML}},
  volume       = {119},
  pages        = {4475--4483},
  year         = {2020}
}

@inproceedings{kingma2015adam,
  author       = {Diederik P. Kingma and
                  Jimmy Ba},
  title        = {{Adam: {A} Method for Stochastic Optimization}},
  booktitle    = {{ICLR}},
  year         = {2015}
}

@inproceedings{kunin2021neural,
  author       = {Daniel Kunin and
                  Javier Sagastuy{-}Bre{\~{n}}a and
                  Surya Ganguli and
                  Daniel L. K. Yamins and
                  Hidenori Tanaka},
  title        = {{Neural Mechanics: Symmetry and Broken Conservation Laws in Deep Learning Dynamics}},
  booktitle    = {{ICLR}},
  year         = {2021}
}

@inproceedings{loshchilov2019decoupled,
  author       = {Ilya Loshchilov and
                  Frank Hutter},
  title        = {{Decoupled Weight Decay Regularization}},
  booktitle    = {{ICLR}},
  year         = {2019}
}

@inproceedings{merity2017pointer,
  author       = {Stephen Merity and
                  Caiming Xiong and
                  James Bradbury and
                  Richard Socher},
  title        = {{Pointer Sentinel Mixture Models}},
  booktitle    = {{ICLR}},
  year         = {2017}
}

@inproceedings{neyshabur2015path,
  author       = {Behnam Neyshabur and
                  Ruslan Salakhutdinov and
                  Nathan Srebro},
  title        = {{Path-SGD: Path-Normalized Optimization in Deep Neural Networks}},
  booktitle    = {{NIPS}},
  pages        = {2422--2430},
  year         = {2015}
}

@inproceedings{salimans2016weight,
  author       = {Tim Salimans and
                  Diederik P. Kingma},
  title        = {{Weight Normalization: {A} Simple Reparameterization to Accelerate Training of Deep Neural Networks}},
  booktitle    = {{NIPS}},
  pages        = {901},
  year         = {2016}
}

@article{shazeer2020glu,
  author       = {Noam Shazeer},
  title        = {{{GLU} Variants Improve Transformer}},
  journal      = {CoRR},
  volume       = {abs/2002.05202},
  year         = {2020}
}

@inproceedings{tanaka2021noether,
  author       = {Hidenori Tanaka and
                  Daniel Kunin},
  title        = {{Noether's Learning Dynamics: Role of Symmetry Breaking in Neural Networks}},
  booktitle    = {NeurIPS},
  pages        = {25646--25660},
  year         = {2021}
}

@inproceedings{vaswani2017attention,
  author       = {Ashish Vaswani and
                  Noam Shazeer and
                  Niki Parmar and
                  Jakob Uszkoreit and
                  Llion Jones and
                  Aidan N. Gomez and
                  Lukasz Kaiser and
                  Illia Polosukhin},
  title        = {{Attention is All you Need}},
  booktitle    = {{NIPS}},
  pages        = {5998--6008},
  year         = {2017}
}

@inproceedings{xiong2020layer,
  author       = {Ruibin Xiong and
                  Yunchang Yang and
                  Di He and
                  Kai Zheng and
                  Shuxin Zheng and
                  Chen Xing and
                  Huishuai Zhang and
                  Yanyan Lan and
                  Liwei Wang and
                  Tie{-}Yan Liu},
  title        = {{On Layer Normalization in the Transformer Architecture}},
  booktitle    = {{ICML}},
  volume       = {119},
  pages        = {10524--10533},
  year         = {2020}
}

@inproceedings{yang2022tensor,
  author       = {Ge Yang and
                  Edward J. Hu and
                  Igor Babuschkin and
                  Szymon Sidor and
                  Xiaodong Liu and
                  David Farhi and
                  Nick Ryder and
                  Jakub Pachocki and
                  Weizhu Chen and
                  Jianfeng Gao},
  title        = {{Tuning Large Neural Networks via Zero-Shot Hyperparameter Transfer}},
  booktitle    = {NeurIPS},
  pages        = {17084--17097},
  year         = {2021}
}

@inproceedings{zhai2023stabilizing,
  author       = {Shuangfei Zhai and
                  Tatiana Likhomanenko and
                  Etai Littwin and
                  Dan Busbridge and
                  Jason Ramapuram and
                  Yizhe Zhang and
                  Jiatao Gu and
                  Joshua M. Susskind},
  title        = {{Stabilizing Transformer Training by Preventing Attention Entropy Collapse}},
  booktitle    = {{ICML}},
  volume       = {202},
  pages        = {40770--40803},
  year         = {2023}
}

@inproceedings{ainsworth2023gitrebasin,
  author       = {Samuel K. Ainsworth and
                  Jonathan Hayase and
                  Siddhartha S. Srinivasa},
  title        = {{Git Re-Basin: Merging Models modulo Permutation Symmetries}},
  booktitle    = {{ICLR}},
  year         = {2023}
}

@inproceedings{ainslie2023gqa,
  author       = {Joshua Ainslie and
                  James Lee{-}Thorp and
                  Michiel de Jong and
                  Yury Zemlyanskiy and
                  Federico Lebr{\'{o}}n and
                  Sumit Sanghai},
  title        = {{{GQA:} Training Generalized Multi-Query Transformer Models from Multi-Head Checkpoints}},
  booktitle    = {{EMNLP}},
  pages        = {4895--4901},
  year         = {2023}
}

@article{gu2023mamba,
  author       = {Albert Gu and
                  Tri Dao},
  title        = {{Mamba: Linear-Time Sequence Modeling with Selective State Spaces}},
  journal      = {CoRR},
  volume       = {abs/2312.00752},
  year         = {2023}
}

@inproceedings{shazeer2017outrageously,
  author       = {Noam Shazeer and
                  Azalia Mirhoseini and
                  Krzysztof Maziarz and
                  Andy Davis and
                  Quoc V. Le and
                  Geoffrey E. Hinton and
                  Jeff Dean},
  title        = {{Outrageously Large Neural Networks: The Sparsely-Gated Mixture-of-Experts Layer}},
  booktitle    = {{ICLR}},
  year         = {2017}
}

@article{fedus2022switch,
  author       = {William Fedus and
                  Barret Zoph and
                  Noam Shazeer},
  title        = {{Switch Transformers: Scaling to Trillion Parameter Models with Simple and Efficient Sparsity}},
  journal      = {J. Mach. Learn. Res.},
  volume       = {23},
  pages        = {120:1--120:39},
  year         = {2022}
}

@inproceedings{tran2025equivariant,
  author       = {Hoang V. Tran and
                  Thieu Vo and
                  An Nguyen The and
                  Tho Tran Huu and
                  Minh{-}Khoi Nguyen{-}Nhat and
                  Thanh Tran and
                  Duy{-}Tung Pham and
                  Tan Minh Nguyen},
  title        = {{Equivariant Neural Functional Networks for Transformers}},
  booktitle    = {{ICLR}},
  year         = {2025}
}

@article{su2024roformer,
  author       = {Jianlin Su and
                  Murtadha H. M. Ahmed and
                  Yu Lu and
                  Shengfeng Pan and
                  Wen Bo and
                  Yunfeng Liu},
  title        = {{RoFormer: Enhanced transformer with Rotary Position Embedding}},
  journal      = {Neurocomputing},
  volume       = {568},
  pages        = {127063},
  year         = {2024}
}

@inproceedings{zhang2019rmsnorm,
  author       = {Biao Zhang and
                  Rico Sennrich},
  title        = {{Root Mean Square Layer Normalization}},
  booktitle    = {NeurIPS},
  pages        = {12360--12371},
  year         = {2019}
}

@inproceedings{zhao2022teleportation,
  author       = {Bo Zhao and
                  Nima Dehmamy and
                  Robin Walters and
                  Rose Yu},
  title        = {{Symmetry Teleportation for Accelerated Optimization}},
  booktitle    = {NeurIPS},
  year         = {2022}
}

@article{touvron2023llama,
  author       = {Hugo Touvron and
                  Thibaut Lavril and
                  Gautier Izacard and
                  Xavier Martinet and
                  Marie{-}Anne Lachaux and
                  Timoth{\'{e}}e Lacroix and
                  Baptiste Rozi{\`{e}}re and
                  Naman Goyal and
                  Eric Hambro and
                  Faisal Azhar and
                  Aur{\'{e}}lien Rodriguez and
                  Armand Joulin and
                  Edouard Grave and
                  Guillaume Lample},
  title        = {{LLaMA: Open and Efficient Foundation Language Models}},
  journal      = {CoRR},
  volume       = {abs/2302.13971},
  year         = {2023}
}

@inproceedings{bernstein2018signsgd,
  author       = {Jeremy Bernstein and
                  Yu{-}Xiang Wang and
                  Kamyar Azizzadenesheli and
                  Animashree Anandkumar},
  title        = {{{SIGNSGD:} Compressed Optimisation for Non-Convex Problems}},
  booktitle    = {{ICML}},
  volume       = {80},
  pages        = {559--568},
  year         = {2018}
}

@inproceedings{entezari2022role,
  author       = {Rahim Entezari and
                  Hanie Sedghi and
                  Olga Saukh and
                  Behnam Neyshabur},
  title        = {{The Role of Permutation Invariance in Linear Mode Connectivity of Neural Networks}},
  booktitle    = {{ICLR}},
  year         = {2022}
}

@inproceedings{henry2020query,
  author       = {Alex Henry and
                  Prudhvi Raj Dachapally and
                  Shubham Shantaram Pawar and
                  Yuxuan Chen},
  title        = {{Query-Key Normalization for Transformers}},
  booktitle    = {{EMNLP} (Findings)},
  volume       = {{EMNLP} 2020},
  pages        = {4246--4253},
  year         = {2020}
}

@inproceedings{liu2020understanding,
  author       = {Liyuan Liu and
                  Xiaodong Liu and
                  Jianfeng Gao and
                  Weizhu Chen and
                  Jiawei Han},
  title        = {{Understanding the Difficulty of Training Transformers}},
  booktitle    = {{EMNLP} {(1)}},
  pages        = {5747--5763},
  year         = {2020}
}

@inproceedings{pascanu2013difficulty,
  author       = {Razvan Pascanu and
                  Tom{\'{a}}s Mikolov and
                  Yoshua Bengio},
  title        = {{On the difficulty of training recurrent neural networks}},
  booktitle    = {{ICML} {(3)}},
  volume       = {28},
  pages        = {1310--1318},
  year         = {2013}
}

@inproceedings{stock2019equinormalization,
  author       = {Pierre Stock and
                  Benjamin Graham and
                  R{\'{e}}mi Gribonval and
                  Herv{\'{e}} J{\'{e}}gou},
  title        = {{Equi-normalization of Neural Networks}},
  booktitle    = {{ICLR}},
  year         = {2019}
}

@inproceedings{zhang2019three,
  author       = {Guodong Zhang and
                  Chaoqi Wang and
                  Bowen Xu and
                  Roger B. Grosse},
  title        = {{Three Mechanisms of Weight Decay Regularization}},
  booktitle    = {{ICLR}},
  year         = {2019}
}

@article{radford2019language,
  title={{Language models are unsupervised multitask learners}},
  author={Radford, Alec and Wu, Jeffrey and Child, Rewon and Luan, David and Amodei, Dario and Sutskever, Ilya and others},
  journal={OpenAI blog},
  volume={1},
  number={8},
  pages={9},
  year={2019}
}

@book{cohen1988statistical,
  title={{Statistical power analysis for the behavioral sciences}},
  author={Cohen, Jacob},
  year={2013},
}

@article{lakens2013calculating,
  title={{Calculating and reporting effect sizes to facilitate cumulative science: a practical primer for t-tests and ANOVAs}},
  author={Lakens, Dani{\"e}l},
  journal={Frontiers in psychology},
  volume={4},
  pages={863},
  year={2013},
}

@misc{Gokaslan2019OpenWeb,
	title={{OpenWebText Corpus}},
	author={Aaron Gokaslan and Vanya Cohen},
	howpublished={\url{http://Skylion007.github.io/OpenWebTextCorpus}},
	year={2019}
}

\clearpage
\section*{Checklist}

\begin{enumerate}
  \item For all models and algorithms presented, check if you include:
  \begin{enumerate}
    \item A clear description of the mathematical setting, assumptions, algorithm, and/or model.
    [Yes] See Secs.~\ref{sec:symmetries}, \ref{sec:theory}, and~\ref{sec:method}.
    \item An analysis of the properties and complexity (time, space, sample size) of any algorithm.
    [Yes] See Sec.~\ref{sec:theory} and Appendix~\ref{app:practical_full}.
    \item (Optional) Anonymized source code, with specification of all dependencies, including external libraries.
    [Yes] See Appendix~\ref{app:implementation}.
  \end{enumerate}

  \item For any theoretical claim, check if you include:
  \begin{enumerate}
    \item Statements of the full set of assumptions of all theoretical results.
    [Yes] See Sec.~\ref{sec:theory} and Appendix~\ref{app:proofs}.
    \item Complete proofs of all theoretical results.
    [Yes] See Appendix~\ref{app:proofs}.
    \item Clear explanations of any assumptions.
    [Yes] See Secs.~\ref{sec:symmetries} and~\ref{sec:theory}.
  \end{enumerate}

  \item For all figures and tables that present empirical results, check if you include:
  \begin{enumerate}
    \item The code, data, and instructions needed to reproduce the main experimental results (either in the supplemental material or as a URL).
    [Yes] See Appendix~\ref{app:implementation}.
    \item All the training details (e.g., data splits, hyperparameters, how they were chosen).
    [Yes] See Appendices~\ref{app:exp_details}, \ref{app:moved_round2}, \ref{app:additional_experiments}, and~\ref{app:qualitative_protocol}.
    \item A clear definition of the specific measure or statistics and error bars (e.g., with respect to the random seed after running experiments multiple times).
    [Yes] See Appendix~\ref{app:stat_details}.
    \item A description of the computing infrastructure used. (e.g., type of GPUs, internal cluster, or cloud provider).
    [Yes] See Appendix~\ref{app:compute}.
  \end{enumerate}

  \newpage
  \item If you are using existing assets (e.g., code, data, models) or curating/releasing new assets, check if you include:
  \begin{enumerate}
    \item Citations of the creator If your work uses existing assets.
    [Yes] See Appendices~\ref{app:exp_details} and~\ref{app:qualitative_protocol}.
    \item The license information of the assets, if applicable.
    [No] Asset licenses are not listed.
    \item New assets either in the supplemental material or as a URL, if applicable.
    [Not Applicable] No new datasets or model checkpoints are released.
    \item Information about consent from data providers/curators.
    [Not Applicable] No new participant data are collected.
    \item Discussion of sensible content if applicable, e.g., personally identifiable information or offensive content.
    [No] Sensitive corpus content is not analyzed.
  \end{enumerate}

  \item If you used crowdsourcing or conducted research with human subjects, check if you include:
  \begin{enumerate}
    \item The full text of instructions given to participants and screenshots.
    [Not Applicable] No crowdsourcing or human-subject research.
    \item Descriptions of potential participant risks, with links to Institutional Review Board (IRB) approvals if applicable.
    [Not Applicable] No human-subject research.
    \item The estimated hourly wage paid to participants and the total amount spent on participant compensation.
    [Not Applicable] No human participants.
  \end{enumerate}
\end{enumerate}

\clearpage
\appendix
\onecolumn
\thispagestyle{empty}
\aistatstitle{Same Function, Different Trajectory: No Standard Optimizer\\ Is Equivariant to Transformer Reparameterization\\ Supplementary Materials}

\startcontents[apx]
\section*{Appendix Table of Contents}
\printcontents[apx]{l}{1}{\setcounter{tocdepth}{2}}

\section{Related Work}
\label{app:related}
\label{sec:related}

\paragraph{Parameterization and initialization.}
Variance-preserving initialization~\citep{glorot2010understanding,he2015delving} fixes a single scale at the start of training. The Maximal Update Parametrization ($\mu$P) of~\citet{yang2022tensor} provides a principled framework for hyperparameter transfer across widths. T-Fixup~\citep{huang2020improving} instead rescales residual branches so that deep Transformers can be trained more stably at a fixed width. Our work is complementary: $\mu$P governs how to scale across widths, and T-Fixup chooses a particular residual scale, while reparameterization symmetry reveals a continuous family of forward-equivalent configurations at fixed width and depth.

\paragraph{Weight normalization and reparameterization.}
Weight normalization (WeightNorm)~\citep{salimans2016weight} decouples the direction and magnitude of weight vectors, implicitly addressing some scale ambiguities. \citet{dinh2017sharp} explicitly derive the inverse gradient-coordinate law under function-preserving reciprocal rescaling. We use that law as background for the full-state optimizer analysis. \citet{neyshabur2015path} introduced path-regularization to account for rescaling invariances in ReLU networks. Transformer-NFN~\citep{tran2025equivariant} characterizes more general matrix symmetries of multi-head attention; the scalar Q--K and V--O rescalings used here are special cases. Our focus is the response of first-order optimizer states and resumed trajectories to these edits. SAGGS instead selects the Frobenius-balanced representative of an exact scalar pair orbit. Standard WeightNorm has a trainable magnitude and does not impose this pairwise canonicalization.

\paragraph{Depth-dependent techniques.}
Pre-LN~\citep{xiong2020layer} places the normalization before the attention and MLP sub-layers, improving gradient flow relative to post-layer normalization (Post-LN). Admin~\citep{liu2020understanding} diagnoses unbalanced residual-branch magnitudes in Transformers and rescales those magnitudes at initialization. $\sigma$Reparam~\citep{zhai2023stabilizing} constrains Q--K spectral scales to prevent attention entropy collapse, a failure mode adjacent to unconstrained motion along the Q--K orbit. Query-Key Normalization~\citep{henry2020query} removes that free Q--K scale by unit-normalizing the two factors. These techniques act on initialization, the residual stream, or spectral constraints, whereas our analysis concerns reciprocal rescalings inside attention and MLP blocks. The existence of an internal symmetry alone does not determine a training recipe's stability or final PPL.

\paragraph{Weight balancing and equivariance.}
\citet{du2018algorithmic} proved conservation of differences of squared layer norms under gradient flow in homogeneous networks and established discrete-time balancing results for matrix factorization. Conservation alone does not drive an arbitrary initial imbalance to zero. Equi-normalization~\citep{stock2019equinormalization} uses the same ReLU scale freedom to iteratively rebalance hidden units toward a minimum-norm representative; the procedure does not canonicalize first-order optimizer state. Symmetry Teleportation~\citep{zhao2022teleportation} moves along a function-preserving orbit to change the local descent geometry. SAGGS uses a fixed orbit-selection criterion: the Frobenius-balanced representative of an exact scalar pair.

\paragraph{Symmetry and optimization dynamics.}
\citet{kunin2021neural} and \citet{tanaka2021noether} analyze how continuous network symmetries constrain learning dynamics and how finite-step and adaptive updates break associated conservation laws. Discrete permutation symmetries, whose role in linear mode connectivity was isolated by~\citet{entezari2022role} and later used for model merging~\citep{ainsworth2023gitrebasin}, identify a different orbit: Those methods rearrange hidden units rather than rescaling a bilinear pair. Our focus is on complete discrete optimizer states and canonicalized restarts along the scalar rescaling orbit. The moment-scaling identities follow from the Adam recursions~\citep{kingma2015adam}. Theorem~\ref{thm:adam_equivariance} and Corollary~\ref{cor:canonical_restart} give the corresponding discrete-time AdamW and restart rules, complementing the continuous-time analysis of~\citet{kunin2021neural}.

\FloatBarrier
\section{Notation and Architectural Details}
\label{app:arch_variants}

\subsection{Attention, MLPs, and Affine Projections}
\label{app:notation_setup_moved}
\label{sec:notation}

Consider an $L$-layer Transformer~\citep{vaswani2017attention} with hidden dimension $d$, attention head dimension $d_k$, and MLP intermediate dimension $d_{\mathrm{ff}}$. At layer $\ell \in \{1, \ldots, L\}$, the self-attention sub-layer computes:
\begin{equation}
  \mathrm{Attn}_\ell(\mat{X}) \coloneqq \mat{W}^O_\ell \left[ \mathrm{softmax}\left( \frac{(\mat{W}^Q_\ell \mat{X})^\transpose (\mat{W}^K_\ell \mat{X})} {\sqrt{d_k}} \right) (\mat{W}^V_\ell \mat{X})^\transpose \right]^\transpose,
  \label{eq:attention}
\end{equation}
where $\mat{W}^Q_\ell, \mat{W}^K_\ell \in \reals^{d_k \times d}$, $\mat{W}^V_\ell \in \reals^{d_k \times d}$, and $\mat{W}^O_\ell \in \reals^{d \times d_k}$. The equation is for a single head; multi-head attention concatenates the per-head matrices.

The MLP sub-layer with ReLU activation is:
\begin{equation}
  \mathrm{MLP}_\ell(\mat{x}) \coloneqq \mat{W}^2_\ell \sigma(\mat{W}^1_\ell \mat{x}),
  \label{eq:mlp}
\end{equation}
where $\mat{W}^1_\ell \in \reals^{d_{\mathrm{ff}} \times d}$, $\mat{W}^2_\ell \in \reals^{d \times d_{\mathrm{ff}}}$, and $\sigma$ denotes the activation function.

Let $\loss(\param)$ denote the data loss evaluated at parameters $\param \coloneqq \{\mat{W}^Q_\ell, \mat{W}^K_\ell, \mat{W}^V_\ell, \mat{W}^O_\ell, \mat{W}^1_\ell, \mat{W}^2_\ell\}_{\ell=1}^L$. The equations omit biases.

For affine projections, scale the Q--K biases with their respective matrices, the V bias with V, and the ReLU up-projection or SwiGLU value bias with its up matrix; leave output biases and the SwiGLU gate unchanged.  These co-transforms preserve the forward and gradient-coordinate identities. Without these co-transforms, matrix-only rescaling need not be a symmetry of a biased layer.

\subsection{Grouped-Query Attention and Rotary Embeddings}
\label{app:gqa_detail}

\paragraph{Modern variants.}
\label{par:modern_variants}
The pairs of Table~\ref{tab:symmetry_catalogue} remain exact in the modern architecture that we use for the 31M replication and the checkpoint-resume experiments (Sec.~\ref{sec:experiments}). This architecture combines SwiGLU with GQA, RMSNorm, and RoPE. RMSNorm acts on the residual stream rather than on the paired matrices, and RoPE only rotates queries and keys, so the factors $c$ and $c^{-1}$ still cancel. GQA shares each key and value head across a group of query heads, so the Q--K and V--O rescalings use one factor per group.

\paragraph{Grouped-query attention.}
GQA~\citep{ainslie2023gqa} shares one $\mat{W}^K$ and one $\mat{W}^V$ across $H/G$ query heads, where $G$ is the group count. The Q--K orbit becomes group-tied: Any rescaling $\mat{W}^Q_{\ell,h} \mapsto c\mat{W}^Q_{\ell,h}$ for the $H/G$ query heads in a group must be paired with a single $\mat{W}^K_{\ell,g} \mapsto c^{-1}\mat{W}^K_{\ell,g}$ for the group's shared key. The dimension of the orbit is reduced from $H$ scalars per layer in the multi-head case to $G$ scalars per layer; the gradient covariance in \eqref{eq:grad_transform} applies to each group-pair independently. The per-pair Frobenius rebalance of SAGGS extends by treating each group's $H/G$ Q heads and shared K head as one balancing unit: The Q-side norm is $(\sum_{h\in g}\frobnorm{\mat{W}^Q_{\ell,h}}^2)^{1/2}$. For V--O, scale the shared V head by $c$ and all $H/G$ corresponding output-projection blocks by $c^{-1}$, using their joint Frobenius norm on the output side.

\paragraph{Residual-stream normalization and rotary embeddings.}
RMSNorm~\citep{zhang2019rmsnorm} pre-normalization acts on the residual stream and therefore leaves the internal pair identities unchanged. RoPE~\citep{su2024roformer} applies a position-dependent linear orthogonal map to queries and keys, so $(c\mat{W}^Q\mat{x})^\transpose(c^{-1}\mat{W}^K\mat{x})$ is invariant before and after the rotary map. The LLaMA-style block combining RMSNorm, RoPE, and SwiGLU~\citep{touvron2023llama} with GQA~\citep{ainslie2023gqa} remains inside the catalog for the Q--K, V--O, and SwiGLU value--output pairs used by SAGGS.

\subsection{Interaction with Layer Normalization}
\label{app:layernorm_detail}

\paragraph{Layer-normalization interaction.}
Pre-LN normalizes the input to each sub-layer, which might seem to eliminate the effect of weight norm imbalance. However, layer normalization acts on the residual stream, \ie{} the sub-layer input, not on the internal weight matrices. The gradient transformation law operates on the internal bilinear structure, and layer normalization does not interact with this symmetry. Specifically, the layer-normalization output $\hat{\mat{x}} \coloneqq \mathrm{LN}(\mat{x})$ is a fixed function of the residual stream regardless of the internal parameterization of the sub-layer, so the reparameterization $(\mat{A}_\ell, \mat{B}_\ell) \to (c\mat{A}_\ell, c^{-1}\mat{B}_\ell)$ preserves $\mathrm{SubLayer}_\ell(\hat{\mat{x}})$ and hence the gradient transformation law continues to hold.

To make the layer-normalization argument precise, write $f_{\mat{A},\mat{B}}(\mat{X})$ for the sub-layer map with all other sub-layer parameters fixed. Every exact pair in Table~\ref{tab:symmetry_catalogue} satisfies
\begin{align*}
  f_{c\mat{A},c^{-1}\mat{B}}(\mat{X}) &=f_{\mat{A},\mat{B}}(\mat{X}).
\end{align*}
The form $\mat{B}g(\mat{A}\mat{X})$ describes a ReLU MLP, but does not represent a Q--K attention pair: Both Q and K enter the logits, with $(c\mat{A}\mat{X})^\transpose(c^{-1}\mat{B}\mat{X}) =(\mat{A}\mat{X})^\transpose(\mat{B}\mat{X})$. In Pre-LN, the residual-block map is $\mat{X}+f_{\mat{A},\mat{B}}(\mathrm{LN}(\mat{X}))$; in Post-LN, the residual-block map is $\mathrm{LN}(\mat{X}+f_{\mat{A},\mat{B}}(\mat{X}))$. Substituting the same sub-layer identity into either expression proves forward invariance of the whole block. Lemma~\ref{lem:invariance_grad_law} then gives the same inverse gradient-coordinate factors in both cases.

\subsection{Other Architectural Variants}

The following extensions and boundaries complete the catalog in Sec.~\ref{sec:symmetry_catalogue}. Cross-attention, mixture-of-experts, and state-space blocks are not used in the primary training comparisons.

\paragraph{Cross-attention in encoder and decoder models.}
The attention pairs in Table~\ref{tab:symmetry_catalogue} also apply when queries, keys, and values come from different sources: Their bilinear and linear compositions do not require the two input sequences to coincide.

\paragraph{Mixture-of-experts models.}
Each expert in a sparse mixture-of-experts MLP~\citep{shazeer2017outrageously,fedus2022switch} with ReLU experts contains an independent $(\mat{W}^1_e, \mat{W}^2_e)$ pair. Conditional routing means a given token is processed by only $k$ experts, but the symmetry $\mat{W}^{1\prime}_e \coloneqq c_e \mat{W}^1_e$, $\mat{W}^{2\prime}_e \coloneqq c_e^{-1}\mat{W}^2_e$ is per-expert and forward-equivalent regardless of the routing decision: The routing gate is shared across experts and unaffected. The catalog adds $E$ independent scalar orbits per layer, where $E$ is the expert count.  For SwiGLU experts, the exact pair is V--O with the gate fixed. GELU up--down rescaling is not an exact symmetry.

\paragraph{State-space models.}
Mamba's selective state-space block~\citep{gu2023mamba} contains an input projection $\mat{W}^{\rm in}$ and an output projection $\mat{W}^{\rm out}$ around a state-space recurrence. The input-dependent discretization step $\Delta$ enters nonlinearly, so the recurrence is not generally homogeneous in its input. Reciprocal rescaling of the input and output projections need not preserve the function, and state-space blocks are excluded from Table~\ref{tab:symmetry_catalogue}. Any sub-block that has the form $\mat{B} g(\mat{A}\mat{x})$ with $g$ positively homogeneous inherits the corresponding pair.  ReLU has this property; in SwiGLU, the same pair inheritance holds for the value path with the gate fixed, not for joint scaling of the gate and value projections.

\FloatBarrier
\section{Proofs and Optimizer Analysis}
\label{app:proofs}

\subsection{Gradient Covariance from Forward Invariance}
\label{app:grad_law_proof}
\label{app:grad_law_general}

We record the general chain-rule identity underlying \eqref{eq:grad_transform} and the optimizer-state proofs. The reciprocal-scaling instance is standard~\citep[Theorem~3]{dinh2017sharp}.

\begin{lemma}[Forward invariance implies the gradient law]
\label{lem:invariance_grad_law}
Let $F : \reals^{n} \to \reals$ and let $J \coloneqq \mathrm{diag}(s_1, \ldots, s_n)$ with all $s_i > 0$. Suppose that $F(J\param) = F(\param)$ for every $\param \in \reals^{n}$. Then $F$ is differentiable at $\param$ if and only if $F$ is differentiable at $J\param$, and at every such point
\begin{equation}
  \grad F(J\param) = J^{-1}\grad F(\param), \qquad\text{\ie}\qquad \partial_i F(J\param) = s_i^{-1}\partial_i F(\param) \quad\text{for } i = 1, \ldots, n.
  \label{eq:lemma_invariance}
\end{equation}
\end{lemma}

\begin{proof}
Let $G \coloneqq F \circ J$, \ie{} $G(\param) = F(J\param)$.  By hypothesis, $G = F$ as functions on $\reals^{n}$.  Since $J$ is an invertible linear map, $G$ is differentiable at $\param$ if and only if $F$ is differentiable at $J\param$; for the converse write $F = G \circ J^{-1}$. In that case, the chain rule gives $\grad G(\param) = J^{\transpose}\grad F(J\param) = J\grad F(J\param)$. Because $G = F$, $G$ is differentiable at $\param$ exactly when $F$ is, with $\grad G(\param) = \grad F(\param)$.  Combining the two statements, $F$ is differentiable at $\param$ if and only if $F$ is differentiable at $J\param$, and then $J\grad F(J\param) = \grad F(\param)$; multiplying by $J^{-1}$ yields~\eqref{eq:lemma_invariance}.
\end{proof}

For a symmetry pair, take $J=\mathrm{diag}(c I_A,c^{-1}I_B)$ and leave all other coordinates unchanged. Forward invariance supplies the hypothesis, and the two blocks of~\eqref{eq:lemma_invariance} give~\eqref{eq:grad_transform}. This hypothesis holds for the data loss $\loss$ but not for a loss that also includes penalties on the weights, such as L2 regularization, because rescaling a pair changes such penalties. Disjoint exact pairs can be transformed simultaneously by the same argument with a diagonal map containing their respective scales.

\subsection{Unique Balanced Representative}
\label{app:orbit_canonicalization_proof}

\begin{proof}[Proof of Proposition~\ref{cor:saggs_orbit}]
Throughout, $\frobnorm{\mat{A}_p} > 0$ and $\frobnorm{\mat{B}_p} > 0$, so that $\rho_p \coloneqq \frobnorm{\mat{A}_p}/\frobnorm{\mat{B}_p}$ and $c_p^\star \coloneqq \rho_p^{1/2}$ are well defined, and $T_c(\mat{A}, \mat{B}) \coloneqq (c\mat{A}, c^{-1}\mat{B})$ denotes the orbit move with factor $c > 0$.

\paragraph{Balance.}
$\frobnorm{\mat{A}_p^\star} = \frobnorm{\mat{A}_p}/c_p^\star = \frobnorm{\mat{A}_p}^{1/2}\frobnorm{\mat{B}_p}^{1/2}$ and $\frobnorm{\mat{B}_p^\star} = c_p^\star\frobnorm{\mat{B}_p} = \frobnorm{\mat{A}_p}^{1/2}\frobnorm{\mat{B}_p}^{1/2}$, so $\frobnorm{\mat{A}_p^\star} = \frobnorm{\mat{B}_p^\star}$.

\paragraph{Uniqueness and independence of the orbit representative.}
The orbit of $(\mat{A}_p, \mat{B}_p)$ is $\mathcal{O}_p \coloneqq \{T_c(\mat{A}_p, \mat{B}_p) : c > 0\}$, and $c \mapsto T_c(\mat{A}_p, \mat{B}_p)$ is injective because $c\mat{A}_p = \tilde{c}\mat{A}_p$ with $\mat{A}_p \neq \mat{0}$ forces $c = \tilde{c}$.  The point $T_c(\mat{A}_p, \mat{B}_p)$ is Frobenius-balanced if and only if $c\frobnorm{\mat{A}_p} = c^{-1}\frobnorm{\mat{B}_p}$, \ie{} if and only if $c^{2} = \frobnorm{\mat{B}_p}/\frobnorm{\mat{A}_p}$, whose unique positive solution is $c = 1/c_p^\star$.  Hence $(\mat{A}_p^\star, \mat{B}_p^\star) = T_{1/c_p^\star}(\mat{A}_p, \mat{B}_p)$ is the unique balanced point of $\mathcal{O}_p$.  The balanced point depends only on the orbit and not on the point from which the representative is computed: If $(\tilde{\mat{A}}, \tilde{\mat{B}}) = T_c(\mat{A}_p, \mat{B}_p)$, then $\tilde{c}^\star = (c\frobnorm{\mat{A}_p}/(c^{-1}\frobnorm{\mat{B}_p}))^{1/2} = c c_p^\star$, so $\tilde{\mat{A}}/\tilde{c}^\star = c\mat{A}_p/(c c_p^\star) = \mat{A}_p^\star$ and $\tilde{c}^\star\tilde{\mat{B}} = c c_p^\star c^{-1}\mat{B}_p = \mat{B}_p^\star$.  In particular, the step is idempotent: At $(\mat{A}_p^\star, \mat{B}_p^\star)$ one has $\rho_p = 1$ and $c_p^\star = 1$, so a second application is the identity.

\paragraph{Forward pass.}
Since $(\mat{A}_p^\star, \mat{B}_p^\star) \in \mathcal{O}_p$, the exact-pair assumption ensures that the step leaves the forward map unchanged.
\end{proof}

\subsection{Recovery of a Balanced Checkpoint}
\label{app:recovery_detail}

\paragraph{Recovery after an orbit edit of a balanced checkpoint.}
\label{par:balanced_checkpoint_recovery}
If the imbalance was created from a balanced pair by an exact orbit move $(\mat{A},\mat{B})\mapsto(c\mat{A},c^{-1}\mat{B})$, this step applies the inverse orbit move and removes the associated diagonal coordinate factor. Suppose $(\mat{A}_p, \mat{B}_p) = T_c(\mat{A}_p^{0}, \mat{B}_p^{0})$ for some balanced pair with $\frobnorm{\mat{A}_p^{0}} = \frobnorm{\mat{B}_p^{0}}$ and some $c > 0$.  Then $c_p^\star = (c\frobnorm{\mat{A}_p^{0}}/(c^{-1}\frobnorm{\mat{B}_p^{0}}))^{1/2} = c$, so the rebalancing step is $T_{1/c_p^\star} = T_{c^{-1}} = T_c^{-1}$ and returns $(\mat{A}_p^{0}, \mat{B}_p^{0})$ exactly.  By \eqref{eq:grad_transform}, returning to the original balanced point also restores its gradient coordinates.

\paragraph{One-shot canonicalization.}
Algorithm~\ref{alg:rebalancing} applies exactly $\mat{A}_p \leftarrow \rho_p^{-1/2}\mat{A}_p$ and $\mat{B}_p \leftarrow \rho_p^{1/2}\mat{B}_p$, which is $T_{1/c_p^\star}$. By Proposition~\ref{cor:saggs_orbit}, a single application reaches the unique balanced representative of the orbit without changing the forward pass, and any further application is the identity; this fact is the one-shot canonicalizer property for pair-norm imbalance on the symmetry orbit.

Canonicalization preserves the current function and data loss. At a fixed checkpoint, every point of its exact symmetry orbit has the same validation PPL. Proposition~\ref{cor:saggs_orbit} selects the balanced representative; subsequent training from that point is evaluated empirically.

\subsection{AdamW Moment Covariance at Corresponding Orbit Points}
\label{app:adam_extension}

For a fixed sequence of corresponding orbit points, the Adam moment recursions determine the moments in the transformed coordinates. The following lemma supplies the identities used in Theorem~\ref{thm:adam_equivariance}; the lemma does not assume that common hyperparameters preserve the correspondence during training.

\paragraph{Corresponding-history setup.}
Let $g_t \in \reals^{n}$ denote the data-loss gradient at step $t$ and let $\hat{m}_t,\hat{v}_t$ be the bias-corrected first- and second-moment estimates of AdamW with $\beta_1\in[0,1)$ and $\beta_2\in(0,1)$. Under the reparameterization $\param' \coloneqq T_{c}(\param)$ with diagonal Jacobian $J_c\coloneqq\mathrm{diag}(s_1,\ldots,s_n)$, where $s_i\coloneqq c$ on $A$, $s_i\coloneqq c^{-1}$ on $B$, and $s_i\coloneqq1$ elsewhere, let the primed quantities be computed from the gradients in the new coordinates at the corresponding orbit points $\param'_\tau \coloneqq T_c(\param_\tau)$, $\tau \le t$, with the same hyperparameters and zero-initialized moment buffers.

\begin{lemma}[Adam moment covariance]
\label{lem:adam_moment_covariance}
Under the corresponding-history setup above, the new-coordinate gradients satisfy $g_{t,i}' = s_i^{-1}g_{t,i}$ for every $i$ and every $t$, so $\hat{m}_{t,i}' = s_i^{-1}\hat{m}_{t,i}$ and $\hat{v}_{t,i}' = s_i^{-2}\hat{v}_{t,i}$.
\end{lemma}

\begin{proof}[Proof of Lemma~\ref{lem:adam_moment_covariance}]
With all parameters outside the pair held fixed, the data loss satisfies $\loss \circ T_c = \loss$ by the exact-pair assumption, and $T_c$ is the linear map $J_c$.  Lemma~\ref{lem:invariance_grad_law} therefore gives, at every pair of corresponding points, $g'_\tau = J_c^{-1}g_\tau$, \ie{} $g'_{\tau,i} = s_i^{-1}g_{\tau,i}$ for all $\tau \le t$ and all $i$.  The moment recursions $m_\tau \coloneqq \beta_1 m_{\tau-1} + (1-\beta_1)g_\tau$ and $v_\tau \coloneqq \beta_2 v_{\tau-1} + (1-\beta_2)g_\tau^{\odot 2}$ with $m_0 \coloneqq v_0 \coloneqq 0$ are, coordinate-wise, linear in the sequence $(g_{\tau,i})_\tau$ and in $(g_{\tau,i}^{2})_\tau$, respectively.  By induction on $\tau$: $m'_{0,i} = 0 = s_i^{-1}m_{0,i}$, and if $m'_{\tau-1,i} = s_i^{-1}m_{\tau-1,i}$ then $m'_{\tau,i} = \beta_1 s_i^{-1}m_{\tau-1,i} + (1-\beta_1)s_i^{-1}g_{\tau,i} = s_i^{-1}m_{\tau,i}$; likewise $v'_{\tau,i} = \beta_2 s_i^{-2}v_{\tau-1,i} + (1-\beta_2)s_i^{-2}g_{\tau,i}^{2} = s_i^{-2}v_{\tau,i}$.  The bias corrections $\hat{m}_t \coloneqq m_t/(1-\beta_1^{t})$ and $\hat{v}_t \coloneqq v_t/(1-\beta_2^{t})$ are positive scalars that do not depend on the coordinate, so $\hat{m}'_{t,i} = s_i^{-1}\hat{m}_{t,i}$ and $\hat{v}'_{t,i} = s_i^{-2}\hat{v}_{t,i}$. Decoupled weight decay acts separately on the parameters and does not enter the moments.
\end{proof}

\subsection{AdamW Equivariance and Canonical Restart}
\label{app:adam_equivariance_proof}

We prove the coordinate criterion, the functional counterexample, and the canonical restart identity separately.  All calculations are in exact arithmetic.

\begin{proof}[Proof of Theorem~\ref{thm:adam_equivariance}]
For the moment recursions and the coordinate criterion, suppose at step $t$ that $\param'_t=S\param_t$, $m'_t=S^{-1}m_t$, and $v'_t=S^{-2}v_t$. Since the sampled loss is the same in the two runs and $\loss_t\circ S=\loss_t$, differentiating this identity gives $S^\transpose\grad\loss_t(S\param_t)=\grad\loss_t(\param_t)$. The diagonal matrix $S$ is symmetric, so $g'_t=S^{-1}g_t$, where $g_t\coloneqq\grad\loss_t(\param_t)$. The Adam moment recursions therefore give
\begin{align*}
 m'_{t+1} &=\beta_1 S^{-1}m_t+(1-\beta_1)S^{-1}g_t =S^{-1}m_{t+1},\\
 v'_{t+1} &=\beta_2 S^{-2}v_t+(1-\beta_2)(S^{-1}g_t)^{\odot2} =S^{-2}v_{t+1}.
\end{align*}
Here $x^{\odot2}$ denotes componentwise squaring.  Shared bias-correction counters multiply each moment by the same scalar in both runs, so the same identities hold for $\hat m$ and $\hat v$. This argument is inductive: The induction assumes corresponding parameters only at the current step, not at future iterates.

Fix a coordinate and a step, suppress their indices, and put $s\coloneqq s_i$, $a\coloneqq\hat m_{t+1,i}$, $w\coloneqq\sqrt{\hat v_{t+1,i}}$. Because $s>0$, the primed adaptive term is $\eta'a/(w+s\epsilon')$.  Thus universal commutation of the parameter-update formula is precisely the identity
\begin{equation}
 (1-\eta'\omega')s\theta-\frac{\eta'a}{w+s\epsilon'} =s(1-\eta\omega)\theta-\frac{s\eta a}{w+\epsilon}
 \label{eq:adam_commutation_identity}
\end{equation}
for every $\theta,a$ and every positive $w$, and by continuity where $w=0$ and the denominators remain positive. First set $a=0$ and vary $\theta$.  This substitution yields $\eta'\omega'=\eta\omega$. With these decay terms equal, take $a\ne0$ and cancel $a$ to obtain
\begin{align*}
  \eta'(w+\epsilon)&=s\eta(w+s\epsilon')\qquad\text{for every }w>0.
\end{align*}
Equality of the coefficients of $w$ gives $\eta'=s\eta$. Equality of the constant coefficients, together with $s\eta>0$, then gives $\epsilon'=\epsilon/s$.  Substituting the learning-rate identity into the decay identity gives $\omega'=\omega/s$. This calculation proves necessity for the update formula on its full state domain; two particular trajectories may still agree by accident without these identities. Conversely these three identities make both sides of~\eqref{eq:adam_commutation_identity} equal, proving sufficiency. Combined with the moment calculation above, sufficiency completes an induction from the initial state and proves trajectory equivariance. The reasoning also permits a deterministic time-dependent schedule: The three identities must hold at each step of that schedule.
\end{proof}

\begin{proof}[Verification of Example~\ref{rem:adam_moment_covariance}]
For $\loss(a,b)\coloneqq(ab)^2/2$, the gradients are $g_a=ab^2$ and $g_b=a^2b$. The gradients are $(1,1)$ at $(1,1)$ and $(c^{-1},c)$ at $(c,c^{-1})$. With zero initial moments, the first bias-corrected moments are $\hat m_1=g$ and $\hat v_1=g^{\odot2}$, for any $\beta_1,\beta_2\in[0,1)$.  Every gradient component is positive, so all four denominators are positive and the normalized updates with $\epsilon=0$ are exactly $1$. With no weight decay, this update rule gives
\begin{align*}
  (a_1,b_1)&=(1-\eta,1-\eta),\\
  (a'_1,b'_1)&=(c-\eta,c^{-1}-\eta).
\end{align*}
Hence
\begin{align*}
  a_1b_1&=1-2\eta+\eta^2,\\
  a'_1b'_1&=1-\eta(c+c^{-1})+\eta^2,\\
  a'_1b'_1-a_1b_1&=-\eta(c+c^{-1}-2).
\end{align*}
The inequality $c+c^{-1}\geq2$ proves \eqref{eq:adam_functional_counterexample}.  For fixed $\eta>0$, $a'_1b'_1\to-\infty$ as $c\to\infty$, so $\loss(a'_1,b'_1)=(a'_1b'_1)^2/2\to\infty$. The initial product is $1$ in both runs.  The first and second moments satisfy exactly the relations in the proof of Theorem~\ref{thm:adam_equivariance}, so the discrepancy cannot be attributed to stale moments.  Finally, $c=e^{\lambda\ell}$ implies $c+c^{-1}-2=2[\cosh(\lambda\ell)-1]$, giving the stated depth-schedule consequence.
\end{proof}

\begin{proof}[Proof of Corollary~\ref{cor:canonical_restart}]
For one pair, the norm ratio at $S\param$ is $c^2\rho$. Consequently its $A$-side balancing multiplier is $(c^2\rho)^{-1/2}=c^{-1}\rho^{-1/2}$, and its $B$-side multiplier is $(c^2\rho)^{1/2}=c\rho^{1/2}$. The pairs are disjoint and all other coordinates are unchanged, so
\begin{equation}
 R_{S\param}S=R_{\param}.
 \label{eq:canonical_matrix_identity}
\end{equation}
All matrices here are positive diagonal matrices and commute.  These commutation relations imply that
\begin{align*}
  R_{S\param}S\param&=R_{\param}\param,\\
  R_{S\param}^{-1}S^{-1}m&=R_{\param}^{-1}m,\\
  R_{S\param}^{-2}S^{-2}v&=R_{\param}^{-2}v.
\end{align*}
These three identities are exactly the three components of~\eqref{eq:canonical_restart_identity}.  The transformations are well-defined because both norms in every pair are positive.  Each parameter balancing step is an exact symmetry and hence preserves the forward map.

After the two canonicalizations, the numerical parameters and moments are identical.  With equal counters, common hyperparameters, and the same next minibatch and random choices, the next gradients, moments, and parameter updates are identical.  Induction proves equality at every subsequent step.  This implication also holds for common deterministic clipping applied after canonicalization, because the two gradients are then identical; clipping need not commute with an arbitrary $S$ for this conclusion.  If moments are reset, the same proof applies since $S^{-1}0=S^{-2}0=0$. No identity with an uncanonicalized reference or an arbitrary stale state is used or implied.
\end{proof}

\subsection{Orbit-Equivariant First-Order Updates}
\label{app:equivariant_rules_proof}

Throughout, $S_c$ denotes the element of $\mathcal G$ that acts on the fixed pair $p$ with factor $c>0$ and as the identity on every other block; that is, $S_c$ multiplies the $\mat{A}_p$-coordinates by $c$ and the $\mat{B}_p$-coordinates by $c^{-1}$, while $S_c^{-1}$ multiplies those coordinates by $c^{-1}$ and $c$, respectively.

\begin{proof}[Proof of Theorem~\ref{thm:equivariant_rules}]
For the degree $-1$ condition and the stated obstructions, fix $c>0$, a step $t$, and an admissible history $(\param_{1:t},g_{1:t})$, and write $x\coloneqq x_A\coloneqq g^{A_p}_{1:t}$.  The $\mat{A}_p$-coordinates of $S_c^{-1}g_{1:t}$ form the history $c^{-1}x$, and the $\mat{A}_p$-block of $S_c\param_t$ is $c\mat{A}_{p,t}$.  By \eqref{eq:block_local_rule}, the $\mat{A}_p$-block of the left-hand side of~\eqref{eq:orbit_equivariance} equals $-\eta_t\Psi_t(c^{-1}x)-\eta_t\omega_tc\mat{A}_{p,t}$, whereas the $\mat{A}_p$-block of the right-hand side equals $c(-\eta_t\Psi_t(x)-\eta_t\omega_t\mat{A}_{p,t})$.  Equating the two, canceling the decay terms, and dividing by $-\eta_t<0$ yields \eqref{eq:degree_minus_one}.  The history $x$ was arbitrary in the domain of $\Psi_t$.

\paragraph{Homogeneous case.}  Suppose $\Psi_t(sx)=s^{k}\Psi_t(x)$ for all $s>0$ with $k\neq-1$.  Applying~\eqref{eq:degree_minus_one} with $c=2$ and homogeneity with $s=1/2$ gives $2\Psi_t(x)=\Psi_t(x/2)=2^{-k}\Psi_t(x)$, so $(2-2^{-k})\Psi_t(x)=0$. Because $k\neq-1$ we have $2^{-k}\neq2$, hence $\Psi_t(x)=0$.

\paragraph{Bounded case.}  Suppose $\norm{\Psi_t(y)}\le M$ whenever $\norm{y}<\delta$.  Given any $x$, choose $c>\norm{x}/\delta$; then $\norm{c^{-1}x}<\delta$ and~\eqref{eq:degree_minus_one} gives $\norm{\Psi_t(x)}=c^{-1}\norm{\Psi_t(c^{-1}x)}\le M/c$.  Letting $c\to\infty$ shows $\Psi_t(x)=0$.

For the $\mat{B}_p$-block, $S_c^{-1}$ multiplies the gradient history by $c$ and $S_c$ multiplies the parameters by $c^{-1}$, so the same computation yields $\Psi^{B}_t(cx_B)=c^{-1}\Psi^{B}_t(x_B)$, which is \eqref{eq:degree_minus_one} with $c$ replaced by $c^{-1}$; the two consequences follow verbatim.

\paragraph{Coverage.}  Gradient descent, $\Psi_t(x)\coloneqq x_t$, and heavy-ball momentum, $\Psi_t(x)\coloneqq\sum_{\tau\le t}\beta^{t-\tau}x_\tau$, are homogeneous of degree $1$.  AdaGrad, RMSProp, and Adam with $\epsilon=0$ divide a linear function of the history by the square root of a quadratic one and are homogeneous of degree $0$, as are signSGD and Lion, since $\mathrm{sgn}(sy)=\mathrm{sgn}(y)$ for $s>0$. For Shampoo, put $L_t\coloneqq\sum_{\tau\le t}x_\tau x_\tau^\transpose$ and $R_t\coloneqq\sum_{\tau\le t}x_\tau^\transpose x_\tau$. These Gram factors can be singular, including at the first step for a rectangular gradient. The zero-stabilizer limiting direction is
\begin{align*}
  (L_t^\dagger)^{1/4}x_t(R_t^\dagger)^{1/4},
\end{align*}
where $\dagger$ denotes the Moore--Penrose pseudoinverse; equivalently, the inverse fourth root is zero on each nullspace. This pseudoinverse formula agrees with $L_t^{-1/4}x_tR_t^{-1/4}$ when both factors are positive definite. This pseudoinverse direction is the limit of the regularized direction because $\ker L_t\subseteq\ker x_t^\transpose$ and $\ker R_t\subseteq\ker x_t$: No component of $x_t$ lies in the divergent nullspace factors. Scaling the history by $s>0$ scales each Gram factor by $s^2$, so the limiting direction scales as $s^{-1/2}\cdot s\cdot s^{-1/2}=s^0$. For Adam or AdamW with $\epsilon>0$, the bias-corrected first moment is a convex combination of the past gradients of the coordinate, so $\abs{\Psi_{t,i}(x)}\le\max_{\tau\le t}\abs{x_{\tau,i}}/\epsilon$, which tends to $0$ with the history; thus $\Psi_t$ is bounded near the zero history.  Decoupled weight decay is already included in~\eqref{eq:block_local_rule}, and Lion's decay term $\eta\omega\param$ has the same form.  None of these rules is identically zero, so none is orbit-equivariant.
\end{proof}

\begin{proof}[Proof of Corollary~\ref{cor:half_compensation}]
For the orbit-related histories, the $\mat{A}_p$-history is $c^{-1}x_A$ and the $\mat{B}_p$-history is $cx_B$.  By homogeneity, $\Delta\mat{A}'=-\eta_t\Psi_t(c^{-1}x_A)=c^{-k}(-\eta_t\Psi_t(x_A)) =c^{-k}\Delta\mat{A}$ and $\Delta\mat{B}'=-\eta_t\Psi^{B}_t(cx_B)=c^{k}\Delta\mat{B}$. Since $\frobnorm{\mat{A}'}=c\frobnorm{\mat{A}}$ and $\frobnorm{\mat{B}'}=c^{-1}\frobnorm{\mat{B}}$, \eqref{eq:relative_update_tilt} follows.
\end{proof}

\paragraph{Depth-schedule specialization.}
\label{app:depth_detail}
For the nonzero increments in Corollary~\ref{cor:half_compensation}, taking logarithms of the ratio of the two nonzero relative increments gives a shift of $\log c^{1+k}-\log c^{-(1+k)}=2(1+k)\log c$, which equals $2(1+k)\lambda\ell$ for $c=e^{\lambda\ell}$; the same difference is $4\lambda\ell$ for $k=1$ and $2\lambda\ell$ for $k=0$. For $\lambda\ell\ne0$, the shift is zero only for $k=-1$; an identity orbit move gives zero for every $k$.

\paragraph{Full increments with decoupled decay.}
When $\omega_t\ne0$, the corresponding full increments instead satisfy
\begin{align*}
  \Delta\mat{A}'&=-\eta_t c^{-k}\Psi_t(x_A)-\eta_t\omega_t c\mat{A},\\
  \Delta\mat{B}'&=-\eta_t c^k\Psi^B_t(x_B)-\eta_t\omega_t c^{-1}\mat{B}.
\end{align*}
The data and decay terms carry different powers of $c$ for $k\ne-1$, so their sum has no single scaling exponent in general. For example, at the scalar pair $a=b=1$ with loss $(ab)^2/2$, a first zero-$\epsilon$ AdamW step has data directions $u_a=u_b=1$. With $c=2$, $\omega=1$, and a common $\eta>0$, the multipliers of the full relative increments are $3/4$ on $A$ and $3/2$ on $B$, instead of the data-only values $1/2$ and $2$. The B-to-A multiplier is $2$, whereas the data-only law gives $c^2=4$.

\subsection{Normalized Directions and the Scope of the Obstruction}
\label{app:theory_details_moved}

\paragraph{Adam normalized direction.}
For corresponding gradient histories under $\theta'_i\coloneqq s_i\theta_i$, $s_i>0$, and covariantly transformed initial moments, including zero initialization, Adam's moment recursions give $\hat m'_i=\hat m_i/s_i$ and $\hat v'_i=\hat v_i/s_i^2$. With a common additive stabilizer $\epsilon_i$, its normalized data direction is
\begin{equation}
\begin{aligned}
  u'_i&\coloneqq\frac{\hat m_i/s_i}{\sqrt{\hat v_i}/s_i+\epsilon_i}\\
      &=\frac{\hat m_i}{\sqrt{\hat v_i}+s_i\epsilon_i},\\
  u_i&\coloneqq\frac{\hat m_i}{\sqrt{\hat v_i}+\epsilon_i}.
\end{aligned}
  \label{eq:adam_direction_moved}
\end{equation}
Thus the gradient-scale cancellation~\citep{kingma2015adam} is exact when $\epsilon_i=0$ and the denominators are positive; the cancellation generally fails for a common positive $\epsilon_i$. Even when $u'_i=u_i$, preserving an orbit-related training trajectory requires the parameter displacement to scale with the parameters.

\paragraph{Block-locality examples.}
Block locality and the stated regularity alternatives are essential. For example, the unbounded block-local rule that sets $\Psi(x)\coloneqq x_t/\frobnorm{x_t}^2$ for $x_t\ne0$ and $\Psi(x)\coloneqq0$ at $x_t=0$ has degree $-1$ and satisfies the required scaling.  Moreover, the parameter-free, pair-coupled directions $\Psi_A\coloneqq\frobnorm{g^B_t}^2g^A_t$ and $\Psi_B\coloneqq\frobnorm{g^A_t}^2g^B_t$ are polynomial and equivariant: These directions transform by $c$ and $c^{-1}$, respectively.  Thus the obstruction does not cover all parameter-free first-order methods.

\paragraph{Standard optimizer instances.}
Theorem~\ref{thm:equivariant_rules} applies to gradient descent with or without heavy-ball momentum, both homogeneous of degree~$1$. The degree-zero class includes AdaGrad and RMSProp with zero additive stabilizers, Adam with $\epsilon=0$, signSGD~\citep{bernstein2018signsgd}, and Lion. This class also includes per-tensor preconditioned updates such as the zero-stabilizer limit of Shampoo~\citep{gupta2018shampoo}, with inverse roots taken on the ranges of the Gram factors. Adam and AdamW with $\epsilon>0$ are bounded near the zero history and also fall within the theorem's scope. These rules are not orbit-equivariant under a common learning rate, with or without decoupled weight decay.

\subsection{Soft Rebalancing: Contraction and Orbit Dependence}
\label{app:soft_contraction_moved}

\paragraph{Contraction of the log-ratio.}
\label{par:soft_contraction}
The residual log-imbalance after a single damped event contracts geometrically:
\begin{equation}
  \log \rho_p^{\mathrm{post}} \coloneqq \log \frac{\frobnorm{\mat{A}_p^{\mathrm{new}}}} {\frobnorm{\mat{B}_p^{\mathrm{new}}}} = (1 - \alpha)\log \rho_p,
  \label{eq:soft_contraction}
\end{equation}
so $K$ events with no intervening optimizer updates multiply the log-imbalance by $(1-\alpha)^K$. During training, optimizer updates can introduce new imbalance between events.

\paragraph{Dependence on the starting representative.}
\label{app:soft_detail}

The soft map is not itself an orbit canonicalizer for $\alpha<1$. Writing $\Phi_\alpha$ for the damped rebalancing map of~\eqref{eq:soft_saggs} and $T_c(\mat{A},\mat{B})\coloneqq(c\mat{A},c^{-1}\mat{B})$, direct substitution of $\rho(T_c(\mat{A},\mat{B}))=c^2\rho(\mat{A},\mat{B})$ gives
\begin{equation}
 \Phi_\alpha(T_c(\mat{A},\mat{B})) =T_{c^{1-\alpha}}(\Phi_\alpha(\mat{A},\mat{B})).
 \label{eq:soft_orbit_dependence}
\end{equation}
Thus one soft event retains dependence on the starting representative; away from balance the soft map is also not idempotent, since a second event changes $(1-\alpha)\log\rho$ to $(1-\alpha)^2\log\rho$. Hard rebalancing with $\alpha=1$ has both canonicalizer properties. Corollary~\ref{cor:canonical_restart} uses that hard map. The seven-seed Lion comparison in Sec.~\ref{sec:exp66_lion_soft_n7} evaluates the damped variant on Lion training.

\subsection{Pair-Norm Dynamics}
\label{app:balance_diagnostics_moved}

For one matrix pair, let $(\mat{U}^A_t,\mat{U}^B_t)$ denote the optimizer direction: the data gradient for gradient descent, the normalized moment direction for AdamW, or a sign vector for Lion. With $\eta_t>0$ and $0\le\eta_t\omega_t<1$, write
\begin{equation}
 \mat{A}_{t+1}\coloneqq(1-\eta_t\omega_t)\mat{A}_t-\eta_t\mat{U}^A_t,\qquad \mat{B}_{t+1}\coloneqq(1-\eta_t\omega_t)\mat{B}_t-\eta_t\mat{U}^B_t.
 \label{eq:pair_update_generic}
\end{equation}

Let $\Delta_t\coloneqq\frobnorm{\mat{A}_t}^2-\frobnorm{\mat{B}_t}^2$ be the balance defect, and let $r_t\coloneqq\log(\frobnorm{\mat{A}_t}/\frobnorm{\mat{B}_t})$ be the log-ratio underlying the experimental diagnostics wherever both norms are positive. The numerical stabilizer used for measurements is specified in Appendix~\ref{app:exp2_detail_moved}. For the optimizer direction in~\eqref{eq:pair_update_generic}, define the radial forcing and squared-direction norm difference by $F_t\coloneqq\langle\mat{A}_t,\mat{U}^A_t\rangle_{\mathrm{F}} -\langle\mat{B}_t,\mat{U}^B_t\rangle_{\mathrm{F}}$ and $Q_t\coloneqq\frobnorm{\mat{U}^A_t}^2-\frobnorm{\mat{U}^B_t}^2$.

\paragraph{Exact defect recursion.}
For any direction $(\mat{U}^A_t,\mat{U}^B_t)$ in~\eqref{eq:pair_update_generic}, expanding the squared norms gives
\begin{equation}
 \Delta_{t+1}=(1-\eta_t\omega_t)^2\Delta_t -2\eta_t(1-\eta_t\omega_t)F_t+\eta_t^2Q_t.
 \label{eq:defect_recursion}
\end{equation}
This algebraic recursion does not require a symmetry assumption.

\paragraph{From the balance defect to the log-ratio.}
\label{par:balance_log_ratio}
For positive norms, decay alone leaves $r_t$ unchanged. The balance defect bounds the log-ratio as $\abs{r_t}\le\abs{\Delta_t}/(2\min\{\frobnorm{\mat{A}_t}^2,\frobnorm{\mat{B}_t}^2\})$. Appendix~\ref{app:balance_dynamics_proof} derives this bound and the exact recursion.

\paragraph{Radial parameter and update components.}
Balancing equalizes the weight norms of a pair, not the norms of its data gradients. Even at Frobenius balance, whether at initialization or right after a SAGGS rebalance, the matrices $\mat{A}_\ell$ and $\mat{B}_\ell$ in a symmetry pair can receive gradients with different magnitudes due to their positions in the computation graph. For instance, $\mat{W}^Q_\ell$ receives gradients that depend on $\mat{W}^K_\ell$ through the attention logit Jacobian, and vice versa. Gradient magnitudes alone do not determine norm drift. Under \eqref{eq:pair_update_generic}, the expansion at a fixed iterate, with fixed directions and decay coefficient, is
\begin{equation}
 r_{t+1}-r_t =-\eta_t\left( \frac{\langle\mat{A}_t,\mat{U}^A_t\rangle_{\mathrm{F}}}{\frobnorm{\mat{A}_t}^2} -\frac{\langle\mat{B}_t,\mat{U}^B_t\rangle_{\mathrm{F}}}{\frobnorm{\mat{B}_t}^2} \right)+O(\eta_t^2).
 \label{eq:log_ratio_first_order}
\end{equation}
The common decay contribution cancels from this first-order log-ratio increment.

Appendix~\ref{app:balance_dynamics_proof} gives the exact logarithmic recursion. At Frobenius balance, with both squared norms equal to $N>0$, the first-order term is $-\eta_tF_t/N$. The corresponding exact balance-defect recursion is \eqref{eq:defect_recursion}. Interpreting a measured drift requires the radial components of the actual optimizer directions.

\paragraph{Common weight decay preserves the pair ratio.}
AdamW applies weight decay as $\theta \leftarrow (1 - \eta\omega) \theta$ uniformly to all parameters. This uniform decay multiplies both matrices of a pair by the same factor, so a decay step leaves the pair ratio unchanged and cannot restore balance by itself. The exact one-step defect recursion in \eqref{eq:defect_recursion} records the decay factor and data-update terms. The log-ratio increment depends on the radial components of the actual optimizer directions, as in~\eqref{eq:log_ratio_first_order}.

\subsection{Derivations for Pair-Norm Dynamics}
\label{app:balance_dynamics_proof}

All inner products and norms are Frobenius; we suppress the step index when no confusion can arise.

\paragraph{Derivation of the exact defect recursion.}
Expanding the squared norms of~\eqref{eq:pair_update_generic},
\begin{align*}
 \frobnorm{\mat{A}_{t+1}}^2 &=(1-\eta_t\omega_t)^2\frobnorm{\mat{A}_t}^2 -2\eta_t(1-\eta_t\omega_t)\langle\mat{A}_t,\mat{U}^A_t\rangle +\eta_t^2\frobnorm{\mat{U}^A_t}^2,\\
 \frobnorm{\mat{B}_{t+1}}^2 &=(1-\eta_t\omega_t)^2\frobnorm{\mat{B}_t}^2 -2\eta_t(1-\eta_t\omega_t)\langle\mat{B}_t,\mat{U}^B_t\rangle +\eta_t^2\frobnorm{\mat{U}^B_t}^2,
\end{align*}
and subtracting gives~\eqref{eq:defect_recursion}.

\paragraph{Decay and log-ratio diagnostics.}
For the positive block norms of Appendix~\ref{par:balance_log_ratio}, multiplying both matrices by the same positive scalar $1-\eta_t\omega_t$ leaves the ratio of their norms, hence $r_t$, unchanged.  For the bound on $r_t$, put $x\coloneqq\Delta_t/\frobnorm{\mat{B}_t}^2>-1$, so that $r_t=(1/2)\log(1+x)$.  If $x\ge0$ then $\frobnorm{\mat{B}_t}\le\frobnorm{\mat{A}_t}$ and $\log(1+x)\le x=\abs{\Delta_t}/\min\{\frobnorm{\mat{A}_t}^2,\frobnorm{\mat{B}_t}^2\}$. If $-1<x<0$ then $\frobnorm{\mat{A}_t}<\frobnorm{\mat{B}_t}$ and $-\log(1+x)=\log(1/(1+x))\le1/(1+x)-1 =-x/(1+x)=\abs{\Delta_t}/(\frobnorm{\mat{B}_t}^2+\Delta_t) =\abs{\Delta_t}/\frobnorm{\mat{A}_t}^2$, using $\log u\le u-1$.  In both cases, $\abs{r_t}\le\abs{\Delta_t}/(2\min\{\frobnorm{\mat{A}_t}^2,\frobnorm{\mat{B}_t}^2\})$.

\paragraph{Exact log-ratio increments.}
For positive block norms before and after the update, define $\nu_t^A\coloneqq\langle\mat{A}_t,\mat{U}^A_t\rangle_{\mathrm{F}}/ \frobnorm{\mat{A}_t}^2$ and $\chi_t^A\coloneqq\frobnorm{\mat{U}^A_t}^2/\frobnorm{\mat{A}_t}^2$, with the analogous definitions for $B$, and set $\kappa_t\coloneqq1-\eta_t\omega_t$. Dividing the squared-norm expansions above by the respective old squared norms gives
\begin{equation}
 r_{t+1}-r_t =\frac12\log\frac{ \kappa_t^2-2\eta_t\kappa_t\nu_t^A+\eta_t^2\chi_t^A}{ \kappa_t^2-2\eta_t\kappa_t\nu_t^B+\eta_t^2\chi_t^B}.
 \label{eq:log_ratio_increment}
\end{equation}
At a fixed iterate with fixed directions and decay coefficient, the two arguments have expansions $1-2\eta_t(\omega_t+\nu_t^A)+O(\eta_t^2)$ and $1-2\eta_t(\omega_t+\nu_t^B)+O(\eta_t^2)$. Using $\log(1+x)=x+O(x^2)$ proves \eqref{eq:log_ratio_first_order}. The terms involving $\omega_t$ cancel, and at balance $\nu_t^A-\nu_t^B=F_t/N$.

\paragraph{Numerical verification of the optimizer identities.}
A standalone numerical suite checks the AdamW state identities over 80 updates of a matrix regression with two coupled symmetry pairs, counterexamples obtained by omitting each required hyperparameter transformation, canonicalized restarts with common gradient clipping, and the zero-$\epsilon$ scalar counterexample. The numerical suite also checks the relative-increment exponents of Corollary~\ref{cor:half_compensation} and the exact defect recursion in \eqref{eq:defect_recursion}. These checks complement the proofs above.

\FloatBarrier
\section{Experimental and Implementation Details}
\label{app:exp_details}
\label{app:exp_setup_moved}

\subsection{Models and Training Protocols}

\paragraph{Architecture.}
The original small GPT-style experiments use Pre-LN, with normalization before each sub-layer, learned positional embeddings, tied input and output embeddings, bias-free linear projections, and context length 256. Later experiments specify their architecture and context-length changes. The vocabulary size $|\mathcal{V}|$ is 50,257, as determined by the GPT-2 byte-pair encoding (BPE) tokenizer. Linear and embedding weights in the GPT-style model use $\mathcal N(0,0.02^2)$; normalization gains retain their default initialization. These GPT-style runs apply no additional $1/\sqrt{2L}$ factor to residual-output projections.

The four original GPT-style configurations range from 22M to 354M parameters. Table~\ref{tab:model_configs} lists their widths, depths, head counts, and peak learning rates under the names small, medium, deep, and xlarge.

\paragraph{Training.}
We use AdamW~\citep{loshchilov2019decoupled} with $\beta_1 = 0.9$, $\beta_2 = 0.95$, $\epsilon = 10^{-8}$, weight decay $0.1$, and a cosine learning rate schedule with linear warmup over the first 2,000 steps and a minimum learning-rate ratio of 0.1. Gradient clipping is applied with max norm $1.0$. The default batch size is 64 for small and deep models and 32 for medium models. The xlarge configuration uses batches of 16 with gradient accumulation over 4 micro-batches, giving an effective batch size of 64. Training uses bfloat16 (BF16) mixed precision for efficiency.

\paragraph{Datasets.}
WikiText-103~\citep{merity2017pointer} is the primary dataset, with approximately 103M training words; GPT-2 BPE token counts and processed training-token budgets differ. OpenWebText~\citep{Gokaslan2019OpenWeb} is used for multi-dataset validation. OpenWebText contains approximately 8B tokens, and we use a subset for training.

\paragraph{SAGGS hyperparameters.}
\label{par:saggs_hyperparameters}
Hard SAGGS rebalances all symmetry pairs every 100 steps. Soft SAGGS uses the reported interval and $\alpha=0.5$ unless a sweep specifies otherwise. SAGGS+P, used only in some supplementary experiments, adds to SAGGS an auxiliary penalty on the training loss: the mean of $[\log(\frobnorm{\mat{A}}+\delta)-\log(\frobnorm{\mat{B}}+\delta)]^2$ over the rebalanced pairs, with $\delta\coloneqq10^{-8}$, weighted by $0.01$ unless a protocol specifies otherwise. The penalty is not part of SAGGS or Soft SAGGS. We rebalance Q--K, V--O, and MLP pairs in ReLU and SwiGLU models. GELU models use only Q--K and V--O pairs, since GELU up--down rescaling is not an exact symmetry.

\paragraph{Reparameterization procedure.}
The exponential reparameterization is applied after standard initialization by multiplying $\mat{A}_\ell \leftarrow e^{\lambda\ell}\mat{A}_\ell$ and $\mat{B}_\ell \leftarrow e^{-\lambda\ell}\mat{B}_\ell$ for each symmetry pair. For exact pairs, this rescaling preserves the initial forward pass in exact arithmetic across all $\lambda$ values. GELU MLP rescaling in the approximation diagnostic is excluded from this identity.

The mathematical sections index layers from $1$ to $L$; the training code uses $0,\ldots,L-1$, so its largest applied scale is $c_{L-1}=e^{\lambda(L-1)}$. Numerical examples use the code convention. Small and deep models use 30K steps, medium models 25K to 30K steps, and 354M models 15K to 20K steps unless their protocols specify otherwise. The logged gradients in SAGGS+P experiments include the auxiliary penalty; the exact gradient-coordinate law concerns the data loss alone.

\paragraph{Evaluation.}
Validation PPL is computed on the WikiText-103 validation set or the corresponding OpenWebText validation set, using non-overlapping windows at the experiment's configured sequence length, usually 256 or 512. Standard evaluation traverses the validation loader; any experiment-specific evaluation limits are specified in the corresponding protocol. Most experiments report mean $\pm$ SD over the paired seeds stated in each protocol; Appendix~\ref{app:seeds} lists the number of seeds in every experiment.

\subsection{Seeds and Computing Infrastructure}
\label{app:compute}

\paragraph{Random seeds.}
\label{app:seeds}
Paired comparisons use the same optimization seed for both arms, so the two arms share initialization, data order, and dropout masks. All seed pools were fixed before the corresponding experiments, and no seed was selected retrospectively from a larger pool.

A three-seed core pool was fixed before the multi-seed cross-scale comparisons and the three-seed 101M comparison. Three-seed experiments use this pool: the three-seed 101M comparison, the hard-rebalancing Lion comparison, the Lion state ablation, the OpenWebText replication, the 123.8M and 296.7M scale replications, the 354M stress test at $\lambda=0.2$, the 22M multi-seed cross-scale comparison, and the 101M one at $\lambda=0.2$. Four additional seeds, fixed before the primary seven-seed comparisons, extend the core pool to seven seeds for the seven-seed 101M comparison, the 31M architecture and tokenizer replication, and the soft-rebalancing Lion comparison. The branched checkpoint-surgery resume and the optimizer-state diagnostics use a separate seven-seed pool, which also contains the core pool and was fixed before those experiments. Two seeds from the core pool are used in the natural-drift experiment, the deep-model stress test, the 101M multi-seed cross-scale comparison at $\lambda=0$, and the 354M one. A single seed from the core pool is used in the 354M stress test at $\lambda\in\{0,0.1\}$, the SwiGLU stress test, the 101M rescaling-strength sweep, and the 1.2B training comparison.

\paragraph{Compute.}
The main experimental suite was run on NVIDIA H100 80GB GPUs, using 4$\times$H100 single-node runs with distributed data parallel (DDP) training for the largest configurations. The implementation also supports Intel Data Center GPU Max devices through the PyTorch XPU backend; DDP runs use NCCL on CUDA devices and XCCL on XPU devices.  The 31M architecture and tokenizer replication, the 123.8M scale replication and branched checkpoint-surgery resume, the 31M optimizer-state diagnostics, and the 296.7M scale replication were executed on Aurora Intel Data Center GPU Max 1550 XPU tiles. Runtime depends on the recipe and hardware. For example, the 1.2B training comparison records 2,494 to 2,518 seconds per configuration on four H100s, equivalent to 41.6 to 42.0 minutes or approximately 2.8 H100-hours per configuration. These values are training timers, including periodic evaluation, rather than end-to-end times including data and model setup; these timers should not be extrapolated from parameter count alone.

\subsection{Implementation}
\label{app:implementation}

\paragraph{Implementation checks.}
We verify the implementation on CPU and replay the same suite on Aurora using an XPU device. The checks confirm exact forward preservation, the closed-form gradient law, and SAGGS forward preservation for Q--K, V--O, and MLP pairs on both the GPT-style model and the RMSNorm, RoPE, SwiGLU, and GQA model of the architecture and tokenizer replication.  The checks also verify the optimizer-state co-transform used in the Lion state ablation by confirming that Adam-family first- and second-moment buffers receive the chain-rule scale factors after a hard SAGGS orbit move and after an explicit exact reparameterization surgery.  Both runs pass all checks: The largest relative gradient-law error is below $3\times10^{-7}$, and the optimizer-state relative error is zero.

\paragraph{Reference implementation.}
Listing~\ref{lst:saggs_operator} reproduces the rebalancing operator of our implementation, trimmed to the bias-free projections used in the primary training experiments.  The pair map follows Table~\ref{tab:symmetry_catalogue}: For SwiGLU the MLP pair is the value and output matrices and the gate is untouched, and GELU models use only the Q--K and V--O pair types.  For each pair $p$ the function computes $\rho_p=\frobnorm{\mat{A}_p}/\frobnorm{\mat{B}_p}$ and applies the damped rebalancing map of \eqref{eq:soft_saggs} with $c^{\mathrm{eff}}_p(\alpha)=\rho_p^{\alpha/2}$; $\alpha=1$ gives the hard rebalancing of Algorithm~\ref{alg:rebalancing}.  Pairs with a norm below $\delta_{\rm skip}=10^{-12}$ are skipped, as analyzed in \eqref{eq:guarded_soft_event}.  The optional moment co-transform is the state-covariant variant of Appendix~\ref{app:statecov_detail}, used only in the Lion state ablation and the optimizer-state diagnostics; default hard and Soft SAGGS leave the buffers unchanged.  For affine projections the full implementation additionally applies the bias co-transforms of Appendix~\ref{app:notation_setup_moved}.  The training loop calls the operator on the 32-bit floating-point (FP32) parameters, outside the BF16 autocast region, immediately after the optimizer update whenever $t \bmod T_{\mathrm{bal}}=0$.

\needspace{8\baselineskip}
\begin{lstlisting}[caption={Hard SAGGS with \texttt{alpha=1} and Soft SAGGS with \texttt{alpha<1} for bias-free projections, with optional optimizer-state co-transformation.},label={lst:saggs_operator}]
import torch


def get_pair_modules(block, pair_type):
    """Modules (A, B) of one exact pair in a block (Table 1)."""
    if pair_type == 'qk':
        return block.attn.W_Q, block.attn.W_K
    if pair_type == 'vo':
        return block.attn.W_V, block.attn.W_O
    if pair_type == 'mlp':
        if hasattr(block.mlp, 'W1'):       # ReLU up and down
            return block.mlp.W1, block.mlp.W2
        if hasattr(block.mlp, 'W_value'):  # SwiGLU value and output; gate kept
            return block.mlp.W_value, block.mlp.W_out
    return None, None


def rescale_optimizer_state(optimizer, param, grad_scale):
    """Buffer co-transform for param <- s * param, grad_scale = 1 / s:
    gradient averages scale linearly, squared-gradient averages quadratically."""
    state = optimizer.state.get(param)
    if not state:
        return
    for key, buf in state.items():
        if not torch.is_tensor(buf) or buf.shape != param.shape:
            continue
        if key == 'exp_avg':        # Adam and AdamW first moment; Lion momentum
            buf.mul_(grad_scale)
        elif key == 'exp_avg_sq':   # Adam and AdamW second-moment estimate
            buf.mul_(grad_scale * grad_scale)


@torch.no_grad()
def saggs_rebalance_soft(model, pair_types=('qk', 'vo', 'mlp'),
                         alpha=0.5, optimizer=None,
                         transform_optimizer_state=False):
    """Soft SAGGS with alpha in (0, 1]; alpha = 1 is Algorithm 1."""
    for block in model.blocks:
        for pt in pair_types:
            A_mod, B_mod = get_pair_modules(block, pt)
            if A_mod is None:
                continue
            A, B = A_mod.weight, B_mod.weight
            norm_A, norm_B = A.norm(), B.norm()   # Frobenius norms
            if norm_A < 1e-12 or norm_B < 1e-12:  # delta_skip guard
                continue
            c_eff = (norm_A / norm_B).pow(alpha * 0.5)  # rho^(alpha/2)
            A.div_(c_eff)                        # A <- A / c_eff
            B.mul_(c_eff)                        # B <- c_eff * B
            if transform_optimizer_state:        # state-covariant only
                rescale_optimizer_state(optimizer, A, c_eff)
                rescale_optimizer_state(optimizer, B, c_eff.reciprocal())


def saggs_rebalance(model, pair_types=('qk', 'vo', 'mlp')):
    """Hard SAGGS (Algorithm 1): balanced orbit representative."""
    saggs_rebalance_soft(model, pair_types, alpha=1.0)
\end{lstlisting}

\subsection{Model Configurations}

Table~\ref{tab:model_configs} summarizes the architectures and peak learning rates. Experiment-specific protocols specify any departures from these defaults.

The GPT-2-style models use context length 256 unless otherwise stated; the xlarge, 1B, and later 101M replication recipes use 512. At the same widths, the SwiGLU stress test uses three-projection variants with 25.5M parameters at $L=12$ and 63.3M at $L=48$. Modern-architecture models use RMSNorm, RoPE, SwiGLU, GQA, and T5 SentencePiece tokenization. Their context length is 256 for the 31M model in the architecture and tokenizer replication and the optimizer-state diagnostics, and 512 for the larger models.

\begin{table}[t!]
\centering
\caption{Model configurations. LR is the peak learning rate; modern-architecture head counts are pairs $(h_q,h_{kv})$ of query heads and key-value (KV) heads. The first four parameter counts assume two-projection ReLU or GELU MLPs. Context lengths and SwiGLU variants are specified in the text.}
\label{tab:model_configs}
\vspace{4pt}
\small
\begin{tabular}{@{}lcccccc@{}}
\toprule
\textbf{Name} & $d$ & $d_{\mathrm{ff}}$ & $L$ & \textbf{Heads} & \textbf{Params} & \textbf{LR} \\
\midrule
\multicolumn{7}{@{}l}{GPT-2-style} \\
Small        & 256  & 1024  & 12 & 4  & 22M  & $6\times10^{-4}$ \\
Medium       & 512  & 2048  & 24 & 8  & 101M & $6\times10^{-4}$ \\
Deep         & 256  & 1024  & 48 & 4  & 51M  & $3\times10^{-4}$ \\
Xlarge       & 1024 & 4096  & 24 & 16 & 354M & $3\times10^{-4}$ \\
1B (SwiGLU)  & 2048 & 5504  & 22 & 16 & 1.22B & $3\times10^{-4}$ \\
\midrule
\multicolumn{7}{@{}l}{Modern architecture} \\
Modern-small & 384 & 1024 & 12 & $(6,2)$ & 31M & $4\times10^{-4}$ \\
Modern-medium-plus & 640 & 1728 & 24 & $(10,2)$ & 123.8M & $2.5\times10^{-4}$ \\
Modern-large & 896 & 2432 & 32 & $(14,2)$ & 296.7M & $2\times10^{-4}$ \\
\midrule
\multicolumn{7}{@{}l}{Pre-trained} \\
GPT-2 (pre-trained) & 768 & 3072 & 12 & 12 & 124M & $2\times10^{-5}$ \\
\bottomrule
\end{tabular}
\end{table}

\subsection{Optimizer-State Co-Transformation}
\label{app:statecov_detail}

\paragraph{Optimizer-state co-transform ablation.}
\label{par:state_covariant_saggs}
SAGGS is a coordinate change on the symmetry orbit, whereas practical optimizers also maintain state variables.  To separate the effect of the orbit move from the possible perturbation of optimizer state, we also evaluate a state-covariant variant.  If a rebalance applies $\mat{A}'\coloneqq\mat{A}/c$ and $\mat{B}'\coloneqq c\mat{B}$, then the data-gradient coordinates transform as $\grad_{\mat{A}'}\loss=c\grad_{\mat{A}}\loss$ and $\grad_{\mat{B}'}\loss=c^{-1}\grad_{\mat{B}}\loss$. Accordingly, for an Adam-family first-moment buffer or a Lion momentum buffer $(\mat{m}_A,\mat{m}_B)$ we map $\mat{m}'_A\coloneqq c\mat{m}_A$ and $\mat{m}'_B\coloneqq c^{-1}\mat{m}_B$, and for Adam-family second-moment estimates $(\mat{v}_A,\mat{v}_B)$ we map $\mat{v}'_A\coloneqq c^2\mat{v}_A$ and $\mat{v}'_B\coloneqq c^{-2}\mat{v}_B$. By default, both hard and Soft SAGGS leave the optimizer buffers unchanged, so neither method needs access to the internal state of the optimizer. We use the state-covariant variant only as a diagnostic: In the Lion state ablation (Appendix~\ref{app:exp64_moved}), this variant measures how leaving Lion's momentum buffer in the old coordinates affects the cost of hard rebalancing.

\subsection{Cost, Weight Decay, and Numerical Guards}
\label{app:practical_full}

\paragraph{Computational cost.}
Each rebalance computes two Frobenius norms and rescales all entries, costing $O(n_A+n_B)$ per pair with $O(1)$ tensor operations. At interval $T_{\mathrm{bal}}$, this rebalancing adds amortized work of $O((n_A+n_B)/T_{\mathrm{bal}})$ per pair per optimization step. With full-layer Q--K, V--O, and MLP pairs there are $3L$ pairs; independent headwise Q--K and V--O pairs give $(2H+1)L$ for standard $H$-head attention.

\paragraph{Interaction with weight decay.}
A common decoupled decay factor multiplies both matrices equally and leaves their norm ratio unchanged~\citep{zhang2019three}. The one-step balance-defect recursion in \eqref{eq:defect_recursion} records the contribution of this decay factor; we discuss the measured drift in Appendix~\ref{sec:disc_natural_drift}.

\paragraph{Numerical skip guard.}
The implementation skips a pair when either norm is below $\delta_{\rm skip}\coloneqq10^{-12}$, whereas Algorithm~\ref{alg:rebalancing} states the ideal positive-norm map. For a scheduled soft event $k$ with positive norms, let $r_k\coloneqq\log(\frobnorm{\mat{A}}/\frobnorm{\mat{B}})$ before the event and let $z_k$ be the log-ratio afterward. The implemented correction obeys, in exact arithmetic,
\begin{equation}
 \alpha_k^{\rm eff}\coloneqq\alpha\mathbf1\{ \min\{\frobnorm{\mat{A}},\frobnorm{\mat{B}}\}\geq\delta_{\rm skip}\}, \qquad z_k=(1-\alpha_k^{\rm eff})r_k.
 \label{eq:guarded_soft_event}
\end{equation}
A skipped positive-norm pair has $z_k=r_k$, so the hard canonicalizer and fixed-$\alpha$ contraction identities apply to executed corrections.

\subsection{Code Availability and Reproducibility}
\label{app:code_availability}
Appendix~\ref{app:proofs} gives self-contained proofs under the stated assumptions. The experimental protocols and table captions specify model configurations, datasets, optimizers, training budgets, schedules, and hardware. Appendix~\ref{app:seeds} describes the seed protocol and seed counts, and Appendix~\ref{app:stat_details} defines the paired statistics and multiple-comparison conventions.

Appendix~\ref{app:implementation} gives a reference implementation of the rebalancing operator (Listing~\ref{lst:saggs_operator}) and the checks used to verify this implementation. WikiText-103 and OpenWebText are public datasets.

Reproduction uses fixed seeds and the documented hardware and backend. Because \path{torch.nn.functional.scaled_dot_product_attention} can introduce backend non-determinism, reruns can differ from the reported values within numerical tolerance.

\FloatBarrier
\section{Statistical Methodology and Summary of Results}
\label{app:moved_details}

The primary comparisons and checkpoint-resume experiments use seven paired seeds; OpenWebText and the larger modern architectures use three. We report the mean paired PPL difference $\bar\Delta$, its sample SD $\sigma_\Delta$, paired Cohen's $d$, and exact sign-flip $p$-values. Unless stated otherwise, a difference is Vanilla minus the method, so a positive value favors the method, and $\pm$ denotes a sample SD. With seven seeds, the one-sided sign-flip $p$-value $p_{\mathrm{sf}}^{+}$ takes its smallest possible value, $0.0078$, exactly when all seven differences are positive; with three seeds, its smallest value is $0.125$.

\subsection{Paired Comparisons}
\label{app:stat_pairing_moved}

For every paired comparison of Vanilla with a method at fixed $\lambda$ and fixed model size, we use the same optimization seed $s$ for both arms, covering data ordering, weight initialization, and dropout mask, so the paired difference $\Delta^{(s)} \coloneqq M^{\mathrm{Vanilla}}(s) - M^{\mathrm{Method}}(s)$ controls these stochastic factors. Checkpoint-branch diagnostics share the base checkpoint and use the continuation protocol specified in the corresponding experiment. The within-seed design reduces variation due to these shared stochastic factors. The primary 101M WikiText-103, 31M architecture replication, and Lion comparisons use seven paired seeds (Appendix~\ref{app:seeds}).

\subsection{Statistics and Sign-Flip Tests}
\label{app:stat_details}

\paragraph{Reported statistics.}
The paired summaries use the mean $\bar\Delta$, the paired SD $\sigma_\Delta \coloneqq [(1/(n-1))\sum_s (\Delta^{(s)} - \bar\Delta)^2]^{1/2}$, and paired Cohen's $d \coloneqq \bar\Delta / \sigma_\Delta$. Tables report the subset of these statistics relevant to each comparison. Unless explicitly labeled otherwise, primary $\pm$ values are sample SDs, not standard errors. Earlier small-$n$ tables use a different dispersion convention: Those summaries use population SDs, while the later paired analyses use the $n-1$ denominator. We do not pool these summaries. Differences are computed before rounding; subtracting two rounded means can differ by one last rounded digit from the reported paired difference. Later table captions state only experiment-specific details. Unless a caption specifies otherwise, a positive paired $\Delta$ is the left-minus-right or Vanilla-minus-method contrast on validation PPL, and $p_{\rm sf}^{+}$ is the one-sided sign-flip $p$-value for a positive mean.

\paragraph{Sign-flip tests.}
For finite-sample significance, we use exact paired sign-flip randomization tests as the primary $p$-value whenever the claim is a paired $n=7$ contrast. Their finite-sample validity requires invariance of the joint null distribution under independent sign flips of the paired differences. For example, the condition holds for independent differences symmetric about zero. Pairing seeds or assuming only a zero mean does not establish this invariance. At $n=7$, the smallest one-sided sign-flip $p$ is $1/2^7=0.0078$; at $n=3$ the $p$-value floor is only $0.125$, so three-seed comparisons can establish effect direction and scale but not a $p<0.05$ sign-flip claim.  We also report the paired statistic $t\coloneqq\sqrt{n}\bar\Delta/\sigma_\Delta$ as a supplementary descriptive summary; significance claims use the sign-flip test.

For paired differences $\Delta_1,\ldots,\Delta_n$, we enumerate all $2^n$ sign assignments and compute the one-sided value
\begin{align*}
  p_{\rm sf}^{+}&\coloneqq2^{-n}\sum_{\boldsymbol\sigma\in\{\pm1\}^n} \mathbf1\left\{\frac{1}{n}\sum_i\sigma_i\Delta_i \geq\bar\Delta\right\}.
\end{align*}
The two-sided zero-effect test compares the absolute permuted mean against $|\bar\Delta|$. These tests do not establish equivalence against a prespecified practical margin.

\paragraph{Effect-size thresholds.}
We use the conventional Cohen thresholds adapted to the paired design~\citep{cohen1988statistical,lakens2013calculating}: $|d| < 0.2$ negligible, $0.2 \leq |d| < 0.5$ small, $0.5 \leq |d| < 0.8$ medium, $|d| \geq 0.8$ large, $|d| \geq 2.0$ very large. The PPL effect at $\lambda = 0.2$, $n = 7$ on the 101M medium model is $d \approx +14.2$ in the primary comparison (Sec.~\ref{sec:exp65_core_n7} and Table~\ref{tab:exp65_core_n7}), with all seven paired differences positive. We interpret paired effect sizes together with absolute metric changes and sample size, rather than relying on the $p$-value alone.

\paragraph{Multiple-comparison handling.}
The primary hypothesis that SAGGS protects the language-model objective under a forward-equivalent $\lambda=0.2$ orbit imbalance is tested by the 101M WikiText-103 paired seven-seed contrast in Sec.~\ref{sec:exp65_core_n7}.  The 31M modern-architecture seven-seed replication and the Lion and Soft SAGGS seven-seed comparison are primary replications of the architecture, tokenizer, and optimizer variants. Secondary contrasts, including other $\lambda$ values, scale variants, and workflow diagnostics, provide supporting evidence. Reported $p$-values are unadjusted unless labeled otherwise; the unadjusted $p$-values must not be interpreted as family-wise significance claims. The single pre-declared core PPL contrast is separate from these secondary families.

\paragraph{Single-seed 1.2B comparison.}
The 1.2B comparison in Appendix~\ref{sec:exp_1b_training} uses a single seed because of compute constraints. Its training timers are 2,494 to 2,518 seconds per configuration on four H100s, or approximately 2.8 H100-hours per configuration, excluding data and model setup. We report this comparison's observed differences without an inferred $p$-value or Cohen's $d$ (Table~\ref{tab:1b_training}). The seven-seed 101M, modern-architecture, Lion, and checkpoint-resume comparisons are the primary statistical evidence; the three-seed 123.8M and 296.7M comparisons test transfer across scales.

\paragraph{Pre-specification of $\lambda$ and seed choices.}
Throughout the paper, pre-specified means fixed in our experiment plan before inspecting the corresponding held-out result; the phrase does not imply external public registration. The stress-test setting $\lambda=0.2$ was selected from the empirical sensitivity observed in initial single-seed exploratory runs and then fixed in advance for all subsequent paired confirmation experiments with $\lambda>0$. The paired seed pools were likewise fixed before the corresponding experiments were executed (Appendix~\ref{app:seeds}). The primary paired tables include all seeds of the corresponding pools and the $\lambda$ values fixed in those configurations.

\subsection{Summary of Primary and Supporting Comparisons}
\label{app:evidence_tiers_moved}

\paragraph{Role of each comparison.}
The main protection result is $\bar\Delta_{\mathrm{PPL}}=+0.374\pm0.026$ at $\lambda=0.2$, $n=7$, with paired Cohen's $d=14.21$ (Table~\ref{tab:exp65_core_n7}), measured under deliberate reparameterization. At $\lambda=0$, mean differences are small in the core AdamW and selected Soft and Lion comparisons; the three-seed Lion state ablation (Appendix~\ref{app:exp64_moved}) is the main adverse unperturbed case.

SAGGS and Soft SAGGS are orbit-rebalancing steps for exact bilinear symmetries, complementary to the optimizer: Hard SAGGS selects the balanced representative, and Soft SAGGS contracts the distance to that representative in log-ratio coordinates. Deliberate rescaling tests measure sensitivity to orbit position; checkpoint-surgery and stale-state branches test recovery after an edit. Table~\ref{tab:artifact_evidence_tiers} records the statistical support of each comparison. Paired $n=7$ comparisons are the primary evidence; the $n=3$ modern-architecture comparisons have a sign-flip floor of $0.125$, and the 1.2B run is a single-seed scale check.

\begin{table}[t!]
\centering
\caption{Primary PPL comparisons and checkpoint-resume diagnostics. $\bar\Delta$ is the paired left-minus-right difference, reported as mean $\pm$ sample SD. Rows marked $^{\dagger}$ use three paired seeds; all others use seven. Soft denotes Soft SAGGS, WT-103 denotes WikiText-103, and V denotes Vanilla.}
\label{tab:artifact_evidence_tiers}
\small
\begin{tabular}{@{}lclrcl@{}}
\toprule
\textbf{Setting} & $\lambda$ & \textbf{Contrast} & $\bar\Delta$ & $p_{\mathrm{sf}}^{+}$ & \textbf{Purpose} \\
\midrule
\multicolumn{6}{@{}l}{Primary} \\
101M WT-103 & $0.2$ & V $-$ SAGGS & $+0.374\pm0.026$ & $0.0078$ & Core language-model protection \\
31M modern & $0.2$ & V $-$ Soft & $+0.780\pm0.040$ & $0.0078$ & Architecture and tokenizer \\
101M Lion & $0.2$ & V $-$ Soft & $+34.465\pm25.170$ & $0.0078$ & Optimizer transfer \\
\midrule
\multicolumn{6}{@{}l}{Workflow} \\
123.8M surgery & & No-SAGGS $-$ control & $+3.775\pm0.416$ & $0.0078$ & State-reset resume \\
\midrule
\multicolumn{6}{@{}l}{Mechanism} \\
31M reset & & Reset $-$ reset ctrl & $+0.277\pm0.015$ & $0.0078$ & Reset penalty isolated \\
31M reset & & Soft $-$ reset ctrl & $+0.007\pm0.006$ & $0.0156$ & Residual after Soft \\
31M stale & & Stale $-$ Soft & $+0.062\pm0.016$ & $0.0078$ & Stale-state recovery \\
31M partial & & Deep-stale $-$ Soft & $+0.062\pm0.008$ & $0.0078$ & Partial-state recovery \\
\midrule
\multicolumn{6}{@{}l}{Scale transfer} \\
123.8M modern$^{\dagger}$ & $0.2$ & V $-$ Soft & $+21.156\pm0.667$ & $0.1250$ & Larger modern architecture \\
296.7M modern$^{\dagger}$ & $0.2$ & V $-$ Soft & $+85.813\pm4.730$ & $0.1250$ & 300M-class modern \\
\bottomrule
\end{tabular}
\end{table}

\FloatBarrier
\section{Primary Comparisons and Optimizer-State Ablations}
\label{app:moved_round2}

The subsections below group related comparisons by purpose.

\FloatBarrier
\subsection{101M WikiText-103 Comparisons}
\label{app:exp45_n3_moved}

The three-seed comparison uses a paired $2\times2\times3$ design on WikiText-103: $\lambda\in\{0,0.2\}$, Vanilla or SAGGS, and three paired seeds. The medium ReLU Transformer has $L=24$, $d=512$, 8 heads, about 101M parameters, and context length 512. Each run uses 30K AdamW steps, learning rate $6\times10^{-4}$, $\beta_2=0.95$, weight decay $0.1$, 2K-step warmup, cosine decay to 10\% of peak, gradient clipping at 1.0, BF16 mixed precision, and batch size 32. At $\lambda=0.2$, Q--K pairs receive the depth-exponential edit at initialization. SAGGS rebalances Q--K, V--O, and ReLU up--down pairs every 100 steps. Data sampling and initialization are shared within seed. The primary comparison in Table~\ref{tab:exp65_core_n7} extends this protocol to seven paired seeds.

Table~\ref{tab:exp45_ppl_moved} reports paired differences $\Delta_s=\mathrm{PPL}_{\rm Vanilla}(s)-\mathrm{PPL}_{\rm SAGGS}(s)$. At $\lambda=0$, the mean is $-0.044\pm0.026$ PPL: a small observed cost on a base near 22.9 PPL. At $\lambda=0.2$, the mean is $+0.380\pm0.025$ PPL and all three seeds favor SAGGS. The paired effect sizes are $d=-1.69$ and $d=15.35$, respectively. With $n=3$, the one-sided exact sign-flip $p$-value floor is $0.125$.

\begin{table}[t!]
\centering
\caption{Three-seed 101M WikiText-103 comparison. Table~\ref{tab:exp65_core_n7} gives the seven-seed extension. Method columns report PPL as mean $\pm$ sample SD; $\bar\Delta$ reports paired Vanilla-minus-SAGGS differences in the same format.}
\label{tab:exp45_ppl_moved}
\vspace{4pt}
\small
\begin{tabular}{@{}lccc@{}}
\toprule
$\lambda$ & Vanilla & SAGGS & $\bar\Delta$ \\
\midrule
$0.0$ & $22.88\pm0.060$ & $22.92\pm0.033$ & $-0.044\pm0.026$ \\
$0.2$ & $23.25\pm0.024$ & $22.87\pm0.036$ & $+0.380\pm0.025$ \\
\bottomrule
\end{tabular}
\end{table}

\paragraph{Pair-norm measurements in the seven-seed comparison.}
\label{app:geometric_moved}
SAGGS keeps the final pair log-ratio imbalance below $10^{-3}$ in these comparisons, whereas Vanilla Q--K imbalance is about $0.15$ at $\lambda=0$ and about $3$ at $\lambda=0.2$. Rebalancing controls pair norms in both conditions; the PPL benefit appears under deliberate rescaling.

\FloatBarrier
\subsection{Architecture and Tokenizer Replication}
\label{app:exp63_moved}

This experiment tests transfer across architecture and tokenizer choices. The experiment replaces the GPT-style layer normalization, learned absolute positions, and standard multi-head attention with RMSNorm, RoPE, SwiGLU, and GQA, and replaces GPT-2 BPE with T5 SentencePiece tokenization. We evaluate the resulting 31M model over seven paired seeds.

We train a 12-layer causal language model with $d=384$, six query heads, two KV heads, SwiGLU hidden width 1024, tied input and output embeddings, no biases, RMSNorm pre-normalization, RoPE, and T5 SentencePiece vocabulary.  We re-tokenize WikiText-103 with the T5 tokenizer.

The optimizer is AdamW with $\eta=4\times10^{-4}$, weight decay $0.1$, cosine decay, 10\% warmup, gradient clipping at $1.0$, BF16 XPU autocast, batch size 48, sequence length 256, and 12K steps.  The paired design compares Vanilla with Soft SAGGS at $\alpha=0.5$, crossed with $\lambda=0$ and $\lambda=0.2$, over seven paired seeds. The reparameterization acts on Q--K, V--O, and SwiGLU value--output pairs, and Soft SAGGS rebalances those same pairs every 100 steps. A pre-training verification on a WikiText-103 batch confirms exact forward equivalence, with loss difference $0$ and maximum logit difference below $2.4\times10^{-6}$, and gradient-law relative errors below $5\times10^{-7}$ for Q--K, V--O, and SwiGLU value--output pairs.

The results reproduce both patterns observed in the 101M comparison (Table~\ref{tab:exp63_modern_n7_cells}). First, the unperturbed $\lambda=0$ comparison has a small observed difference: Vanilla and Soft SAGGS differ by only $+0.019\pm0.024$ PPL, taking Vanilla minus Soft, relative to a baseline near 20.8 PPL.  Second, the $\lambda=0.2$ reparameterization creates a consistent degradation in Vanilla on all seven seeds, while Soft SAGGS remains near its unperturbed baseline: Vanilla is $21.58\pm0.07$ PPL and Soft SAGGS is $20.80\pm0.06$ PPL, with paired $\bar\Delta=+0.780\pm0.040$, $t=52.18$, Cohen $d=19.72$, and exact one-sided sign-flip $p=0.0078$.

The final pair-norm statistics show corresponding balance control: At $\lambda=0.2$, Vanilla ends with Q--K max absolute log-ratio about $2.65$, V--O about $1.21$, and MLP about $1.81$, while Soft SAGGS keeps all three below $8\times10^{-4}$.  The scale replications in Appendices~\ref{sec:exp67_modern_medium} and~\ref{sec:exp69_modern_large} extend the same architecture to 123.8M and 296.7M parameters. PPL depends on the tokenizer, so these values are not comparable with the values of the GPT-2-tokenized models; all performance comparisons here use the same tokenizer and paired seeds.

\begin{table}[t!]
\centering
\caption{Seven-seed modern architecture and cross-tokenizer replication. The paired block is Vanilla minus Soft SAGGS with $\alpha=0.5$.}
\label{tab:exp63_modern_n7_cells}
\label{tab:exp63_modern_n7_paired}
\small
\begin{tabular}{@{}lccccccc@{}}
\toprule
& \multicolumn{2}{c}{\textbf{Mean PPL}} & \multicolumn{5}{c}{\textbf{Paired difference}} \\
\cmidrule(lr){2-3}\cmidrule(lr){4-8}
$\lambda$ & Vanilla & Soft, $\alpha=0.5$ & $\bar\Delta$ & $\sigma_\Delta$ & $t$ & $d$ & $p_{\mathrm{sf}}^{+}$ \\
\midrule
$0.0$ & $20.83\pm0.06$ & $20.81\pm0.06$ & $+0.019$ & $0.024$ & $+2.11$ & $+0.80$ & $0.0391$ \\
$0.2$ & $21.58\pm0.07$ & $20.80\pm0.06$ & $+0.780$ & $0.040$ & $+52.18$ & $+19.72$ & $0.0078$ \\
\bottomrule
\end{tabular}
\end{table}

\FloatBarrier
\subsection{Hard Rebalancing with Lion}
\label{app:exp41_moved}

Lion~\citep{chen2023symbolic} uses sign-based updates. Positive-scale invariance of its sign direction does not imply orbit equivariance of its parameter trajectory (Theorem~\ref{thm:equivariant_rules}). We test whether hard SAGGS protects Lion training under an exact orbit tilt.

\paragraph{Design.}
The protocol uses the medium ReLU Transformer with $L=24$, $d=512$, 8 heads, and 101M parameters, WikiText-103 with GPT-2 BPE, context length 512, and three paired seeds. We compare Vanilla and SAGGS at $\lambda=0.2$, applied to Q--K, V--O, and ReLU up--down pairs at initialization. Training uses 15K Lion steps with learning rate $2\times10^{-4}$, $\beta=(0.9,0.99)$, weight decay $0.1$, batch size 32, gradient clipping at 1.0, and BF16 mixed precision. The cosine schedule uses 1.5K warmup steps and decays to 10\% of peak learning rate. SAGGS rebalances all three pair types every 100 steps.

\paragraph{Results.}
All three $\lambda=0.2$ pairs favor SAGGS, with $\bar\Delta_{\mathrm{Vanilla-SAGGS}}=+52.896\pm31.543$ PPL, paired Cohen's $d=1.68$, and paired $t=2.90$ (Table~\ref{tab:exp41_lion}). With $n=3$, the one-sided exact sign-flip $p$-value floor is $0.125$. The soft-rebalancing comparison in Appendix~\ref{app:exp66_moved} evaluates Soft SAGGS with seven paired seeds under both unperturbed and deliberately tilted starts. The Lion state ablation in Appendix~\ref{app:exp64_moved} compares hard, soft, and state-covariant rebalancing under matched Lion controls.

Table~\ref{tab:exp41_lion} records the Lion validation PPL of each paired seed in this comparison.

\begin{table}[t!]
\centering
\caption{Three-seed hard-rebalancing Lion comparison at $\lambda=0.2$: 101M parameters, $L=24$, and 15K steps. Rows index the paired seeds. V denotes Vanilla, S denotes SAGGS, and $\Delta=\mathrm{PPL}_{\mathrm{V}}-\mathrm{PPL}_{\mathrm{S}}$.}
\label{tab:exp41_lion}
\small
\begin{tabular}{@{}crrr@{}}
\toprule
\textbf{Seed} & $\mathrm{PPL}_{\mathrm{V}}$ & $\mathrm{PPL}_{\mathrm{S}}$ & $\Delta$ \\
\midrule
        1 & $55.992$ & $23.954$ & $+32.038$ \\
        2 & $113.248$ & $24.065$ & $+89.183$ \\
        3 & $61.429$ & $23.963$ & $+37.466$ \\
\bottomrule
\end{tabular}
\end{table}

\FloatBarrier
\subsection{Soft Rebalancing with Lion}
\label{app:exp66_moved}

\paragraph{Design.}
We use the 101M ReLU Transformer and 15K-step Lion recipe of the hard-rebalancing comparison: learning rate $2\times10^{-4}$, $\beta=(0.9,0.99)$, weight decay $0.1$, 1.5K-step warmup, cosine decay to 10\% of peak learning rate, batch size 32, context length 512, gradient clipping at 1.0, and BF16 mixed precision on WikiText-103. The paired design compares Vanilla and Soft SAGGS at $\lambda\in\{0,0.2\}$, with all-pairs reparameterization and Soft $\alpha=0.5$ every 100 steps. The hard and state-covariant comparisons in the Lion state ablation use three paired seeds.

\begin{table}[t!]
\centering
\caption{Seven-seed Lion comparison on the 101M WikiText-103 model. PPL is reported as mean $\pm$ sample SD; the paired block reports Vanilla-minus-Soft SAGGS differences.}
\label{tab:exp66_lion_soft_n7_cells}
\label{tab:exp66_lion_soft_n7_paired}
\small
\begin{tabular}{@{}lccccccc@{}}
\toprule
& \multicolumn{2}{c}{\textbf{Mean PPL}} & \multicolumn{5}{c}{\textbf{Paired difference}} \\
\cmidrule(lr){2-3}\cmidrule(lr){4-8}
$\lambda$ & Vanilla & Soft SAGGS & $\bar\Delta$ & $\sigma_\Delta$ & $t$ & $d$ & $p_{\mathrm{sf}}^{+}$ \\
\midrule
$0.0$ & $24.15\pm0.39$ & $24.20\pm0.58$ & $-0.055$ & $0.232$ & $-0.62$ & $-0.24$ & $0.6953$ \\
$0.2$ & $58.32\pm25.20$ & $23.86\pm0.09$ & $+34.465$ & $25.170$ & $+3.62$ & $+1.37$ & $0.0078$ \\
\bottomrule
\end{tabular}
\end{table}

Table~\ref{tab:exp66_lion_soft_n7_cells} provides the seven-seed summaries. At $\lambda=0.2$, every one of the seven paired seeds favors Soft SAGGS, giving $\bar\Delta_{\rm Vanilla-Soft}=+34.465$ PPL and exact one-sided sign-flip $p_{\rm sf}^{+}=0.0078$.  At $\lambda=0$, the paired mean is $-0.055\pm0.232$ PPL, including one seed where Soft is $0.51$ PPL worse than Vanilla. The results support protection under the deliberate tilt, with a small observed unperturbed mean cost.

\FloatBarrier
\subsection{Optimizer-State Co-Transformation with Lion}
\label{app:exp64_moved}

A hard rebalance changes gradient coordinates while Lion retains a momentum buffer storing an EMA of past gradients. To measure the effect of this state mismatch, we compare Vanilla, hard SAGGS, Soft SAGGS, and the state-covariant variant of Appendix~\ref{par:state_covariant_saggs} under matched medium-scale Lion protocols.

\paragraph{Design.}
The state-covariant experiment comprises six runs: $\lambda\in\{0.0,0.2\}$, three paired seeds, 101M parameters and $L=24$, WikiText-103, Lion $\eta=2\times10^{-4}$, $\beta=(0.9,0.99)$, weight decay $0.1$, batch~32, 15K steps, all-pairs reparameterization, and $T_{\mathrm{bal}}=100$.  The only changed variable relative to hard SAGGS is that the Lion momentum buffer (\texttt{exp\_avg}) is multiplied by the same chain-rule factor as the gradient coordinate after every rebalance. The Vanilla, hard SAGGS, and Soft SAGGS baselines use the matched three-seed Lion baselines from the hard-rebalancing comparison. Their means and sample SDs are reported alongside the state-covariant variant in Table~\ref{tab:exp64_state_cov_cells_moved}.

\begin{table}[t!]
\centering
\caption{Lion state-covariant ablation on the 101M model with $n=3$. The upper block reports PPL as mean $\pm$ sample SD, and the lower block reports paired Vanilla-minus-State differences. Soft denotes Soft SAGGS; State denotes hard SAGGS with covariantly transformed momentum buffers.}
\label{tab:exp64_state_cov_cells_moved}
\label{tab:exp64_state_cov_paired_moved}
\small
\begin{tabular}{@{}lcccc@{}}
\toprule
$\lambda$ & Vanilla & SAGGS & Soft & State \\
\midrule
$0.0$ & $23.92\pm0.02$ & $30.12\pm10.56$ & $23.97\pm0.11$ & $24.14\pm0.15$ \\
$0.2$ & $76.89\pm31.60$ & $23.99\pm0.06$ & $23.86\pm0.12$ & $23.97\pm0.08$ \\
\bottomrule
\end{tabular}

\vspace{0.65em}

\begin{tabular}{@{}lccccc@{}}
\toprule
$\lambda$ & $\bar\Delta$ & $\sigma_\Delta$ & $t$ & $d$ & $p_{\mathrm{sf}}^{+}$ \\
\midrule
$0.0$ & $-0.225$ & $0.140$ & $-2.78$ & $-1.60$ & $1.000$ \\
$0.2$ & $+52.917$ & $31.539$ & $+2.91$ & $+1.68$ & $0.125$ \\
\bottomrule
\end{tabular}
\end{table}

\paragraph{Interpretation.}
At $\lambda=0$, state-covariant SAGGS has a lower mean and less seed variation than hard SAGGS: $24.14\pm0.15$ PPL compared with $30.12\pm10.56$ PPL in Table~\ref{tab:exp64_state_cov_cells_moved}. State-covariant SAGGS still has an observed mean cost of $0.225$ PPL relative to Vanilla, with $\bar\Delta_{\rm Vanilla-State}=-0.225$ PPL and $n=3$. These results support an optimizer-state contribution to the cost of hard rebalancing, while showing that co-transforming the momentum buffer does not eliminate the measured unperturbed mean cost.

At $\lambda=0.2$, the state-covariant variant retains the full protective effect: Vanilla minus state-covariant SAGGS is $+52.917\pm31.539$ PPL, with $d=+1.68$ (Table~\ref{tab:exp64_state_cov_paired_moved}), matching the direction and order of magnitude of the matched Vanilla-minus-hard and Vanilla-minus-Soft contrasts in Table~\ref{tab:exp64_state_cov_cells_moved}. State-covariant SAGGS and hard SAGGS differ by only $-0.021$ PPL on average at $\lambda=0.2$, while Soft SAGGS is slightly lower than state-covariant SAGGS by $0.117$ PPL on average.  These results motivate Soft SAGGS as the default when optimizer buffers are left unchanged. State-covariant hard rebalancing provides a complementary diagnostic when momentum-buffer transformations are available.

\FloatBarrier
\subsection{Branched Checkpoint-Surgery Resume}
\label{app:exp68b_moved}

\paragraph{Motivation.}
The 123.8M scale replication in Appendix~\ref{sec:exp67_modern_medium} measures sensitivity to orbit imbalance at initialization. We now test a checkpoint-resume setting in which weights are edited and training continues with a fresh optimizer. For each seed, one shared phase-1 checkpoint is cloned into four continuation branches, isolating the effect of the edit and subsequent rebalancing.

\paragraph{Design.}
For each of the seven paired seeds, we train the same RMSNorm, RoPE, SwiGLU, and GQA T5 SentencePiece model used in the 123.8M scale replication for 10K normal WikiText-103 AdamW steps.  We then reset the optimizer and branch into four 5K-step resumes: control, surgery without SAGGS, surgery with hard SAGGS, and surgery with Soft SAGGS. The non-control branches receive an exact all-pairs $\lambda=0.2$ orbit surgery at the handoff; the SAGGS branches then rebalance Q--K, V--O, and SwiGLU value--output pairs every 50 steps. Hard SAGGS applies an initial exact rebalance at the handoff; Soft SAGGS applies $\alpha=0.5$ at the handoff and during the resume. The optimizer state is deliberately not carried across the handoff. Each phase uses a fresh cosine schedule with 10\% warmup: 1K steps in phase 1 and 500 steps in phase 2. The base learning rate, weight decay, batch size, context length, and precision match the 123.8M scale replication.

\begin{table}[t!]
\centering
\caption{Branched checkpoint-surgery full paired contrasts on the 123.8M modern architecture with $n=7$.}
\label{tab:exp68b_checkpoint_surgery_paired_moved}
\small
\begin{tabular}{@{}lccccc@{}}
\toprule
\textbf{Contrast} & $\bar\Delta$ & $\sigma_\Delta$ & $t$ & $d$ & $p_{\mathrm{sf}}^{+}$ \\
\midrule
Surgery no SAGGS $-$ control & $+3.775$ & $0.416$ & $+24.00$ & $+9.07$ & $0.0078$ \\
Surgery no SAGGS $-$ SAGGS & $+3.765$ & $0.413$ & $+24.14$ & $+9.12$ & $0.0078$ \\
Surgery no SAGGS $-$ Soft & $+3.766$ & $0.414$ & $+24.06$ & $+9.09$ & $0.0078$ \\
\midrule
Control $-$ SAGGS & $-0.010$ & $0.008$ & $-3.53$ & $-1.33$ & $0.9922$ \\
Control $-$ Soft & $-0.009$ & $0.007$ & $-3.36$ & $-1.27$ & $0.9844$ \\
\bottomrule
\end{tabular}
\end{table}

The handoff uses an exact symmetry edit: Across all edited branches, the absolute validation-loss change after surgery and any initial rebalance is below $7.4\times10^{-5}$. These changes are floating-point residuals of the full handoff operation.

The optimizer-state-reset resume exposes a large coordinate effect.  Surgery with no SAGGS reaches $24.62\pm0.58$ PPL, while the control resume reaches $20.84\pm0.25$ PPL, a paired degradation of $+3.775\pm0.416$ PPL with all seven seeds in the same direction, with $p_{\rm sf}^{+}=0.0078$. Hard SAGGS and Soft SAGGS recover the final validation PPL closely: Their final PPLs are $20.85\pm0.25$ and $20.85\pm0.25$, only $0.010$ and $0.009$ PPL above the control resume, respectively. The final Q--K log-ratio makes the mechanism visible: No-SAGGS surgery ends near $6.14$, whereas both SAGGS variants end below $1.7\times10^{-4}$.

The experiment demonstrates protection in a controlled checkpoint-resume workflow with a deliberately introduced symmetry-orbit displacement and an optimizer-state reset. The comparison measures recovery of the final PPL; identical resumed trajectories require canonicalizing both checkpoints (Corollary~\ref{cor:canonical_restart}).

\FloatBarrier
\subsection{Optimizer-State Diagnostics}
\label{app:optimizer_state_diagnostics_moved}

Instead of being reset, the AdamW moments can be kept and transformed along with the weights, or loaded unchanged, which leaves the moments stale. Seven-seed experiments on the 31M model compare these cases, including moments that are stale only in the deeper half of the layers. With reset moments, the edit adds $0.277\pm0.015$ PPL relative to the matched control (Table~\ref{tab:exp70_state_preserved_cells}). Keeping and transforming the moments leaves a smaller but nonzero gap of $0.058\pm0.017$ PPL relative to a control that also keeps its optimizer state, consistent with Theorem~\ref{thm:adam_equivariance}: Under a shared learning rate, transformed moments alone do not keep the two runs equivalent. Fully stale and half-stale moments yield penalties of $0.056\pm0.015$ and $0.056\pm0.007$ PPL, respectively (Tables~\ref{tab:exp71_stale_state_cells} and~\ref{tab:exp72_partial_state_cells}). In the reset and stale cases, Soft SAGGS returns PPL to approximately the PPL of the control, with a residual of at most $0.007\pm0.006$ PPL.

\FloatBarrier
\subsection{State-Preserved Checkpoint-Surgery Diagnostic}
\label{app:exp70_moved}

The branched checkpoint-surgery resume deliberately resets optimizer state at the handoff, matching the common workflow of importing a checkpoint without its AdamW moments.  We now measure the residual penalty when optimizer moments are preserved and transformed covariantly with the exact orbit edit.

\paragraph{Design.}
For each of the seven paired seeds, we train the 31M modern RMSNorm, RoPE, SwiGLU, and GQA T5 SentencePiece model for 5K WikiText-103 AdamW steps.  We then clone the same checkpoint into five 2.5K-step branches: control reset, surgery reset, surgery reset + Soft SAGGS, control state, and surgery state-covariant. The model uses context length 256, batch size 48, and peak learning rate $4\times10^{-4}$. Phase 1 follows a 7.5K-step cosine schedule with 750-step warmup. The reset branches create a fresh AdamW optimizer and a 2.5K-step cosine schedule with 250-step warmup; the state branches load the phase-1 AdamW and scheduler state. In the state-covariant surgery branch, the exact $\lambda=0.2$ all-pairs orbit move also maps AdamW first moments by the chain-rule factor and second moments by its square.  This branch applies the covariant moment map without periodic rebalancing during phase 2.

\begin{table}[t!]
\centering
\caption{State-preserved checkpoint-surgery diagnostic on the 31M modern-architecture WikiText-103 model with $n=7$. The lower block reports paired contrasts.}
\label{tab:exp70_state_preserved_cells}
\label{tab:exp70_state_preserved_paired}
\small
\begin{tabular}{@{}lc@{}}
\toprule
\textbf{Resume condition} & \textbf{Final PPL} \\
\midrule
Control, optimizer reset & $23.27\pm0.05$ \\
Surgery, optimizer reset & $23.54\pm0.06$ \\
Surgery reset + Soft SAGGS & $23.27\pm0.06$ \\
\midrule
Control, optimizer preserved & $23.92\pm0.06$ \\
Surgery, state-covariant optimizer & $23.97\pm0.07$ \\
\bottomrule
\end{tabular}

\vspace{0.65em}

\begin{tabular}{@{}lccccc@{}}
\toprule
\textbf{Contrast} & $\bar\Delta$ & $\sigma_\Delta$ & $t$ & $d$ & $p_{\mathrm{sf}}^{+}$ \\
\midrule
Surgery reset $-$ control reset & $+0.277$ & $0.015$ & $+50.21$ & $+18.98$ & $0.0078$ \\
Surgery reset $-$ Soft & $+0.270$ & $0.013$ & $+56.23$ & $+21.25$ & $0.0078$ \\
Soft $-$ control reset & $+0.007$ & $0.006$ & $+2.98$ & $+1.13$ & $0.0156$ \\
\midrule
State-covariant $-$ state control & $+0.058$ & $0.017$ & $+8.83$ & $+3.34$ & $0.0078$ \\
Reset control $-$ state control & $-0.650$ & $0.019$ & $-88.91$ & $-33.60$ & $1.0000$ \\
\bottomrule
\end{tabular}
\end{table}

Table~\ref{tab:exp70_state_preserved_cells} reports the seven-seed summaries. The reset branch reproduces the checkpoint-surgery mechanism at a smaller model size.  Surgery with optimizer reset is $+0.277\pm0.015$ PPL worse than the reset control over seven paired seeds, with exact sign-flip $p_{\rm sf}^{+}=0.0078$. Soft SAGGS recovers the penalty almost entirely, leaving only $+0.007\pm0.006$ PPL relative to the reset control.  The largest immediate absolute validation-loss change after the edit, including the initial Soft rebalance where used, is below $6.7\times10^{-5}$.

When optimizer state is preserved and co-transformed, the surgery gap falls to $+0.058\pm0.017$ PPL relative to the state-preserved control. Moment co-transformation alone does not enforce trajectory equivariance: The shared learning rates already violate the coordinate rule in Theorem~\ref{thm:adam_equivariance}, even if $\epsilon=0$. Global clipping, finite precision, and state history can introduce further discrepancies. The residual is about one fifth of the reset-surgery gap. This comparison changes both moment history and the learning-rate continuation, so the roughly 79\% reduction cannot be attributed to moment preservation alone or used to partition the 123.8M checkpoint-surgery penalty at a different model scale. The small measured handoff loss change separately supports forward preservation. The state-preserved control itself is worse than the reset control in this short-horizon diagnostic. The experiment characterizes the residual penalty under the state-preserved continuation.

\FloatBarrier
\subsection{Stale Optimizer-State Checkpoint Edit}
\label{sec:exp71_stale_state}

The state-preserved diagnostic compares optimizer reset against a carefully co-transformed AdamW state.  A second checkpoint-resume condition arises when a checkpoint is edited by an exact symmetry transform, but its AdamW moments are loaded unchanged.  This mismatch can occur whenever model weights are imported, rescaled, or surgically modified while optimizer internals are treated as opaque checkpoint metadata.

\paragraph{Design.}
We reuse the 31M modern RMSNorm, RoPE, SwiGLU, and GQA T5 SentencePiece WikiText-103 setting of the state-preserved diagnostic. For each of the seven paired seeds, we train the model for 5K phase-1 steps, then clone the resulting checkpoint's model, AdamW, and scheduler states into four 2.5K-step branches: control state, surgery state stale, surgery state-covariant, and surgery state stale + Soft SAGGS. All surgery branches receive the exact $\lambda=0.2$ all-pairs orbit edit.  The stale branch leaves AdamW first and second moments in the old coordinates; the covariant branch maps first moments by the chain-rule factor and second moments by its square; the Soft branch uses the same stale optimizer state but applies Soft SAGGS at the handoff and every 50 steps.

\begin{table}[t!]
\centering
\caption{Stale AdamW-state diagnostic on the 31M model with $n=7$: final PPL as mean $\pm$ sample SD and paired contrasts after an exact $\lambda=0.2$ edit. Stale branches retain the pre-edit moments.}
\label{tab:exp71_stale_state_cells}
\label{tab:exp71_stale_state_paired}
\small
\begin{tabular}{@{}lc@{}}
\toprule
\textbf{Resume condition} & \textbf{Final PPL} \\
\midrule
Control, preserved state & $23.91\pm0.05$ \\
Surgery + stale AdamW state & $23.96\pm0.07$ \\
Surgery + covariant AdamW state & $23.96\pm0.07$ \\
Surgery + stale state + Soft SAGGS & $23.90\pm0.05$ \\
\bottomrule
\end{tabular}

\vspace{0.65em}

\begin{tabular}{@{}lccccc@{}}
\toprule
\textbf{Contrast} & $\bar\Delta$ & $\sigma_\Delta$ & $t$ & $d$ & $p_{\mathrm{sf}}^{+}$ \\
\midrule
Stale $-$ control & $+0.056$ & $0.015$ & $+9.86$ & $+3.73$ & $0.0078$ \\
Stale $-$ covariant & $+0.002$ & $0.001$ & $+6.42$ & $+2.42$ & $0.0078$ \\
Stale $-$ Soft & $+0.062$ & $0.016$ & $+10.52$ & $+3.98$ & $0.0078$ \\
\midrule
Covariant $-$ control & $+0.054$ & $0.015$ & $+9.21$ & $+3.48$ & $0.0078$ \\
Soft $-$ control & $-0.005$ & $0.003$ & $-5.45$ & $-2.06$ & $1.0000$ \\
\bottomrule
\end{tabular}
\end{table}

Table~\ref{tab:exp71_stale_state_cells} reports the seven-seed summaries. The stale-state edit creates a small but highly consistent penalty: $+0.056\pm0.015$ PPL relative to the state-preserved control over seven paired seeds, with $p_{\rm sf}^{+}=0.0078$, despite an immediate handoff validation-loss change below $7.3\times10^{-5}$.  Co-transforming AdamW moments is slightly better than stale reuse by only $+0.002\pm0.001$ PPL, and still remains $+0.054\pm0.015$ PPL worse than control.  This residual is consistent with Theorem~\ref{thm:adam_equivariance}: Keeping common learning rates while co-transforming moments is already insufficient for trajectory equivariance.  Clipping, finite precision, and the $\epsilon$ floor can add to this effect; the contrast does not isolate their shares.

Soft SAGGS removes the stale-state penalty.  The stale-minus-Soft contrast is $+0.062\pm0.016$ PPL over all seven seeds, with $p_{\rm sf}^{+}=0.0078$, while Soft itself is $-0.005\pm0.003$ PPL relative to the state-preserved control.  The effect size is much smaller than in the optimizer-reset workflow of the branched checkpoint-surgery resume, but the exact checkpoint edit still changes subsequent training when moments are stale. Periodic Soft SAGGS recovers the control PPL in these runs.

\FloatBarrier
\subsection{Partial Optimizer-State Checkpoint Edit}
\label{sec:exp72_partial_state}

The stale-state diagnostic treats the AdamW state as either fully stale or fully co-transformed. Real checkpoint workflows can fail in a more granular way: Optimizer-state shards may be missing or stale only for a subset of layers.  This experiment tests that intermediate failure mode while keeping the forward edit exact.

\paragraph{Design.}
We again use the 31M modern RMSNorm, RoPE, SwiGLU, and GQA T5 SentencePiece WikiText-103 setting of the state-preserved and stale-state diagnostics. For each of the seven paired seeds, one 5K-step phase-1 checkpoint is cloned into five 2.5K-step branches: control state, surgery covariant, surgery all-stale, surgery deep-stale, and surgery deep-stale + Soft SAGGS. All surgery branches receive the exact $\lambda=0.2$ all-pairs orbit edit. The covariant branch maps every AdamW first and second moment into the new coordinates.  The all-stale branch leaves every moment in the old coordinates.  The deep-stale branch maps only the shallow half of layers and leaves layers 6 through 11 stale, mimicking a partially missing optimizer-state shard.  The Soft branch starts from this same partial-state condition and uses Soft SAGGS at handoff and every 50 steps.

\begin{table}[t!]
\centering
\caption{Partial AdamW-state diagnostic on the 31M model with $n=7$: final PPL as mean $\pm$ sample SD and paired contrasts. In deep-half-stale branches, only shallow-layer moments are co-transformed.}
\label{tab:exp72_partial_state_cells}
\label{tab:exp72_partial_state_paired}
\small
\begin{tabular}{@{}lc@{}}
\toprule
\textbf{Resume condition} & \textbf{Final PPL} \\
\midrule
Control state preserved & $23.90\pm0.07$ \\
Exact edit, covariant state & $23.96\pm0.07$ \\
Exact edit, all state stale & $23.96\pm0.07$ \\
Exact edit, deep-half stale & $23.96\pm0.07$ \\
Deep-half stale + Soft SAGGS & $23.90\pm0.06$ \\
\bottomrule
\end{tabular}

\vspace{0.65em}

\begin{tabular}{@{}lccccc@{}}
\toprule
\textbf{Contrast} & $\bar\Delta$ & $\sigma_\Delta$ & $t$ & $d$ & $p_{\mathrm{sf}}^{+}$ \\
\midrule
All stale $-$ control & $+0.056$ & $0.008$ & $+19.19$ & $+7.25$ & $0.0078$ \\
Deep stale $-$ control & $+0.056$ & $0.007$ & $+19.86$ & $+7.51$ & $0.0078$ \\
Deep stale $-$ covariant & $+0.003$ & $0.002$ & $+4.45$ & $+1.68$ & $0.0078$ \\
All stale $-$ deep stale & $+0.000$ & $0.001$ & $+0.71$ & $+0.27$ & $0.2500$ \\
\midrule
Deep stale $-$ Soft & $+0.062$ & $0.008$ & $+19.92$ & $+7.53$ & $0.0078$ \\
Soft $-$ control & $-0.005$ & $0.002$ & $-8.05$ & $-3.04$ & $1.0000$ \\
\bottomrule
\end{tabular}
\end{table}

Table~\ref{tab:exp72_partial_state_cells} reports the seven-seed summaries. The partial-state condition changes subsequent training despite preserving the current function: The immediate handoff validation-loss change, including the initial Soft rebalance where used, is below $6.2\times10^{-5}$, but the deep-stale branch is $+0.056\pm0.007$ PPL worse than the state-preserved control over seven paired seeds, with $p_{\rm sf}^{+}=0.0078$.  This partial-state penalty is essentially the same penalty as leaving all moments stale, $+0.056\pm0.008$ PPL. Leaving the additional shallow-half moments stale changes mean PPL by only $0.00027$; this difference measures the incremental cost of stale moments, not the entire surgery penalty. Relative to the fully covariant branch, the deep-stale condition is still $+0.003\pm0.002$ PPL worse, so the partial shard mismatch is measurable but much smaller than the orbit edit itself under the practical protocol.

Soft SAGGS again removes the penalty.  The deep-stale-minus-Soft contrast is $+0.062\pm0.008$ PPL, with all seven paired seeds in the same direction, and Soft is $-0.005\pm0.002$ PPL relative to the control. Together with the reset and stale-state comparisons, these results show that periodic rebalancing recovers control-level PPL across the tested optimizer-state conditions.

\FloatBarrier
\section{Additional Experiments}
\label{app:additional_experiments}

\label{app:precursors_moved}
\label{app:supporting_moved}
These experiments evaluate numerical identities, naturally occurring imbalance, and transfer across architectures, corpora, and model sizes. Seed counts and method variants are stated for each protocol.

\subsection{Gradient Law and Orbit Imbalance}
\label{sec:mechanism_checks}
\label{sec:exp_verification}
\label{sec:exp_natural_drift}

\subsubsection{Gradient-Law Verification}

We evaluate the gradient-norm scaling implied by \eqref{eq:grad_transform} on the same batch at orbit-related points, using FP32, or BF16 at 1.2B parameters. The exact pairs agree with the predicted scaling within floating-point error, whereas GELU MLP pairs show substantially larger deviations. Appendix~\ref{app:exp1_detail_moved} defines the residual and slope measurements. Table~\ref{tab:exact_verification} reports their maxima.

\subsubsection{Natural Drift}
We test whether pair-scale imbalance also develops without a deliberate edit. We train three ReLU models at $\lambda=0$ for 30K steps with two seeds: a small model with 22M parameters and $L=12$, a medium model with 101M parameters and $L=24$, and a deep model with 51M parameters and $L=48$. We track the maximum absolute log-ratio across layers and measure its growth from initialization to the last logged step. Every reported seed shows positive growth. Q--K pairs have the largest mean increases, ranging from $0.14$ to $0.20$ (Table~\ref{tab:natural_drift}). Appendix~\ref{app:exp2_detail_moved} defines the measurement and its numerical stabilizer. These results establish naturally occurring imbalance; its effect on PPL is assessed by the paired training comparisons in Sec.~\ref{sec:lm_comparisons}.

\subsubsection{Additional Stress and Scale Checks}
Supplementary experiments examine deeper models, SwiGLU, 354M models, and a single-seed 1.2B training run. These experiments generally show larger PPL effects under imposed rescaling than during unperturbed training, with exceptions documented in the experiments below. These comparisons complement the primary seven-seed results.

\subsection{Supporting Ablations and Replications}
\label{app:supporting_comparisons_summary}
Three-seed replications give Vanilla-minus-Soft PPL differences of $+18.278$ on OpenWebText, $+21.156$ at 123.8M parameters, and $+85.813\pm4.730$ at 296.7M under $\lambda=0.2$. Appendices~\ref{sec:exp62_owt_paired}, \ref{sec:exp67_modern_medium}, and~\ref{sec:exp69_modern_large} give the results. Their smaller seed counts support comparisons of direction and magnitude. Synthetic contraction and optimizer-state ablations provide separate diagnostics of the rebalancing operation, as reported in Appendices~\ref{sec:exp59_simulation} and~\ref{app:exp64_moved}.

\FloatBarrier
\subsection{FP32 Gradient-Law Verification}
\label{app:exp1_detail_moved}

We verify the gradient transformation law of \eqref{eq:grad_transform} by computing gradients at the original parameterization and the reparameterized configuration for the same batch. Table~\ref{tab:exact_verification} uses FP32; the 1.2B check uses BF16 mixed precision. The verification compares gradient norms. Writing $g_A\coloneqq\frobnorm{\grad_A\loss}$ and $g'_A\coloneqq\frobnorm{\grad_{A'}\loss}$, its relative residual is
\begin{equation}
  \varepsilon_A\coloneqq\frac{|g'_A-c^{-1}g_A|} {\max\{c^{-1}g_A,\epsilon_{\rm num}\}},\qquad \varepsilon_B\coloneqq\frac{|g'_B-cg_B|} {\max\{cg_B,\epsilon_{\rm num}\}},\qquad \epsilon_{\rm num}\coloneqq10^{-12}.
  \label{eq:exp1_residual_moved}
\end{equation}
The reported $\varepsilon_{\rm rel}$ is the maximum over tested blocks and layers. This residual verifies the norm consequence of the law; small norm residuals alone do not verify gradient directions.

\begin{table}[t!]
\centering
\caption{Maximum relative gradient-norm residual $\varepsilon_{\mathrm{rel}}$ (\eqref{eq:exp1_residual_moved}) in FP32, by depth, tilt, and pair type. GELU MLPs are included as a non-exact comparison.}
\label{tab:exact_verification}
\vspace{4pt}
\small
\begin{tabular}{@{}lcccc@{}}
\toprule
\textbf{Pair} & $L$ & $\lambda = 0.01$ & $\lambda = 0.05$ & $\lambda = 0.1$ \\
\midrule
\multirow{3}{*}{Q--K (ReLU)} & 12 & $4.7\times10^{-7}$ & $2.7\times10^{-7}$ & $2.4\times10^{-7}$ \\
 & 24 & $7.4\times10^{-5}$ & $4.9\times10^{-5}$ & $1.2\times10^{-4}$ \\
 & 48 & $1.8\times10^{-4}$ & $2.2\times10^{-4}$ & $1.3\times10^{-4}$ \\
\midrule
\multirow{3}{*}{V--O (ReLU)} & 12 & $1.2\times10^{-7}$ & $9.8\times10^{-8}$ & $1.0\times10^{-7}$ \\
 & 24 & $8.6\times10^{-5}$ & $7.6\times10^{-5}$ & $8.6\times10^{-5}$ \\
 & 48 & $1.1\times10^{-5}$ & $4.7\times10^{-5}$ & $5.2\times10^{-5}$ \\
\midrule
\multirow{3}{*}{MLP (ReLU)} & 12 & $8.7\times10^{-8}$ & $1.6\times10^{-7}$ & $6.0\times10^{-5}$ \\
 & 24 & $1.9\times10^{-5}$ & $6.1\times10^{-5}$ & $6.2\times10^{-5}$ \\
 & 48 & $5.7\times10^{-5}$ & $5.5\times10^{-5}$ & $7.4\times10^{-5}$ \\
\midrule
\multirow{3}{*}{MLP (GELU)} & 12 & $1.2\times10^{-2}$ & $7.0\times10^{-2}$ & $1.4\times10^{-1}$ \\
 & 24 & $3.1\times10^{-2}$ & $1.6\times10^{-1}$ & $2.0\times10^{-1}$ \\
 & 48 & $6.3\times10^{-2}$ & $1.5\times10^{-1}$ & $1.1\times10^{-1}$ \\
\midrule
\multirow{3}{*}{MLP (SwiGLU)} & 12 & $1.6\times10^{-7}$ & $1.5\times10^{-7}$ & $1.1\times10^{-7}$ \\
 & 24 & $1.3\times10^{-7}$ & $1.9\times10^{-7}$ & $1.3\times10^{-7}$ \\
 & 48 & $1.5\times10^{-7}$ & $1.1\times10^{-7}$ & $1.6\times10^{-7}$ \\
\bottomrule
\end{tabular}
\end{table}

The reported gradient-ratio depth slopes fit $h_{\ell,\delta_g}\coloneqq\log(g_{B,\ell}+\delta_g)- \log(g_{A,\ell}+\delta_g)$ against layer index, with $\delta_g\coloneqq10^{-12}$. The exact $2\log c_\ell$ shift under an orbit move applies to the unstabilized log-ratio; a fixed additive $\delta_g$ can change this shift when a gradient norm is near the floor. We test across depths $L \in \{12, 24, 48\}$, tilt parameters $\lambda \in \{0.01, 0.05, 0.1\}$, and all symmetry pairs Q--K, V--O, and MLP.

For the exact symmetries, \ie{} Q--K and V--O across all activations, ReLU MLP, and SwiGLU value--output, the reported relative errors range from approximately $10^{-7}$ to $2.2\times10^{-4}$ across the reported FP32 configurations. In particular, the $L=12$ ReLU MLP row reaches $6.0\times10^{-5}$ at $\lambda=0.1$; the largest Q--K residual is $2.2\times10^{-4}$ at $L=48$. These residuals are consistent with floating-point error in an exact symmetry. GELU MLP shows substantially larger errors with a non-monotone dependence on $\lambda$ in the deepest row, up to $2\times10^{-1}$, confirming that the deviation is dominated by GELU's lack of exact homogeneity rather than numerical precision. In the 1.2B gradient-law verification (Table~\ref{tab:1b_verify}), the law continues to hold with relative error below $0.16\%$ under BF16 mixed precision. The SwiGLU value--output rows have norm residuals near $10^{-7}$.

\FloatBarrier
\subsection{Naturally Occurring Pair Imbalance}
\label{app:exp2_detail_moved}

At each logging step, before the optimizer update, we track the maximum stabilized log-ratio imbalance across all layers:
\begin{equation}
 r_{\ell,\delta}^{(t)}\coloneqq \log\frac{\frobnorm{\mat{A}_\ell^{(t)}}+\delta} {\frobnorm{\mat{B}_\ell^{(t)}}+\delta},\qquad \Delta_{\max}^{(t)}\coloneqq\max_\ell|r_{\ell,\delta}^{(t)}|, \qquad \delta\coloneqq10^{-12}.
 \label{eq:logged_imbalance_moved}
\end{equation}
We record this statistic for each symmetry pair type Q--K, V--O, and MLP. This statistic is the absolute imbalance at step $t$. Table~\ref{tab:natural_drift} instead reports its increase between the first and last logged states:
\begin{equation}
 G\coloneqq\Delta_{\max}^{(29.5\mathrm{K})}-\Delta_{\max}^{(0)}.
 \label{eq:natural_drift_growth_moved}
\end{equation}
The reported $G$ is a difference of maxima. The growth statistic $G$ is distinct from both the endpoint $\Delta_{\max}^{(29.5\mathrm{K})}$ and the maximum layerwise change $\max_\ell|r_{\ell,\delta}^{(29.5\mathrm{K})}-r_{\ell,\delta}^{(0)}|$.

For positive norms $a\coloneqq\frobnorm{\mat{A}}$ and $b\coloneqq\frobnorm{\mat{B}}$, the ideal log-ratio $r\coloneqq\log(a/b)$ used in the theory satisfies $r_\delta-r=\log(1+\delta/a)-\log(1+\delta/b)$. Thus the stabilized measurement approximates $r$ when both norms are large relative to $\delta$, but the stabilized measurement does not obey the exact orbit-shift or contraction identities near the floor. Experimental references to measured log-ratios use this numerical convention. Since initialization is not exactly balanced, a nonzero endpoint alone does not establish growth from initialization; positive $G$ does establish an increase in the maximum absolute imbalance. We train for 30,000 steps with two seeds, context length 256, 2K-step warmup, and logging every 500 steps. Peak learning rate is $6\times10^{-4}$ for the small and deep models and $3\times10^{-4}$ for the medium model. The last logged drift state is at step 29.5K, before the final 500 updates.

The log-ratio increment depends on parameter and update inner products, as in \eqref{eq:log_ratio_first_order}; different gradient magnitudes alone do not determine the direction or persistence of drift.

\begin{table}[t!]
\centering
\caption{Increase in maximum absolute symmetry-pair log-ratio during standard ReLU training at $\lambda=0$. The reported $G=\Delta_{\max}^{(29.5\mathrm{K})}-\Delta_{\max}^{(0)}$ uses $\delta=10^{-12}$ as in Appendix~\ref{app:exp2_detail_moved}. Values are mean $\pm$ population SD over two seeds.}
\label{tab:natural_drift}
\vspace{4pt}
\begin{tabular}{@{}lrcccc@{}}
\toprule
\textbf{Model} & \textbf{Params} & $L$ & \textbf{Q--K $G$} & \textbf{V--O $G$} & \textbf{MLP $G$} \\
\midrule
Small       & 22M  & 12 & $0.200 \pm 0.026$ & $0.068 \pm 0.004$ & $0.099 \pm 0.002$ \\
Medium      & 101M & 24 & $0.136 \pm 0.003$ & $0.047 \pm 0.002$ & $0.037 \pm 0.002$ \\
Deep        & 51M  & 48 & $0.163 \pm 0.021$ & $0.120 \pm 0.026$ & $0.093 \pm 0.003$ \\
\bottomrule
\end{tabular}
\end{table}

\begin{figure}[t!]
\centering
\includegraphics[width=0.68\linewidth]{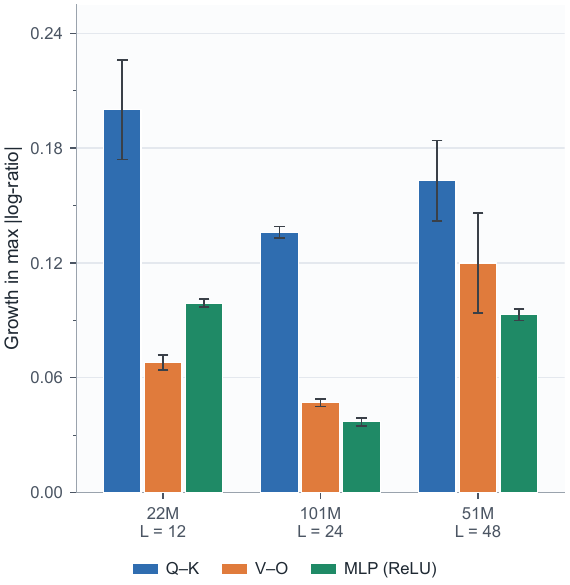}
\caption{Growth $G$ in maximum absolute pair log-ratio from step 0 to 29.5K during standard ReLU training. Q--K pairs show the largest mean growth. The ordering of MLP and V--O differs across model configurations. Error bars show $\pm1$ population SD over two seeds.}
\label{fig:natural_drift_moved}
\end{figure}

\FloatBarrier
\subsection{Pair Imbalance in a Pre-Trained Model}
\label{sec:exp_pretrained_drift}

\paragraph{Motivation.}
The natural-drift experiment measures changes in pair imbalance during training from scratch. Here we inspect a released GPT-2 checkpoint to determine whether its learned attention pairs have unequal norms.

\paragraph{Checkpoint measurement.}
We load GPT-2 small~\citep{radford2019language}, with 124M parameters, $d = 768$, $L = 12$, and GELU, from Hugging Face and measure the symmetry pair-norm balance. The recorded maximum absolute log-ratios are $0.221$ for Q--K and $0.931$ for V--O, so V--O has the larger imbalance in this checkpoint. These pre-fine-tuning measurements use stabilizer $\delta\coloneqq10^{-8}$. Without the corresponding initial norms, these endpoint measurements do not determine how much imbalance accumulated during pre-training. We report the attention-pair maxima; GELU up--down rescaling is not an exact symmetry. Exact attention-pair edits of this biased model also require the bias co-transforms stated in Appendix~\ref{app:notation_setup_moved}. These observations motivate studying attention-pair canonicalization during fine-tuning.

\FloatBarrier
\subsection{Deep-Model Sensitivity to Rescaling}
\label{sec:exp_lambda_breakdown}

\paragraph{Motivation.}
We test whether rebalancing protects deep Transformer training from deliberate exact Q--K orbit tilts. The deepest layer's scale factor is $c_{47}\coloneqq e^{47\lambda}$; the gradient-coordinate identity alone does not predict the resulting PPL or a critical value of $\lambda$.

\paragraph{Setup.}
We train the deep ReLU model with $d=256$ and $L=48$ on WikiText-103 for 30,000 AdamW steps with two seeds, applying Q--K rescaling at $\lambda\in\{0,0.2,0.3\}$. The recipe uses peak learning rate $3\times10^{-4}$, batch size 64, context length 256, and 2K-step warmup. We compare Vanilla against SAGGS+P, with rebalancing every 100 steps and auxiliary-penalty weight $0.01$.

\paragraph{Results.}
Table~\ref{tab:lambda_breakdown_deep} shows the results. At $\lambda = 0.2$, where $c_{47} \approx 1.2 \times 10^4$, Vanilla PPL approximately doubles from 30.34 to 61.46, while SAGGS+P maintains near-baseline PPL of 30.31. At $\lambda = 0.3$, where $c_{47} \approx 1.3 \times 10^6$, Vanilla reaches PPL 150.58, while SAGGS+P achieves 30.35, close to the unperturbed baseline in this two-seed stress test. The PPL gap of $+120.23$ at $\lambda = 0.3$ is a stress-test result: Orbit rebalancing removes the injected symmetry-coordinate distortion.

\begin{table}[t!]
\centering
\caption{Deep ReLU model with $L=48$ under Q--K rescaling over 30K steps. Values are best validation PPL, reported as mean $\pm$ population SD over two seeds; $\Delta$ denotes Vanilla minus SAGGS+P.}
\label{tab:lambda_breakdown_deep}
\vspace{4pt}
\begin{tabular}{@{}lcrrr@{}}
\toprule
$\lambda$ & $c_{47}$ & \multicolumn{1}{c}{\textbf{Vanilla}} & \multicolumn{1}{c}{\textbf{SAGGS+P}} & $\Delta$ \\
\midrule
$0.0$ & $1$                         & $30.34 \pm 0.02$  & $30.27 \pm 0.01$ & $+0.07$ \\
$0.2$ & $\approx 1.2 \times 10^{4}$ & $61.46 \pm 2.99$  & $30.31 \pm 0.04$ & $+31.15$ \\
$0.3$ & $\approx 1.3 \times 10^{6}$ & $150.58 \pm 0.96$ & $30.35 \pm 0.06$ & $+120.23$ \\
\bottomrule
\end{tabular}
\end{table}

\FloatBarrier
\subsection{Rescaling a 354M-Parameter Model}
\label{sec:exp_large_reparam}

\paragraph{Motivation.}
We test whether the protection observed in the deep-model stress experiment also occurs at 354M parameters. The experiment compares unperturbed training with exact Q--K reparameterization at $\lambda\in\{0.1,0.2\}$ under the same training recipe.

\paragraph{Setup.}
We train the xlarge GELU model with $d=1024$ and $L=24$ for 15,000 AdamW steps on a single NVIDIA H100 GPU, applying Q--K rescaling at $\lambda\in\{0,0.1,0.2\}$. The sequence length is 512, and batch size 8 with 4$\times$ gradient accumulation gives an effective batch size of 32. SAGGS+P applies only to Q--K and V--O pairs, where the bilinear symmetry is exact regardless of activation, with interval 200 and auxiliary penalty weight 0.005. Warmup is 1.5K steps. The reported PPL is the best validation value; gradient slopes use the last logged state at 14.5K.

\paragraph{Results.}
Table~\ref{tab:large_reparam} reports one seed at $\lambda=0$ and $0.1$, and three seeds at $\lambda=0.2$. At $\lambda=0$, the PPL values are $23.53$ and $23.59$, a $0.06$ increase for SAGGS+P in this single-seed comparison.

At $\lambda = 0.1$, where $c_{23} = e^{0.1 \times 23} \approx 10$, Vanilla achieves $23.89$ while SAGGS+P maintains $23.56$, giving a $+0.33$ PPL gap at this scale factor. At $\lambda = 0.2$, where $c_{23} = e^{0.2 \times 23} \approx 100$, Vanilla PPL degrades to $25.84 \pm 0.04$ over 3 seeds, a $+2.31$ increase over the $23.53$ Vanilla baseline, while SAGGS+P maintains $23.58 \pm 0.06$, essentially unchanged from the unperturbed case. All three seed-paired differences favor SAGGS+P, with small dispersion in these observed runs. The final logged gradient slope is $0.187$ for Vanilla at $\lambda = 0.2$ and $0.017$ for SAGGS+P.

\begin{table}[t!]
\centering
\caption{354M GELU model with Q--K rescaling over 15K steps. Values are best validation PPL, with $\Delta$ denoting Vanilla minus SAGGS+P; gradient slopes are seed means at the last logged state, step 14.5K. The row marked $^{\dagger}$ uses three seeds, with PPL as mean $\pm$ sample SD; the other rows use a single seed.}
\label{tab:large_reparam}
\vspace{4pt}
\begin{tabular}{@{}lcllrcc@{}}
\toprule
& & \multicolumn{3}{c}{\textbf{PPL}} & \multicolumn{2}{c@{}}{\textbf{Gradient slope}} \\
\cmidrule(lr){3-5}\cmidrule(l){6-7}
$\lambda$ & $c_{23}$ & \multicolumn{1}{c}{Vanilla} & \multicolumn{1}{c}{SAGGS+P} & $\Delta$ & Vanilla & SAGGS+P \\
\midrule
$0.0$           & $1$           & $23.53$          & $23.59$          & $-0.06$ & $0.018$ & $0.020$ \\
$0.1$           & $\approx 10$  & $23.89$          & $23.56$          & $+0.33$ & $0.097$ & $0.018$ \\
$0.2^{\dagger}$ & $\approx 100$ & $25.84 \pm 0.04$ & $23.58 \pm 0.06$ & $+2.26$ & $0.187$ & $0.017$ \\
\bottomrule
\end{tabular}
\end{table}

\FloatBarrier
\subsection{SwiGLU Sensitivity to Rescaling}
\label{sec:exp_swiglu_breakdown}

\paragraph{Motivation.}
The gradient-law verification in Appendix~\ref{app:exp1_detail_moved} confirmed SwiGLU's exact V--O symmetry with the nonlinear gate held fixed. SwiGLU is used in modern Transformer architectures, motivating a test of this activation. We vary the rescaling strength for SwiGLU, following the ReLU comparison in the deep-model stress test.

\paragraph{Setup.}
We compare two SwiGLU models with $d=256$ and $d_{\mathrm{ff}}=1024$: a 12-layer model with 25.5M parameters and a 48-layer model with 63.3M parameters. Rescaling acts on the MLP value--output pair, with $\lambda\in\{0,0.1,0.2,0.3\}$ for the smaller model and $\lambda\in\{0,0.2\}$ for the deeper model. Each configuration uses 30,000 WikiText-103 training steps and a single seed. The MLP-only runs use learning rate $6\times10^{-4}$ at $L=12$ and $3\times10^{-4}$ at $L=48$, context 256, batch 64, and 2K-step warmup. The table reports best validation PPL; pair imbalance and gradient slope use the last logged state at step 29.5K.

\paragraph{Results.}
Table~\ref{tab:swiglu_breakdown} reports the results.
\paragraph{Deep model, $L = 48$.}
At $\lambda = 0.2$, where $c_{47} = e^{0.2 \times 47} \approx 1.2\times10^{4}$, Vanilla PPL degrades severely to $63.24$ with an MLP log-ratio imbalance of $8.70$ and gradient slope $0.212$, while SAGGS+P maintains $28.46$ PPL with pair imbalance below $10^{-4}$ and slope $0.0002$. Its PPL is close to the $\lambda=0$ baseline of $28.48$, recovering nearly all of the degradation caused by the imposed rescaling. SAGGS+P at $\lambda = 0$ achieves $28.42$, compared with the Vanilla baseline of $28.48$, a $0.06$ PPL difference in this single-seed protocol.

\paragraph{Small model, $L = 12$.}
The lambda sweep reveals a smooth degradation curve: Vanilla PPL increases from $31.79$ at $\lambda = 0$ to $31.98$ at $\lambda = 0.1$ to $32.48$ at $\lambda = 0.2$ to $33.37$ at $\lambda = 0.3$, with final logged MLP imbalance growing from $0.061$ to $0.193$ to $0.718$ to $1.694$ across the four tilts. SAGGS+P consistently maintains baseline performance across all $\lambda$ values: $31.80$ at $\lambda = 0.1$, $31.77$ at $\lambda = 0.2$, and $31.85$ at $\lambda = 0.3$, all with pair imbalance below $10^{-4}$. At $\lambda=0.3$, the increase above the unperturbed Vanilla baseline of $31.79$ is $+1.58$ PPL for Vanilla, which reaches $33.37$, and $+0.06$ for SAGGS+P.

\begin{figure}[t!]
\centering
\includegraphics[width=\linewidth]{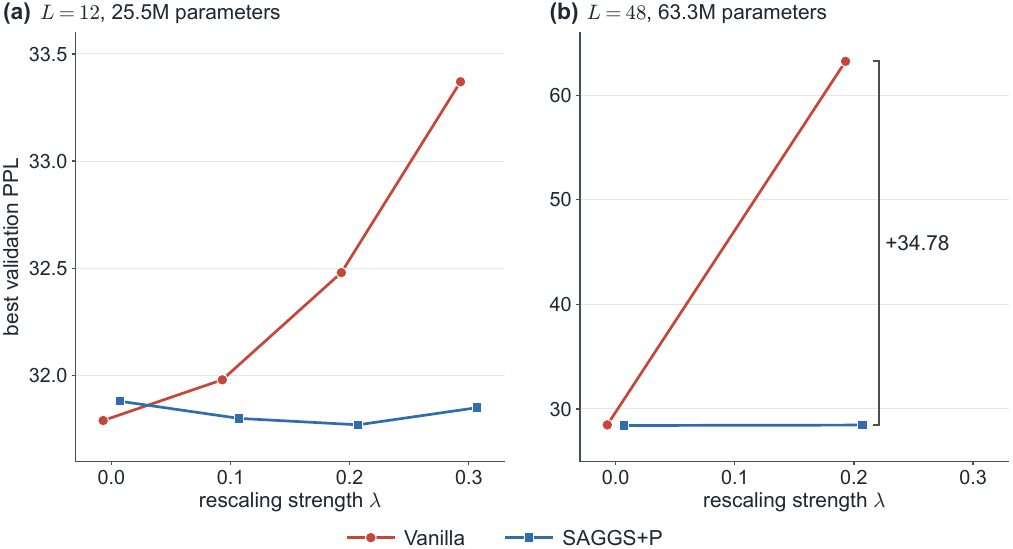}
\caption{Single-seed SwiGLU MLP rescaling at (a)~$L=12$ and (b)~$L=48$: best validation PPL for Vanilla and SAGGS+P (Table~\ref{tab:swiglu_breakdown}). The deeper model uses only $\lambda\in\{0,0.2\}$. Vanilla markers shift left and SAGGS+P markers shift right by $0.007$ for visibility; the bracket in (b) gives Vanilla minus SAGGS+P at $\lambda=0.2$.}
\label{fig:swiglu_breakdown}
\end{figure}

The $+34.78$ PPL gap between Vanilla and SAGGS+P at $\lambda = 0.2$ in the deep model shows substantial sensitivity to the exact SwiGLU value--output reparameterization in this training recipe. Figure~\ref{fig:swiglu_breakdown} plots the corresponding validation-PPL curves at both depths.

\begin{table}[t!]
\centering
\caption{Single-seed SwiGLU MLP rescaling: best validation PPL over 30K steps, with MLP pair imbalance and gradient-ratio depth slope measured at the last logged state, step 29.5K.}
\label{tab:swiglu_breakdown}
\label{tab:swiglu_small}
\vspace{4pt}
\small
\begin{tabular}{@{}clccc@{}}
\toprule
$\lambda$ & \textbf{Method} & \textbf{PPL} & \textbf{MLP imbalance} & \textbf{Gradient slope} \\
\midrule
\multicolumn{5}{@{}l}{Deep, $L=48$} \\
$0.0$ & Vanilla (baseline) & $28.48$ & $0.052$ & $0.001$ \\
      & SAGGS+P            & $28.42$ & $<10^{-4}$ & $0.001$ \\
$0.2$ & Vanilla            & $63.24$ & $8.70$ & $0.212$ \\
      & SAGGS+P            & $28.46$ & $<10^{-4}$ & $0.0002$ \\
\midrule
\multicolumn{5}{@{}l}{Small, $L=12$} \\
$0.0$ & Vanilla & $31.79$ & $0.061$ & $0.015$ \\
      & SAGGS+P & $31.88$ & $<10^{-4}$ & $0.015$ \\
$0.1$ & Vanilla & $31.98$ & $0.193$ & $0.026$ \\
      & SAGGS+P & $31.80$ & $<10^{-4}$ & $0.011$ \\
$0.2$ & Vanilla & $32.48$ & $0.718$ & $0.071$ \\
      & SAGGS+P & $31.77$ & $<10^{-4}$ & $0.012$ \\
$0.3$ & Vanilla & $33.37$ & $1.694$ & $0.165$ \\
      & SAGGS+P & $31.85$ & $<10^{-4}$ & $0.009$ \\
\bottomrule
\end{tabular}
\end{table}

\FloatBarrier
\subsection{1.2B-Parameter Gradient Law Verification}
\label{sec:exp_1b_verify}

\paragraph{Motivation.}
We measure gradient-law residuals at billion-parameter scale under BF16 mixed precision.

\paragraph{Setup.}
We construct a 1.22B-parameter SwiGLU Transformer with $d = 2048$, $L = 22$, $n_{\mathrm{heads}} = 16$, and $d_{\mathrm{ff}} = 5504$, with gradient checkpointing, and measure gradient norms at a single minibatch under BF16 mixed precision. For each of the symmetry pair types Q--K, V--O, and MLP, and $\lambda \in \{0.05, 0.1, 0.2, 0.3\}$, we reparameterize from a common baseline state and compare the measured gradient norms against the theoretical predictions $\norm{\grad_{\mat{A}'_\ell}\loss} = c_\ell^{-1}\norm{\grad_{\mat{A}_\ell}\loss}$ and $\norm{\grad_{\mat{B}'_\ell}\loss} = c_\ell\norm{\grad_{\mat{B}_\ell}\loss}$.

\paragraph{Results.}
Table~\ref{tab:1b_verify} reports the maximum relative error across all layers for each pair type and $\lambda$ value. Across all 12 configurations, the maximum relative error remains below $0.16\%$, with the MLP SwiGLU value--output pair achieving particularly precise agreement below $0.055\%$. At $\lambda = 0.3$, the deepest layer's scale factor is $c_{21}=e^{0.3\times21}\approx545$. The extreme coordinate multiplier ratio is $c_{21}^2=e^{12.6}\approx2.97\times10^5$; the actual gradient-norm spread also depends on the baseline gradients. These measurements show that the norm-scaling identity remains numerically accurate in this billion-parameter BF16 configuration.

\begin{table}[t!]
\centering
\caption{1.2B-parameter gradient law verification in BF16 mixed precision. Values are the maximum relative error across all 22 layers for each symmetry pair type and $\lambda$. All errors are below $0.16\%$.}
\label{tab:1b_verify}
\vspace{4pt}
\begin{tabular}{@{}lcccc@{}}
\toprule
\textbf{Pair type} & \multicolumn{4}{c}{\textbf{Maximum relative error} (\%)} \\
\cmidrule(l){2-5}
 & $\lambda = 0.05$ & $\lambda = 0.1$ & $\lambda = 0.2$ & $\lambda = 0.3$ \\
\midrule
Q--K & $0.152$ & $0.118$ & $0.129$ & $0.146$ \\
V--O & $0.110$ & $0.107$ & $0.110$ & $0.156$ \\
MLP (SwiGLU) & $0.030$ & $0.042$ & $0.038$ & $0.054$ \\
\bottomrule
\end{tabular}
\end{table}

\FloatBarrier
\subsection{Rescaling Strength at 101M Parameters}
\label{sec:exp_medium_lambda}

\paragraph{Motivation.}
We test whether the advantage of SAGGS+P on the medium model with $d=512$, $L=24$, and 101M parameters grows at higher $\lambda$ values where the scale factor $c_{23} = e^{23\lambda}$ becomes large.

\paragraph{Setup.}
We compare Vanilla and SAGGS+P on the medium ReLU model with Q--K rescaling at $\lambda\in\{0.1,0.2,0.3\}$. Training uses a single seed, 30,000 WikiText-103 steps, learning rate $\eta=6\times10^{-4}$, context length 512, batch size 32, and a 2K-step warmup. SAGGS+P uses rebalancing every 100 steps and auxiliary-penalty weight $0.01$.

\paragraph{Results.}
Table~\ref{tab:medium_lambda_ext} reports the results. At $\lambda = 0.1$, where $c_{23} \approx 10$, SAGGS+P and Vanilla differ by $0.01$ PPL, reaching $22.90$ and $22.91$, respectively. At $\lambda = 0.2$, where $c_{23} \approx 100$, SAGGS+P begins to show benefit: SAGGS+P reaches $22.89$ and Vanilla reaches $23.24$, so $\Delta = -0.35$. At $\lambda = 0.3$, where $c_{23} \approx 1000$, the gap widens substantially: SAGGS+P reaches $22.92$ and Vanilla reaches $25.91$, so $\Delta = -3.00$. SAGGS+P at $\lambda = 0.3$ remains within $0.01$ PPL of the $\lambda=0.1$ Vanilla reference of $22.91$, despite the roughly thousand-fold rescaling of the deepest pair.

\begin{table}[t!]
\centering
\caption{Single-seed rescaling-strength sweep at 101M parameters and $L=24$ over 30K steps. Here $\Delta$ is SAGGS+P minus Vanilla. The $\lambda=0.3$ difference is $-2.9954$ before rounding.}
\label{tab:medium_lambda_ext}
\vspace{4pt}
\begin{tabular}{@{}lccc@{}}
\toprule
$\lambda$ & \textbf{Vanilla PPL} & \textbf{SAGGS+P PPL} & $\Delta$ \\
\midrule
$0.1$ & $22.91$ & $22.90$ & $-0.01$ \\
$0.2$ & $23.24$ & $22.89$ & $-0.35$ \\
$0.3$ & $25.91$ & $22.92$ & $-3.00$ \\
\bottomrule
\end{tabular}
\end{table}

\FloatBarrier
\subsection{Multi-Seed Comparisons Across Scales}
\label{sec:exp_multiseed}

\paragraph{Motivation.}
We assess consistency across random seeds at 22M, 101M, and 354M parameters, using two or three paired seeds at $\lambda=0$ and $\lambda=0.2$.

\paragraph{Setup.}
We compare Vanilla and SAGGS+P at $\lambda\in\{0,0.2\}$ in three configurations. The 22M ReLU model uses $L=12$, context length 256, and 3 seeds. The 101M ReLU model uses $L=24$, context length 512, 3 seeds at $\lambda=0.2$, and 2 seeds at $\lambda=0$. The 354M GELU model uses $L=24$, context length 512, 2 seeds, and gradient accumulation over 4 micro-batches for an effective batch size of 32. The 22M and 101M models train for 30K steps; the 354M model trains for 20K steps. The medium and 354M comparisons use Q--K-only tilts, learning rate $6\times10^{-4}$, 2K-step warmup, rebalancing every 100 steps, and auxiliary-penalty weight $0.01$. In particular, this 354M recipe uses a higher learning rate and more frequent rebalancing than the 354M stress test in Appendix~\ref{sec:exp_large_reparam}.

\paragraph{Results.}
Table~\ref{tab:multiseed} reports the multi-seed validation. At $\lambda = 0$, SAGGS+P shows negligible difference from Vanilla across these scales, with absolute mean differences below $0.05$ PPL.

At $\lambda = 0.2$, SAGGS+P protects the medium and 354M models, while the small-model comparison favors Vanilla. The medium-scale gap of $\Delta = -0.38 \pm 0.05$ is consistent with the 101M rescaling-strength sweep (Table~\ref{tab:medium_lambda_ext}), and the 354M gap of $\Delta = -1.17 \pm 0.01$ has a large paired effect size in this protocol. This 20K-step cross-scale protocol is distinct from the 15K-step 354M stress experiment (Table~\ref{tab:large_reparam}), where the same mechanism yields a larger $+2.26$ PPL gap; the comparisons agree in sign and differ in horizon and recipe.

\begin{table}[t!]
\centering
\caption{Multi-seed validation across scales. Rows marked $^{\dagger}$ use two seeds; the others use three. $\Delta$ is SAGGS+P minus Vanilla; paired SD uses within-seed differences. Values are mean $\pm$ SD: Medium and 354M entries use sample SD over the complete specified seed sets, while small-model entries retain their original recorded summaries.}
\label{tab:multiseed}
\vspace{4pt}
\begin{tabular}{@{}llccc@{}}
\toprule
\textbf{Scale} & $\lambda$ & \textbf{Vanilla PPL} & \textbf{SAGGS+P PPL} & $\Delta$ \\
\midrule
\multirow{2}{*}{Small, 22M}   & $0$   & $34.39 \pm 0.08$ & $34.38 \pm 0.08$ & $-0.00 \pm 0.06$ \\
 & $0.2$ & $34.20 \pm 0.04$ & $34.30 \pm 0.06$ & $+0.11 \pm 0.05$ \\
\midrule
\multirow{2}{*}{Medium, 101M} & $0^{\dagger}$ & $22.84 \pm 0.05$ & $22.88 \pm 0.04$ & $+0.042 \pm 0.003$ \\
 & $0.2$ & $23.25 \pm 0.03$ & $22.87 \pm 0.03$ & $-0.38 \pm 0.05$ \\
\midrule
\multirow{2}{*}{354M}          & $0^{\dagger}$ & $20.66 \pm 0.03$ & $20.67 \pm 0.04$ & $+0.012 \pm 0.006$ \\
 & $0.2^{\dagger}$ & $21.82 \pm 0.02$ & $20.65 \pm 0.03$  & $-1.17 \pm 0.01$ \\
\bottomrule
\end{tabular}
\end{table}

\FloatBarrier
\subsection{1.2B-Parameter Training}
\label{sec:exp_1b_training}

\paragraph{Motivation.}
The 1.2B gradient-law verification in Appendix~\ref{sec:exp_1b_verify} confirmed the gradient transformation law via a single forward-backward pass. A separate empirical issue is whether the sensitivity seen in smaller-scale training persists at billion-parameter scale. To test this persistence, we train a 1.22B-parameter SwiGLU model for 10,000 optimizer steps using DDP across 4$\times$H100 GPUs.

\paragraph{Setup.}
The SwiGLU Transformer has $d=2048$, $L=22$, $n_{\mathrm{heads}}=16$, and $d_{\mathrm{ff}}=5504$, with gradient checkpointing and BF16 mixed precision. We train on WikiText-103 using AdamW with $\eta=3\times10^{-4}$, $\beta=(0.9,0.95)$, weight decay $0.1$, a cosine schedule with 1,000-step warmup, and gradient clipping at 1.0. A per-GPU batch size of 4 on 4 GPUs with 2 accumulation steps gives an effective batch size of 32. We compare three configurations: unperturbed Vanilla at $\lambda=0$, Vanilla with all-pairs rescaling at $\lambda=0.2$, and SAGGS with the same rescaling and rebalancing every 100 steps.

\paragraph{Results.}
Table~\ref{tab:1b_training} reports the results. The 1.2B baseline achieves $\mathrm{PPL} = 23.81$ after the specified 10K-step training budget. At $\lambda = 0.2$, where $c_{21} = e^{0.2 \times 21} \approx 66.7$ and the all-pairs reparameterization affects Q--K, V--O, and MLP simultaneously, Vanilla PPL severely degrades to $221.24$, a $9.3\times$ increase. The training loss plateaus above $2.7$, compared with a baseline of $1.5$, and does not recover within the 10K-step budget. Before clipping at $1.0$, the mean gradient norm reaches $27.8$, compared with a baseline of $0.68$, with a peak of about 4,700. The final maximum absolute log-ratios are $4.22$ for Q--K, $4.12$ for V--O, and $4.14$ for MLP.

\begin{figure}[t!]
\centering
\includegraphics[width=\linewidth]{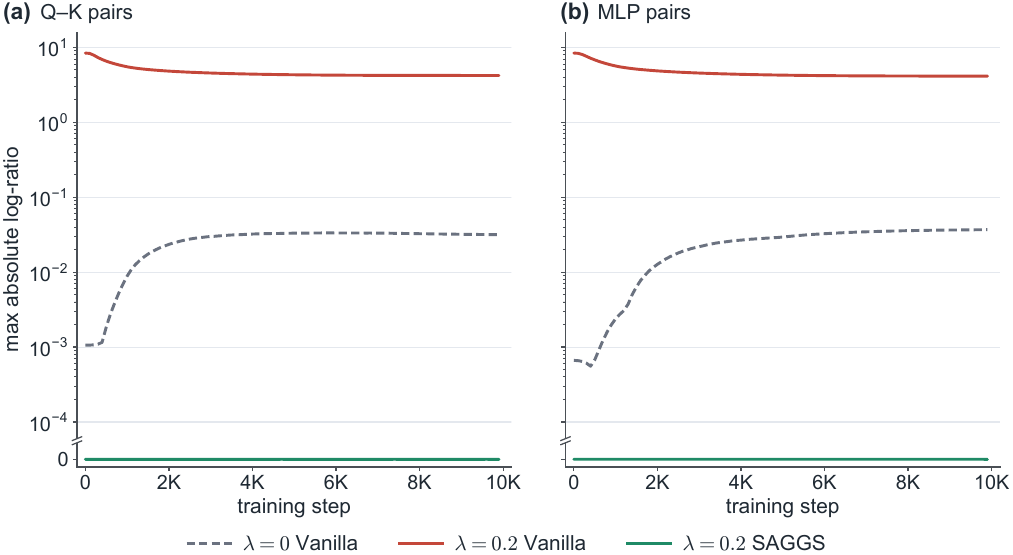}
\caption{Maximum absolute pair log-ratio across layers during 1.2B training: (a)~Q--K and (b)~MLP. The shared vertical scale is linear from $0$ to $10^{-4}$ and logarithmic above; the break marks the transition. Post-rebalance SAGGS samples are at most $4.8\times10^{-7}$ and appear at zero. Samples align with rebalancing events and end at 9.9K; Table~\ref{tab:1b_training} reports the final 10K-step measurements after the intervening updates.}
\label{fig:1b_drift}
\end{figure}

In this single-seed 10K-step run, SAGGS protects against this degradation, achieving $\mathrm{PPL} = 23.70$, slightly better than the baseline, while keeping the final maximum absolute log-ratio below $1.3\times10^{-4}$ for every pair type. V--O is largest at $1.23\times10^{-4}$. Figure~\ref{fig:1b_drift} shows how Q--K and MLP imbalance evolve during this run. This single-seed comparison extends the observed sensitivity to rescaling to billion-parameter training.

\begin{table}[t!]
\centering
\caption{Single-seed 1.2B-parameter SwiGLU training over 10K steps with DDP on 4$\times$H100 GPUs. Gradient norms are measured before clipping at $1.0$.}
\label{tab:1b_training}
\vspace{4pt}
\small
\begin{tabular}{@{}lccc@{}}
\toprule
& $\lambda=0$ & \multicolumn{2}{c@{}}{$\lambda=0.2$} \\
\cmidrule(lr){2-2}\cmidrule(l){3-4}
\textbf{Metric} & Vanilla & Vanilla & SAGGS \\
\midrule
Final validation PPL & $23.81$ & $221.24$ & $\mathbf{23.70}$ \\
Best validation PPL  & $23.81$ & $221.24$ & $\mathbf{23.70}$ \\
\midrule
Mean gradient norm & $0.68$ & $27.79$ & $0.68$ \\
Q--K max-abs log-ratio & $0.032$ & $4.222$ & $0.55\times10^{-4}$ \\
V--O max-abs log-ratio & $0.033$ & $4.117$ & $1.23\times10^{-4}$ \\
MLP max-abs log-ratio & $0.037$ & $4.138$ & $0.43\times10^{-4}$ \\
\bottomrule
\end{tabular}
\end{table}

\paragraph{Interpretation.}
Together with the 1.2B gradient-law verification (Appendix~\ref{sec:exp_1b_verify}), this run shows that both the structural symmetry and the associated optimizer sensitivity appear at billion-parameter size.

\subsection{Exact Contraction and Synthetic Optimizer Diagnostics}
\label{sec:exp59_simulation}

This synthetic study tests the exact disturbance-free contraction law and separately measures Lion sign changes after a soft orbit edit. The sign diagnostic concerns the sampled optimizer state.

\paragraph{Disturbance-free Soft SAGGS contraction.}
Starting from a random pair $(\mat{A}_0,\mat{B}_0)$ with controlled log-ratio $r_0=\log(\frobnorm{\mat{A}_0}/\frobnorm{\mat{B}_0})$, we apply the damped rebalancing map of \eqref{eq:soft_saggs} repeatedly without intervening optimizer updates. The contraction law in \eqref{eq:soft_contraction} predicts $r_K=(1-\alpha)^K r_0$ exactly.  In double precision over $\alpha\in\{0.10,0.25,0.50,0.75,1.0\}$ and three initial imbalances, the reported worst scored single-step relative error is $7.8\times10^{-4}$. This error is scored only when both $|r_k|$ and $|(1-\alpha)r_k|$ exceed $10^{-12}$; tiny denominators can amplify rounding error. For $\alpha=1$, the predicted residual is zero, so this relative-error score has no eligible steps. The 200-event absolute residual is zero for all three starts at $\alpha=1$. The theoretical factor at $\alpha=0.10$ is $0.9^{200}\simeq7.06\times10^{-10}$ after 200 events and $0.9^{150}\simeq1.37\times10^{-7}$ after 150 events.  These factors describe repeated applications of the damped rebalancing map; training-time balance is measured separately in the language-model experiments.

\paragraph{Lion update-sign response to damping.} After a 500-step warmup with $\lambda=0.2$ initial reparameterization, we branch the optimizer state into six parallel runs, each applying Soft SAGGS with a different $\alpha$, and compare the immediate Lion sign update to the $\alpha=0$ branch.  The fraction of update components that flip sign relative to baseline is $0.65\%$ at $\alpha=0.10$, $1.82\%$ at $\alpha=0.25$, $3.91\%$ at $\alpha=0.50$, $5.34\%$ at $\alpha=0.75$, and $7.03\%$ at $\alpha=1.00$. The measured relationship is monotone in $\alpha$ and roughly linear. Hard rebalancing with $\alpha=1$ flips about 11 times as many components as $\alpha=0.10$ in this diagnostic; this pattern is an empirical observation about the sampled state.

\FloatBarrier
\subsection{Cross-Corpus Replication on OpenWebText}
\label{sec:exp62_owt_paired}

The medium-scale paired comparisons in the 101M WikiText-103, multi-seed cross-scale, and Lion experiments all run on a single corpus of about 103M words tokenized with the GPT-2 vocabulary. We repeat the paired comparison of Vanilla and Soft SAGGS with $\alpha=0.5$, crossed with $\lambda=0$ and $\lambda=0.2$, on OpenWebText~\citep{Gokaslan2019OpenWeb}, using a 100M-token GPT-2-tokenized subset, with the same medium-scale architecture as the 101M WikiText-103 comparisons: $L=24$, $d=512$, $H=8$, and 101M parameters, trained with AdamW at $\eta=3\times10^{-4}$, $\beta=(0.9,0.95)$, weight decay $0.1$, a cosine schedule with 1.5K warmup, gradient clipping at $1.0$, batch size $32$, $T=512$, and BF16 autocast, for 15K steps with all-pairs reparameterization. The learning rate, warmup, horizon, and use of all-pairs rather than Q--K-only tilts differ from the 30K-step recipe of the 101M WikiText-103 comparisons, and the optimizer differs from the optimizer of the Lion comparisons. The replication changes both corpus and training recipe. Each configuration uses three paired seeds.

\begin{table}[t!]
\centering
\caption{OpenWebText paired comparison with $n=3$ seeds. The paired block is Vanilla minus Soft SAGGS with $\alpha=0.5$ and reports the two-sided sign-flip $p$-value $p_{\mathrm{sf}}$.}
\label{tab:exp62_owt_cells}
\label{tab:exp62_owt_paired}
\small
\begin{tabular}{@{}lccccccc@{}}
\toprule
& \multicolumn{2}{c}{\textbf{Mean PPL}} & \multicolumn{5}{c}{\textbf{Paired difference}} \\
\cmidrule(lr){2-3}\cmidrule(lr){4-8}
$\lambda$ & Vanilla & Soft, $\alpha=0.5$ & $\bar\Delta$ & $\sigma_\Delta$ & $t$ & $d$ & $p_{\mathrm{sf}}$ \\
\midrule
$0.0$ & $55.92\pm0.07$ & $55.95\pm0.03$ & $-0.032$ & $0.081$ & $-0.69$ & $-0.40$ & $0.500$ \\
$0.2$ & $74.28\pm0.41$ & $56.00\pm0.14$ & $+18.278$ & $0.437$ & $+72.37$ & $+41.79$ & $0.250$ \\
\bottomrule
\end{tabular}
\end{table}

Table~\ref{tab:exp62_owt_cells} reports $55.92\pm0.07$ PPL for unperturbed Vanilla training and $74.28\pm0.41$ after rescaling. At $\lambda=0.2$, the paired Vanilla-minus-Soft difference is $+18.278$ PPL; at $\lambda=0$, the difference is $-0.032$ PPL. The protection has the same direction as the WikiText-103 comparison, but its magnitude is not directly attributable to the corpus change because the optimizer recipe, horizon, and edited pairs also differ. The two-sided sign-flip $p$-value is $0.25$ at $\lambda=0.2$ for three seeds. This experiment provides supporting evidence on another corpus under a different training recipe.

This OpenWebText comparison keeps the GPT-2 BPE vocabulary and the same dense Transformer without GQA or RMSNorm.  The architecture and tokenizer replication in Appendix~\ref{app:exp63_moved} instead tests a smaller but deliberately different LLaMA-like stack~\citep{touvron2023llama,ainslie2023gqa}: RMSNorm, RoPE, SwiGLU, GQA, and a T5 SentencePiece tokenizer.

\FloatBarrier
\subsection{123.8M Modern-Architecture Scale Replication}
\label{sec:exp67_modern_medium}

The 31M replication in Appendix~\ref{app:exp63_moved} changed the architecture and tokenizer at a scale that admits seven paired seeds. We next test the same mechanism in the 100M-parameter regime. We keep the RMSNorm, RoPE, SwiGLU, and GQA block and T5 SentencePiece tokenizer, but use a 24-layer model with $d=640$, ten query heads, two KV heads, SwiGLU hidden width 1728, sequence length 512, and tied embeddings, for 123.8M trainable parameters.

We use the same paired design comparing Vanilla and Soft SAGGS with $\alpha=0.5$, crossed with $\lambda=0$ and $\lambda=0.2$, over three paired seeds. The optimizer is AdamW with $\eta=2.5\times10^{-4}$, weight decay $0.1$, cosine decay, 10\% warmup, gradient clipping at $1.0$, BF16 XPU autocast, batch size 12, sequence length 512, and 15K steps. The reparameterization and Soft SAGGS act on Q--K, V--O, and SwiGLU value--output pairs, with rebalancing every 100 steps. Training uses Intel Data Center GPU Max 1550 XPU tiles.

\begin{table}[t!]
\centering
\caption{123.8M modern-architecture scale replication, 15K AdamW steps, $n=3$. The paired block is Vanilla minus Soft SAGGS with $\alpha=0.5$.}
\label{tab:exp67_modern_medium_cells}
\label{tab:exp67_modern_medium_paired}
\small
\begin{tabular}{@{}lccccccc@{}}
\toprule
& \multicolumn{2}{c}{\textbf{Mean PPL}} & \multicolumn{5}{c}{\textbf{Paired difference}} \\
\cmidrule(lr){2-3}\cmidrule(lr){4-8}
$\lambda$ & Vanilla & Soft, $\alpha=0.5$ & $\bar\Delta$ & $\sigma_\Delta$ & $t$ & $d$ & $p_{\mathrm{sf}}^{+}$ \\
\midrule
$0.0$ & $19.86\pm0.06$ & $19.81\pm0.06$ & $+0.055$ & $0.028$ & $+3.40$ & $+1.96$ & $0.1250$ \\
$0.2$ & $41.12\pm0.65$ & $19.96\pm0.09$ & $+21.156$ & $0.667$ & $+54.90$ & $+31.70$ & $0.1250$ \\
\bottomrule
\end{tabular}
\end{table}

Table~\ref{tab:exp67_modern_medium_cells} reports the three-seed summaries. The 123.8M replication follows the same pattern as the 31M comparison. At $\lambda=0$, the mean Vanilla-minus-Soft difference is $+0.055\pm0.028$ PPL.

At $\lambda=0.2$, Vanilla ends at PPL $41.12\pm0.65$, whereas Soft SAGGS ends at $19.96\pm0.09$; the paired contrast is $\bar\Delta=+21.156\pm0.667$ PPL with all three seeds favoring Soft SAGGS.  With $n=3$, the one-sided exact sign-flip $p$-value floor is $0.125$.

\FloatBarrier
\subsection{296.7M Modern-Architecture Scale Replication}
\label{sec:exp69_modern_large}

The modern-architecture replications at 31M and 123.8M parameters show sensitivity to rescaling. We next test the same paired design at 296.7M parameters, increasing both the width and depth of the RMSNorm, RoPE, SwiGLU, and GQA architecture.

\paragraph{Design.}
The model uses the same T5 SentencePiece tokenization and causal RMSNorm, RoPE, SwiGLU, and GQA block, with $d=896$, 14 query heads, two KV heads, SwiGLU hidden width 2432, 32 layers, tied embeddings, sequence length 512, and 296.7M trainable parameters.  We train the paired grid of Vanilla and Soft SAGGS with $\alpha=0.5$, crossed with $\lambda=0$ and $\lambda=0.2$, over three paired seeds for 12K AdamW steps on WikiText-103 with learning rate $2.0\times10^{-4}$, weight decay $0.1$, 10\% warmup, cosine decay, gradient clipping at $1.0$, BF16 XPU autocast, batch size 8, and rebalancing every 100 steps.  As in the 123.8M scale replication and checkpoint-surgery resume, the intervention acts on Q--K, V--O, and SwiGLU value--output pairs.  The run uses all 12 Intel Data Center GPU Max 1550 XPU tiles.

\begin{table}[t!]
\centering
\caption{296.7M modern-architecture scale replication, 12K AdamW steps, $n=3$. The paired block is Vanilla minus Soft SAGGS with $\alpha=0.5$.}
\label{tab:exp69_modern_large_cells}
\label{tab:exp69_modern_large_paired}
\small
\begin{tabular}{@{}lccccccc@{}}
\toprule
& \multicolumn{2}{c}{\textbf{Mean PPL}} & \multicolumn{5}{c}{\textbf{Paired difference}} \\
\cmidrule(lr){2-3}\cmidrule(lr){4-8}
$\lambda$ & Vanilla & Soft, $\alpha=0.5$ & $\bar\Delta$ & $\sigma_\Delta$ & $t$ & $d$ & $p_{\mathrm{sf}}^{+}$ \\
\midrule
$0.0$ & $25.51\pm0.04$ & $25.42\pm0.10$ & $+0.088$ & $0.052$ & $+2.96$ & $+1.71$ & $0.1250$ \\
$0.2$ & $111.45\pm4.83$ & $25.64\pm0.10$ & $+85.813$ & $4.730$ & $+31.42$ & $+18.14$ & $0.1250$ \\
\bottomrule
\end{tabular}
\end{table}

Table~\ref{tab:exp69_modern_large_cells} reports the three-seed summaries. At $\lambda=0$, Soft SAGGS achieves slightly lower PPL in all three paired seeds, with $\bar\Delta_{\rm Vanilla-Soft}=+0.088\pm0.052$.  Under the same exact $\lambda=0.2$ all-pairs reparameterization, Vanilla ends at $111.45\pm4.83$ PPL, while Soft SAGGS ends at $25.64\pm0.10$ PPL; the paired protection is $+85.813\pm4.730$ PPL with all three seeds favoring Soft SAGGS.

The final pair-norm statistics show balance control: Vanilla runs at $\lambda=0.2$ retain large maximum absolute log-ratios, Q--K about $7.5$, V--O about $5.4$, and MLP about $6.4$, whereas Soft SAGGS keeps all three below $2.5\times10^{-4}$.  Absolute PPL is tokenizer- and budget-dependent; the contrast is the within-tokenizer, seed-paired PPL recovery under deliberate rescaling at a 300M-class modern-architecture scale.

\FloatBarrier
\section{Further Discussion}
\label{app:discussion_moved}
\label{sec:discussion}

The results distinguish invariance of the current function from equivariance of an optimizer trajectory. Standard adaptive and sign-based rules can cancel gradient scales while retaining a dependence on parameter scales. Transforming moments alone does not correct the update-scaling mismatch under shared hyperparameters. Full-state canonicalization addresses this dependence at restart by bringing orbit-related checkpoints to the same numerical state.

Periodic rebalancing serves a different practical role: controlling pair-scale imbalance as training proceeds. Its strongest measured benefits occur after deliberately imposed rescaling, including the checkpoint-resume workflows. Imbalance also arises naturally, in both trained-from-scratch and public models, and rebalancing can be kept on throughout training to control this imbalance: A rebalance rescales each pair only as much as the pair is imbalanced, correcting a large imbalance and leaving pairs that are already close to balance almost untouched. Soft SAGGS brings the same control to Lion, whose update depends on a momentum buffer of past gradients: Soft SAGGS leaves the optimizer state unchanged and applies the correction in several smaller steps, so the update changes gradually, and training recovers from the degradation caused by an edit.

The normalization layers of a Transformer, either layer normalization or RMSNorm, do not remove the pair symmetries, because these layers act on the residual stream rather than on the paired matrices. Weight decay does not restore pair balance either, because the decay shrinks both matrices of a pair by the same factor. Global gradient clipping rescales all gradients by one common factor, set by their total norm, so the gradients enlarged by an orbit edit can shrink the gradients of all other blocks. The following subsections expand on this discussion, covering adaptive and sign-based optimizers, natural drift and weight decay, initialization, and training stability.

\subsection{Adaptive Normalization and Trajectory Equivariance}
\label{app:disc_adaptive_moved}
\label{sec:disc_adaptive}
\label{app:theory_interp_moved}

Adam and AdamW normalize gradients using per-parameter moment estimates. With covariantly transformed moments and $\epsilon=0$, their normalized directions are unchanged by reciprocal rescaling. Equivariance requires the parameter displacements to scale with the parameters, which a shared learning rate does not provide (Theorem~\ref{thm:adam_equivariance}). Theorem~\ref{thm:equivariant_rules} extends this obstruction to the standard block-local rules considered here.

For ideal degree-zero rules on corresponding histories, Corollary~\ref{cor:half_compensation} shows that the distortion in the log-ratio of relative data updates is half the distortion of gradient descent. This factor concerns the data-update term under the stated assumptions; full increments can also reflect weight decay, a fixed positive $\epsilon$, and differences in clipping.

Theorem~\ref{thm:adam_equivariance} supplies a mechanism for a residual surgery gap even with correctly transformed moments. The depth and $\lambda$ sweeps measure the training response, while the state-preserved, stale-state, and partial-state experiments (Appendices~\ref{app:exp70_moved}, \ref{sec:exp71_stale_state}, and~\ref{sec:exp72_partial_state}) examine the additional role of moment history.

\paragraph{Adaptive-optimizer qualifications.}
Reset or stale moments and time-varying orbit moves can break the correspondence between current gradients and past statistics. A fixed positive $\epsilon$ must also be transformed according to Theorem~\ref{thm:adam_equivariance} to preserve trajectory equivariance. Global gradient clipping generally changes its common clipping multiplier under a nonuniform diagonal rescaling of the gradient. Decoupled weight decay alone commutes with a fixed diagonal rescaling when $\eta\omega$ is unchanged. If learning rates are transformed to restore update equivariance, the decay coefficients must also be transformed to keep this product fixed.

\subsection{Sign-Based Optimizers and Soft Rebalancing}
\label{sec:disc_lion}

Lion computes its update sign from a weighted combination of the current gradient and its momentum buffer, an EMA of past gradients. A positive scalar rescaling of both the gradient and the buffer leaves this sign unchanged. As in Example~\ref{rem:adam_moment_covariance}, this sign invariance produces the same numerical displacement, while a rescaled parameter trajectory requires a rescaled displacement.  Sign invariance does not imply immunity to a symmetry tilt, even before accounting for clipping or a stale momentum buffer. Theorem~\ref{thm:equivariant_rules} covers Lion directly as a degree-zero rule. The hard- and soft-rebalancing Lion comparisons (Appendices~\ref{app:exp41_moved} and~\ref{app:exp66_moved}) measure the resulting training sensitivity: The $\lambda=0.2$ tilt degrades Vanilla Lion training, and rebalancing recovers PPL in those comparisons.

The Lion state ablation in Appendix~\ref{app:exp64_moved} reports a higher unperturbed mean PPL and greater seed variation for hard SAGGS than for Soft or state-covariant rebalancing. Soft SAGGS limits the correction at each event by applying only an $\alpha$ fraction of the log-norm correction, as in \eqref{eq:soft_contraction}. Optimizer updates can introduce new imbalance between events. The synthetic study in Appendix~\ref{sec:exp59_simulation} separately reports a sampled sign-response diagnostic. For sign-based optimizers, we use $\alpha=0.5$ as the default, retain the $\lambda=0.2$ protection, and assess the unperturbed cost with paired controls at $\lambda=0$.

\subsection{Natural Drift and Weight Decay}
\label{sec:disc_natural_drift}
\label{app:disc_natural_drift_moved}
\label{app:discussion_detail_moved}

Pair imbalance develops during ordinary training, but the direction of that drift depends on the radial components of the optimizer updates, not just their magnitudes. Appendix~\ref{app:balance_diagnostics_moved} states the exact defect recursion and the first-order log-ratio identity. Appendix~\ref{app:balance_dynamics_proof} gives the derivations. Common weight decay scales both matrices equally and therefore leaves the ratio unchanged. Hard SAGGS restores balance at each executed event, while Soft SAGGS contracts the log-ratio. The effect on validation loss depends on the subsequent training trajectory. Layer normalization preserves the internal pair identities, as shown in Appendix~\ref{app:layernorm_detail}.

\subsection{Implications for Initialization}
\label{app:init_full}

Our analysis reveals that standard initialization schemes, \eg{} Xavier and Kaiming, implicitly choose a specific point in the continuous family of forward-equivalent parameterizations. The gradient geometry at this point need not be the best one for a fixed optimizer, horizon, and regularization recipe. Specifically, if $\mat{W}^Q_\ell$ and $\mat{W}^K_\ell$ are initialized with the same variance, as is standard, the squared Frobenius norms have the same expectation when their shapes match; independent samples are not exactly balanced. Training can then change the imbalance. Residual-output initialization and GQA also give systematic norm differences, so $\lambda=0$ means no deliberate tilt, not exact balance of every pair.

SAGGS enforces balance at executed hard events. Soft SAGGS contracts the current log-ratio. Their effects on validation PPL are evaluated against the unperturbed training controls.

\subsection{Training Stability}
\label{app:stability_full}
\label{sec:disc_stability}

Gradient geometry distortion can also affect training stability. When the reparameterization amplifies gradients of $\mat{B}_\ell$ by $e^{\lambda\ell}$ while attenuating the gradients of $\mat{A}_\ell$ by $e^{-\lambda\ell}$, the spread of gradient norms can change in either direction, depending on the baseline gradients. The amplified $\mat{B}_\ell$ gradients may dominate the global norm and cause clipping to scale the current gradients of every block. This clipping leaves every nonzero pair-gradient norm ratio unchanged. For gradient descent, the clipping scales the data-update direction by the same factor, but the effect on AdamW depends on both moment histories. Theorem~\ref{thm:adam_equivariance} excludes clipping, whose multiplier can change under an orbit edit. Decoupled decay is separate from this gradient clipping.

We track gradient norm spikes, defined as steps where the total gradient norm exceeds $\mu + 3\sigma$, across our experiments. While the absolute number of spikes does not change substantially with moderate $\lambda$ under AdamW in the reported runs, the distribution of gradient mass across parameter groups shifts systematically, which can be detrimental for training efficiency. SAGGS controls pair-norm imbalance; the resulting gradient-norm distribution still depends on the training trajectory.

\FloatBarrier
\clearpage
\section{Released-Checkpoint Case Studies}
\label{sec:qualitative_cases}

The paired experiments establish protection and recovery in controlled training runs. We complement these experiments with four case studies on public checkpoints: naturally occurring pair-scale variation, coordinate-sensitive fine-tuning, the internal evolution of a Lion stress test, and token-level restart differences on real text. These measurements are independent of the primary training runs. Appendix~\ref{sec:released_checkpoint_summary} summarizes the case studies, and Appendix~\ref{app:qualitative_protocol} gives the protocols.

\subsection{Pair Scales in Public Checkpoints}
Figure~\ref{fig:qual_public_orbits} extends the GPT-2 endpoint measurement of Appendix~\ref{sec:exp_pretrained_drift} with a cross-model survey and intermediate Pythia checkpoints. The intermediate checkpoints distinguish initialization offsets from changes accumulated during training: Q--K starts nearly balanced, but reaches a maximum absolute log-ratio of $1.86$ in Pythia 410M, concentrated in deep layers. Pythia 1B remains within $0.04$, and V--O initially moves toward balance. Drift is model- and pair-dependent. These observations show pair-scale variation during ordinary training, without measuring its effect on loss. Training also changes the function, so these checkpoints are not points on a single fixed symmetry orbit.

\begin{figure}[t!]
\centering
\includegraphics[width=\linewidth]{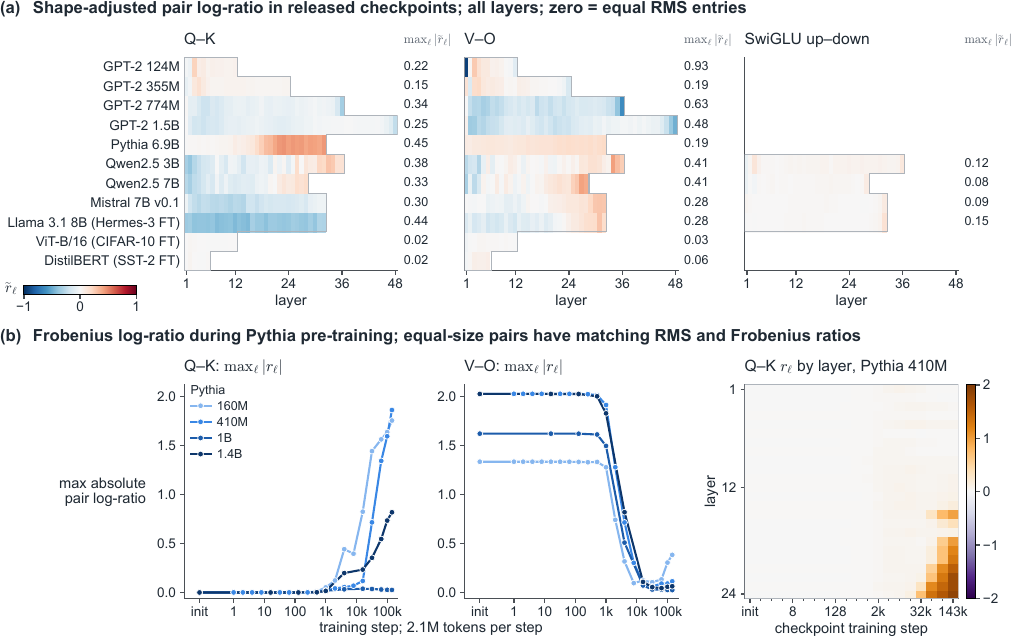}
\caption{Pair-scale variation in released checkpoints. (a)~Eleven checkpoints, using the shape-adjusted log-ratio $\widetilde r_\ell=\log(\frobnorm{A_\ell}/\frobnorm{B_\ell}) -(1/2)\log(n_A/n_B)$, where $n_A$ and $n_B$ count matrix entries. Red denotes larger root mean square (RMS) entry magnitudes in $A_\ell$ and blue denotes larger RMS entry magnitudes in $B_\ell$; zero means equal RMS, which differs from Frobenius balance for unequal-size GQA pairs. Blank SwiGLU rows denote GELU MLPs. (b)~Four Pythia pre-training runs and layerwise Q--K log-ratios for 410M. Equal matrix sizes make RMS and Frobenius log-ratios coincide. Appendix~\ref{app:qualitative_protocol} gives the protocols.}
\label{fig:qual_public_orbits}
\end{figure}

\clearpage
\subsection{Fine-Tuning from Equivalent Checkpoints}
\begin{figure}[t!]
\centering
\includegraphics[height=0.62\textheight,keepaspectratio]{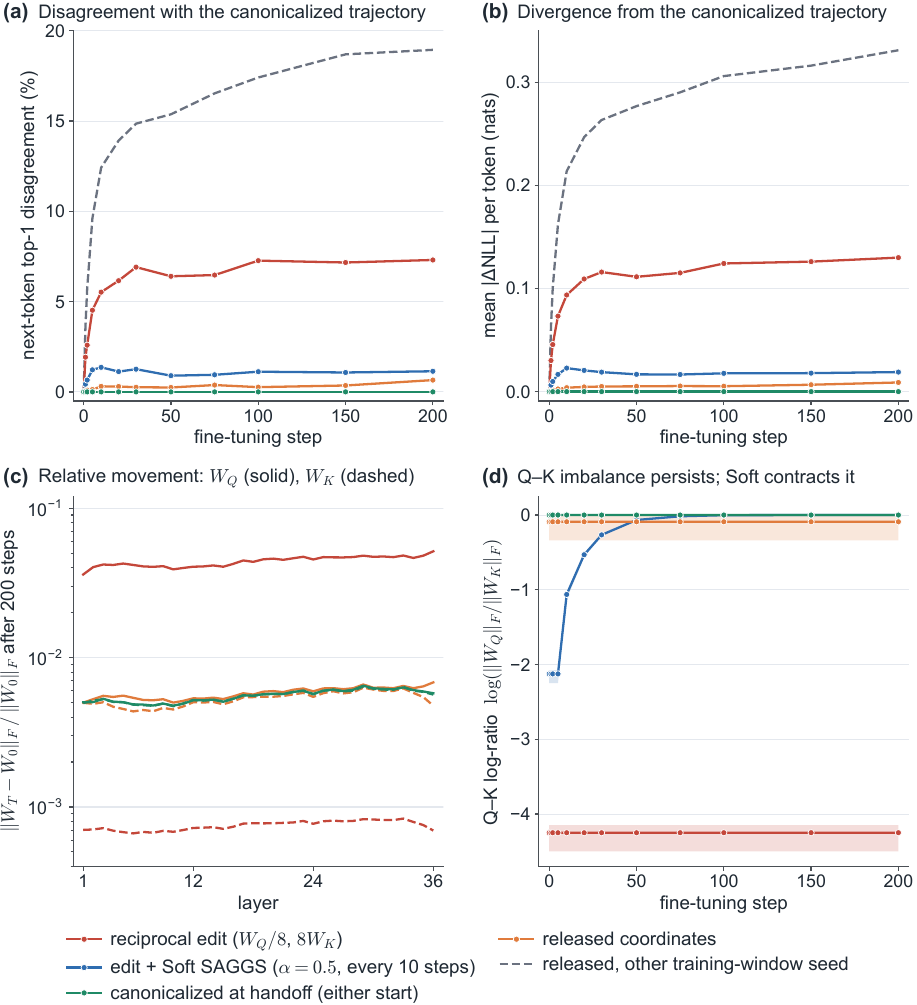}
\caption{Fine-tuning from forward-identical GPT-2 774M checkpoints: 200 AdamW updates with fresh moments and 16,384 held-out targets. (a)~Top-1 disagreement and (b)~mean absolute token-NLL difference, relative to the canonicalized branch. Canonicalizing either start gives the same trajectory, shown in green. The dashed reference uses different sampled training windows; the reference is not an uncertainty band. (c)~Net Q--K movement relative to each branch's starting norm, a cumulative rather than one-step measurement. (d)~Q--K log-ratios: layer mean and min--max band. Soft SAGGS uses $\alpha=0.5$ at handoff and every ten steps. Appendix~\ref{app:qualitative_protocol} gives the protocols.}
\label{fig:qual_released_finetune}
\end{figure}

We next branch from the released GPT-2 774M checkpoint with fresh AdamW moments (Figure~\ref{fig:qual_released_finetune}). An exact reciprocal edit $W_Q\mapsto W_Q/8$, $W_K\mapsto8W_K$ preserves the current function but leads to $7.3\%$ held-out top-1 disagreement with the canonicalized trajectory after 200 updates. Canonicalizing either starting representative gives identical subsequent predictions and token negative log-likelihoods (NLLs) at every evaluation. Soft SAGGS leaves $1.1\%$ disagreement. Continuing from the released coordinates gives $0.65\%$ disagreement with the same canonicalized reference. This comparison illustrates Corollary~\ref{cor:canonical_restart} with reset moments and distinguishes exact restart equivalence from the approximate recovery provided by Soft SAGGS. The reciprocal edit is imposed for this controlled comparison; trajectory differences do not establish a long-run loss advantage.

\clearpage
\subsection{Depth-Resolved Evolution under Lion}
\begin{figure}[t!]
\centering
\includegraphics[height=0.62\textheight,keepaspectratio]{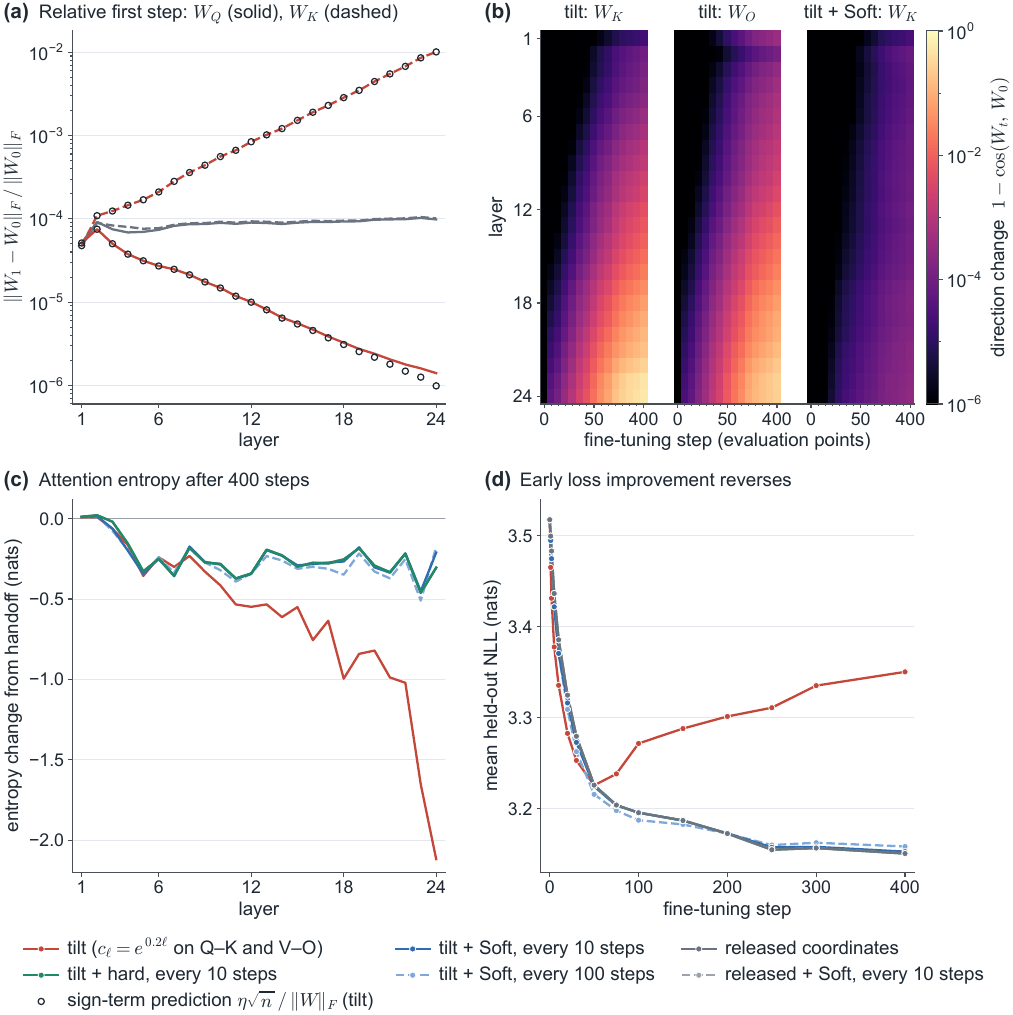}
\caption{Depth-resolved response to an exact attention-pair tilt $c_\ell=e^{0.2\ell}$ in GPT-2 355M with $L=24$ during 400 Lion steps. (a)~First-step relative Q--K movement and the sign-term prediction $\eta\sqrt n/\frobnorm{W}$; full updates also include decay. (b)~Direction change $1-\cos(W_t,W_0)$ for tilted $W_K$ and $W_O$, and $W_K$ with tilt + Soft every ten steps. Columns are 15 evaluation steps on a nonlinear time axis. (c)~Attention-entropy change from handoff after 400 steps, averaged over 16 fixed sequences, heads, and query positions. (d)~Mean held-out NLL. Rebalancing includes handoff and leaves buffers unchanged; Soft uses $\alpha=0.5$. The figure shows a single-seed stress test; protocols are in Appendix~\ref{app:qualitative_protocol}.}
\label{fig:qual_lion_tilt}
\end{figure}

Figure~\ref{fig:qual_lion_tilt} connects the degree-zero relative-update law to the time course of a public-model stress test. The deepest key block receives a relative first step of about $1\%$ under the imposed tilt, compared with $0.01\%$ in released coordinates. Its cosine similarity with the starting matrix falls to $0.46$ after 400 steps, accompanied by a larger attention entropy decrease in deep layers and a reversal of the early loss advantage. The cosine diagnostic is invariant to positive scalar rescalings and therefore separates directional change from the balancing operation itself. Soft SAGGS limits these changes, and hard rebalancing also performs well in this setting. The preference for damping is supported by the unperturbed Lion state ablation in Appendix~\ref{app:exp64_moved}. Here, attention entropy and loss are measured jointly, without isolating a causal relation between the two quantities.

\clearpage
\subsection{Token-Level Differences after Restart}
\begin{figure}[t!]
\centering
\includegraphics{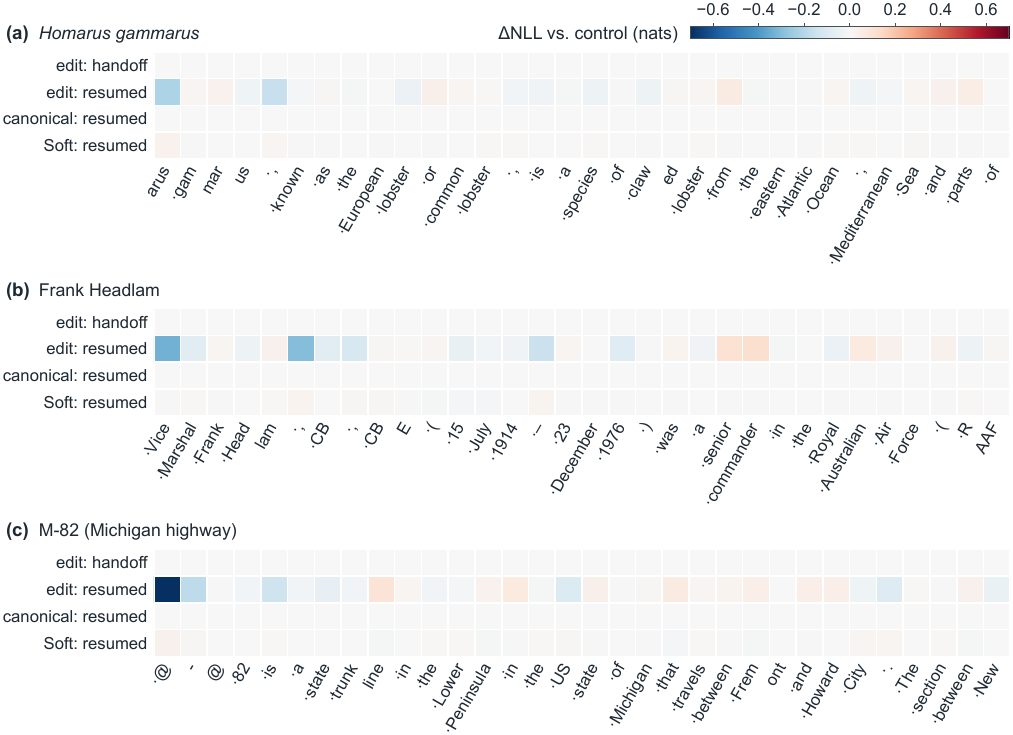}
\caption{Token-level restart differences in GPT-2 124M. (a), (b), and (c)~Outcome-independent excerpts from three WikiText-103 validation articles. Columns are ground-truth BPE tokens conditioned on preceding text; a leading dot denotes a space and \texttt{@-@} is the corpus hyphen. Rows show branch-minus-control token NLL at handoff or after 30 shared AdamW updates. Blue denotes increased and red denotes decreased target probability on a shared signed scale. The edit co-transforms moments; the canonical branch additionally balances weights and moments, while Soft SAGGS with $\alpha=0.5$ leaves buffers unchanged at rebalancing. Appendix~\ref{app:qualitative_protocol} gives the protocols.}
\label{fig:qual_token_restart}
\end{figure}

Figure~\ref{fig:qual_token_restart} makes the full-state distinction visible at individual tokens. After a moment-covariant edit, the probability of \texttt{ Marshal} rises from the control's $0.0582$ to $0.0628$, while \texttt{ commander} falls from $0.0242$ to $0.0215$ after resuming. Full-state canonicalization restores both, within finite-precision residuals. The edited branch's mean NLL is $0.01136$ nats lower on this evaluation set, illustrating that trajectory divergence can also improve predictions. Soft SAGGS leaves a small residual difference from the control; full-state canonicalization recovers the reference trajectory.

\clearpage
\subsection{Summary}
\label{sec:released_checkpoint_summary}

Case studies on public checkpoints complement the paired training comparisons. The case studies show that pair imbalance also arises in ordinary training, to a degree that varies across models and pairs. In Pythia 410M, for example, the query and key matrices start with nearly equal norms, but during pre-training their norms drift apart by a factor of up to about $6$ in deep layers, which corresponds to a log-norm ratio of $1.86$.

The case studies also test canonical restart on a released model. We fine-tune GPT-2 774M with fresh AdamW moments, starting from the released checkpoint and from an exact Q--K edit of this checkpoint that maps $W_Q\mapsto W_Q/8$ and $W_K\mapsto 8W_K$. Canonicalizing both checkpoints at restart makes the two runs identical, with the same predictions and token NLLs at every evaluation. Without canonicalization, the edited run predicts a different top-1 token than the canonicalized runs at $7.3\%$ of held-out positions after 200 updates. Periodic Soft SAGGS reduces this fraction to $1.1\%$ but not to zero. Canonical restart also works when the moments are kept rather than reset: In a GPT-2 124M case whose edit transforms the moments along with the weights, canonicalizing both weights and moments fully restores the trajectory of the unedited run.

\subsection{Protocols}
\label{app:qualitative_protocol}

The four figures in this appendix use supplementary measurements on public checkpoints, independently of the primary training runs. Training case studies use a single fixed seed, FP32, deterministic PyTorch operations, eager attention, and no dropout, TensorFloat-32, or gradient clipping. Within each comparison, all branches use the same fixed training batches, except for the explicitly marked alternative-sampling reference. The training runs use one GPU; the checkpoint-norm survey reads weight tensors on CPU. The case studies are descriptive, single-seed measurements, not additional multi-seed PPL comparisons. Evaluation subsets contain teacher-forced next-token targets, not sampled generations or full-validation PPL measurements.

\paragraph{Public checkpoint survey (Figure~\ref{fig:qual_public_orbits}).}
Figure~\ref{fig:qual_public_orbits}(a) includes GPT-2 124M, 355M, 774M and 1.5B; Pythia 6.9B; Qwen2.5 3B and 7B; Mistral 7B v0.1; Hermes-3-Llama-3.1-8B; a ViT-B/16 CIFAR-10 classifier; and the DistilBERT SST-2 classifier. All layers and available exact matrix pairs are retained, with biases excluded from norms. Fused GPT-2 and Pythia projections are split into Q, K, and V. Pythia stores these projections grouped by head. SwiGLU uses only its up--down value path. GELU MLPs are excluded. Figure~\ref{fig:qual_public_orbits}(a) removes the shape offset $(1/2)\log(n_A/n_B)$, and therefore measures equal-RMS rather than equal-Frobenius balance for GQA. The measured Qwen2.5-3B-Instruct row is omitted because its log-ratios differ from the base by at most $1.9\times10^{-4}$.

Figure~\ref{fig:qual_public_orbits}(b) uses public Pythia 160M, 410M, 1B and 1.4B checkpoints from step 0 through step 143000: 20 each for 160M and 410M and 11 each for 1B and 1.4B. Each training step represents 2,097,152 tokens. Pythia 2.8B was excluded because its early checkpoints contained the same already-trained weights as the endpoint, preventing an initialization comparison. Q--K starts within $0.003$ of balance in all four plotted models. V--O starts near $\log(0.316L)$, consistent with the initialization offset. The model sizes differ in architecture dimensions; no monotone size trend or performance consequence is inferred from these observations.

\paragraph{Fine-tuning and Lion protocols.}
Figures~\ref{fig:qual_released_finetune} and~\ref{fig:qual_lion_tilt} use the released GPT-2 774M (\path{openai-community/gpt2-large}) and 355M (\path{openai-community/gpt2-medium}) checkpoints, respectively. Both use batch size four, context length 128, and fixed random windows from a prefix of the WikiText-103 raw training split containing just over 600,000 tokens, with end-of-text separators. The evaluation set comprises the first 128 non-heading validation paragraphs with at least 129 GPT-2 tokens, truncated to 129 tokens each, yielding 16,384 targets.

The 774M run starts each branch with fresh AdamW moments, learning rate $2\times10^{-5}$, $(\beta_1,\beta_2)=(0.9,0.95)$, $\epsilon=10^{-8}$, decay $0.1$, and 200 updates at constant learning rate without warmup. Q--K reciprocal edits co-transform their biases and leave the attention scaling operation unchanged. Canonicalization and Soft operate on Q--K and V--O. Q, K, and V biases follow their weight scales, while attention output biases are untouched. Soft uses $\alpha=0.5$ at handoff and every ten steps without moment transformation. The alternative-sampling branch uses a different seed to draw new training windows. In Figure~\ref{fig:qual_released_finetune}(a) and Figure~\ref{fig:qual_released_finetune}(b), this branch is compared to the canonicalized branch, just like the other curves, and is a single reference, not a variability estimate. The measured handoff maximum logit difference is $8.8\times10^{-5}$ on the first 16 evaluation sequences, with zero top-1 disagreements over all evaluation targets. The two canonicalized branches have identical predictions and token NLLs at all 12 evaluation steps.

The canonicalized branch's mean held-out NLL falls from $3.389$ to $2.975$ nats. At step 200, the reciprocal-edit, released, and Soft SAGGS branches exceed this NLL by $1.34\times10^{-2}$, $6.1\times10^{-4}$, and $2.3\times10^{-5}$ nats, respectively. Median Q-to-K relative net movement is about $58$ for the edited branch and near one after canonicalization.

The 355M run uses Lion with a zero-initialized momentum buffer, constant learning rate $10^{-5}$, $\beta=(0.9,0.99)$, decay $0.1$ on matrices and no decay on vectors, and 400 updates without warmup. The tilt acts on Q--K and V--O in all 24 layers; biases follow the exact transformations above. Branches comprise released, released+Soft, tilt, tilt+Soft every ten or 100 steps, and tilt+hard every ten steps. Every rebalancing branch also corrects at handoff and leaves buffers unchanged. In particular, the interval-100 Soft branch already has a smaller initial imbalance than the uncorrected tilt branch. Relative steps are measured before each periodic rebalance, and cosine changes are measured against each branch's own post-handoff matrices. The handoff maximum logit residual is below $0.013$ on the logit probe, with zero disagreements on all evaluation targets. For K and O, measured step norms agree with the sign-only prediction within $0.11\%$ throughout the run; decay contributes more for the inflated Q blocks. The large imposed scale is a stress test, not an estimate of typical naturally occurring imbalance.

At layer 24, the relative first-step magnitude of $W_K$ divided by the relative first-step magnitude of $W_Q$ is 7,166 under the tilt and $1.03$ in released coordinates. Final NLL is $3.350$ for the tilted branch, $3.151$ for released, $3.153$ for tilt + Soft, and $3.151$ for tilt + hard.

\paragraph{Token-level restart protocol.}
Figure~\ref{fig:qual_token_restart} uses the released GPT-2 124M model (\path{openai-community/gpt2}) and its tokenizer. We first perform 20 AdamW adaptation steps, then exactly balance attention pairs and co-transform moments to obtain a shared balanced starting state. This complete state is cloned into the control and the edited branches. The edit uses $c_\ell=e^{0.2\ell}$ for $\ell=0,\ldots,11$ and maps $m\mapsto S^{-1}m$, $v\mapsto S^{-2}v$. The full-state branch additionally canonicalizes both weights and moments before resuming. Soft applies $\alpha=0.5$ at handoff and every five steps, leaving buffers unchanged at its own events. This design distinguishes full-state canonical restart from the default buffer-preserving periodic method.

All branches resume for 30 updates with batch four, context 128, constant learning rate $2\times10^{-5}$, AdamW betas $(0.9,0.95)$, $\epsilon=10^{-8}$, and decay $0.1$. Fixed training windows are drawn from a prefix of the WikiText-103 raw training split containing just over 200,000 tokens. The evaluation set takes the first 64 non-heading validation paragraphs with at least 65 tokens and scores their first 64 next-token targets. Displayed cases are the first qualifying paragraphs of the first three distinct validation articles, located at source rows 3, 49, and 80, using their first 33 tokens; the initial token provides context. Selection does not depend on the branch outcomes. Original WikiText punctuation is retained. For the 96 displayed target positions, the maximum handoff logit discrepancy is $2.29\times10^{-4}$; after resuming, the discrepancy is $3.08$ for the edited branch and $1.53\times10^{-4}$ for full-state canonicalization. The three passages are illustrative cases, not independent training replicates.

Across all 4,096 evaluation targets, the branches initially agree on top-1 predictions. Final disagreement with control is $3.32\%$ with mapped moments alone, $0\%$ with full-state canonicalization, and $0.20\%$ with Soft SAGGS. These comparisons measure restoration of the reference trajectory, not a general quality advantage.

\FloatBarrier
\clearpage
\section{List of Notation}
\label{app:notation}

Table~\ref{tab:notation} lists the main mathematical notation.

\begin{table}[t!]
\centering
\caption{Main mathematical notation used in the paper.}
\label{tab:notation}
\vspace{4pt}
\small
\begin{tabular}{@{}ll@{}}
\toprule
\textbf{Symbol} & \textbf{Meaning} \\
\midrule
\multicolumn{2}{@{}l}{Architecture and loss} \\
$\param$ & model parameters \\
$\loss(\param)$ & data loss \\
$L$, $\ell$ & depth and layer index \\
$d$, $d_k$, $d_{\mathrm{ff}}$ & hidden, head, and MLP widths \\
$\mat{W}^Q_\ell$, $\mat{W}^K_\ell$, $\mat{W}^V_\ell$, $\mat{W}^O_\ell$ & attention projections \\
$\mat{W}^1_\ell$, $\mat{W}^2_\ell$ & ReLU MLP up and down projections \\
$\mat{W}^{\mathrm{up}}_\ell$, $\mat{W}^{\mathrm{gate}}_\ell$ & SwiGLU value and gate \\
$\frobnorm{\cdot}$ & Frobenius norm \\
\midrule
\multicolumn{2}{@{}l}{Symmetry and orbits} \\
$(\mat{A},\mat{B})$ & exact scalar symmetry pair \\
$c>0$ & reciprocal scale factor \\
$T_c(\mat{A},\mat{B})$ & orbit map $(c\mat{A},c^{-1}\mat{B})$ \\
$S=\mathrm{diag}(s_i)$ & coordinatewise orbit map \\
$\lambda$ & depth-schedule tilt \\
$c_\ell\coloneqq e^{\lambda\ell}$ & layerwise orbit scale \\
$\rho_p$ & pair-norm ratio $\frobnorm{\mat{A}}/\frobnorm{\mat{B}}$ \\
$c_p^\star$ & balancing scale $\rho_p^{1/2}$ \\
$R_{\param}$ & pair-balancing diagonal map \\
$\mathcal{C}$ & full-state canonicalizer \\
$\mathcal{P}$, $p$ & set and index of disjoint exact pairs \\
\midrule
\multicolumn{2}{@{}l}{Optimizer updates} \\
$\eta_t$ & learning rate \\
$\omega_t$ & decoupled weight-decay coefficient \\
$\epsilon$ & Adam additive stabilizer \\
$\beta_1$, $\beta_2$ & Adam moment-estimate decay rates \\
$m$, $\hat{m}$ & first-moment estimate and bias-corrected version \\
$v$, $\hat{v}$ & second-raw-moment estimate and bias-corrected version \\
$\mat{U}^A_t$, $\mat{U}^B_t$ & pair optimizer directions \\
$\Psi_t$ & block-local direction map \\
$k$ & homogeneity degree of $\Psi_t$ \\
$\Delta\mat{A}$, $\Delta\mat{B}$ & parameter increments \\
\midrule
\multicolumn{2}{@{}l}{Diagnostics and SAGGS} \\
$\Delta_t$ & balance defect $\frobnorm{\mat{A}}^2-\frobnorm{\mat{B}}^2$ \\
$r_t$ & pair log-norm ratio \\
$F_t$ & radial forcing \\
$Q_t$ & squared-direction-norm difference \\
$T_{\mathrm{bal}}$ & rebalancing interval \\
$\alpha$ & Soft SAGGS damping \\
$c^{\mathrm{eff}}_p(\alpha)$ & damped orbit factor \\
\midrule
\multicolumn{2}{@{}l}{Experiments} \\
$\bar\Delta_{\mathrm{PPL}}$ & mean paired PPL difference \\
$\sigma_\Delta$ & paired SD \\
$d$ & paired Cohen's $d$ \\
$p_{\mathrm{sf}}^{+}$ & one-sided sign-flip $p$-value \\
$n$ & number of paired seeds \\
\bottomrule
\end{tabular}
\end{table}

\FloatBarrier
\section{Limitations}
\label{app:limitations}
\label{sec:limitations}

The experiments evaluate a selected collection of Transformer architectures, optimizers, and language-modeling datasets. The experiments do not cover every model family or every corpus, so the empirical findings should be interpreted within this experimental scope.

\end{document}